\documentclass[11pt]{article}

	\usepackage[final]{acl}

	\usepackage{times}
	\usepackage{latexsym}

	\usepackage[T1]{fontenc}

	\usepackage[utf8]{inputenc}

	\usepackage{microtype}

	\usepackage{inconsolata}

	\usepackage{graphicx}

	\usepackage{hyperref}
	\usepackage{amssymb}
	\usepackage{multirow}
	\usepackage{booktabs}
	\usepackage{csquotes}
	\usepackage{listings}
	\usepackage{algorithm}
	\usepackage{algpseudocode}

\algnewcommand{\algorithmicprocess}{\textbf{Process:}}
\algnewcommand{\Process}{\item[\algorithmicprocess]}
	\usepackage[most]{tcolorbox}
	\usepackage{mdframed}
	\usepackage{fancybox}
\lstdefinestyle{promptstyle}{basicstyle=\ttfamily\scriptsize,breaklines=true,columns=fullflexible,keepspaces=true,aboveskip=2pt,belowskip=2pt,literate={·}{{$\cdot$}}1}
\newtcolorbox[auto counter]{promptbox}[2][]{enhanced,breakable,colback=black!3,colframe=black!60,boxrule=0.4pt,arc=1pt,left=3pt,right=3pt,top=2pt,bottom=2pt,title={Box~\thetcbcounter: #2},fonttitle=\bfseries\scriptsize,#1}
	\usepackage{pifont}
	\newcommand{\cmark}{\ding{51}}%
	\newcommand{\xmark}{\ding{55}}%
	\usepackage{array}
	\usepackage{adjustbox}
	\usepackage{twemojis}
	\usepackage{enumitem}
	\usepackage{xcolor}
	\newcommand{\gold}[1]{#1\rlap{\,{\scalebox{0.7}{\twemoji{1f947}}}}}
	\newcommand{\silv}[1]{#1\rlap{\,{\scalebox{0.7}{\twemoji{1f948}}}}}
	\newcommand{\brnz}[1]{#1\rlap{\,{\scalebox{0.7}{\twemoji{1f949}}}}}
	\newcommand{\impv}[1]{\textcolor{green!55!black}{\,(#1)}}
	\newcommand{\decl}[1]{\textcolor{red!75!black}{\,(#1)}}
	\newcommand{\nochg}[1]{\textcolor{gray}{\,(#1)}}
	\newcommand{\medalpad}{\phantom{\,{\scalebox{0.7}{\twemoji{1f947}}}}}
	\newcolumntype{M}{r<{\medalpad}}
	
	\title{MedVision: Benchmarking Quantitative Medical Image Analysis}

	\author{ \textbf{Yongcheng Yao\textsuperscript{$\heartsuit$}} \quad \textbf{Yongshuo Zong\textsuperscript{$\heartsuit$}} \quad \textbf{Raman Dutt\textsuperscript{$\heartsuit$}} \\
	\textbf{Yongxin Yang\textsuperscript{$\spadesuit$}} \quad \textbf{Sotirios A Tsaftaris\textsuperscript{$\diamondsuit$}} \quad \textbf{Timothy Hospedales\textsuperscript{$\heartsuit$}}\thanks{Corresponding author.} \\
	\textsuperscript{$\heartsuit$}School of Informatics, University of Edinburgh \\
	\textsuperscript{$\spadesuit$}Queen Mary University of London \quad \textsuperscript{$\diamondsuit$}School of Engineering, University of Edinburgh \\
	\small{\{yc.yao, yongshuo.zong, raman.dutt, s.tsaftaris, t.hospedales\}@ed.ac.uk}, \small{yongxin.yang@qmul.ac.uk} \\
	\\
	\small{\raisebox{-0.9em}{\includegraphics[height=2.5em]{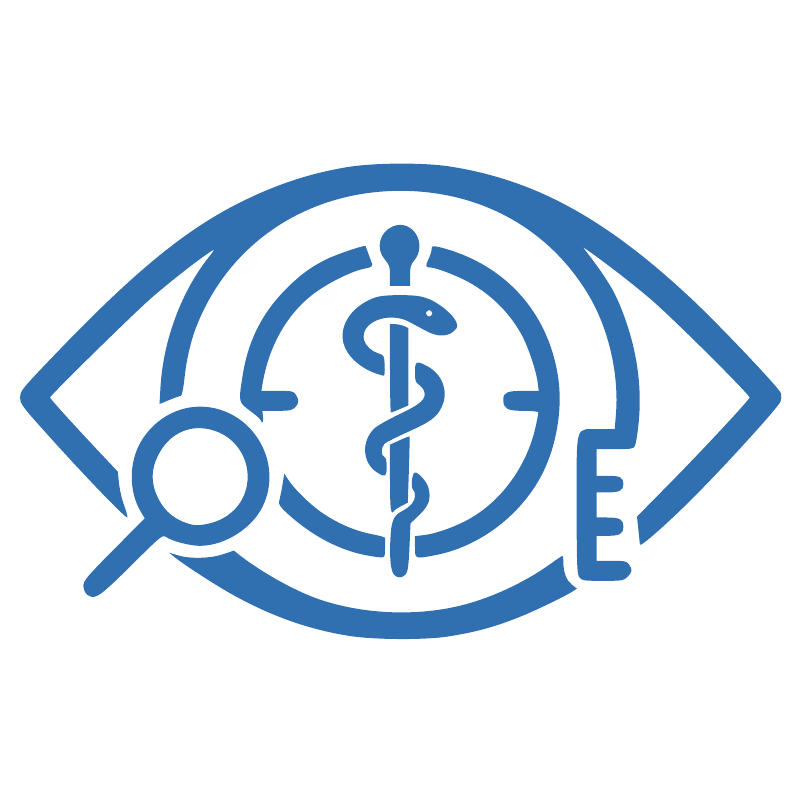}}~\url{https://medvision-vlm.github.io}}}

\begin{document}
	   \maketitle
	   \begin{abstract}
      Current vision-language models (VLMs) in medicine are primarily designed for categorical question answering (e.g., “Is this normal or abnormal?”) or qualitative descriptive tasks. However, clinical decision-making often relies on quantitative assessments, such as measuring the size of a tumor or the angle of a joint, from which clinicians draw their own diagnostic conclusions. This quantitative reasoning capability remains underexplored and poorly supported in existing VLMs. In this work, we introduce \textbf{MedVision}, a large-scale dataset and benchmark specifically designed to evaluate and improve VLMs on quantitative medical image analysis. MedVision spans 22 public datasets covering diverse anatomies and modalities, with \textbf{29.0K} 3D images, \textbf{11.2M} annotated 2D slices, and \textbf{24.3M} single-instance annotations. We focus on three representative quantitative tasks: (1) detection of anatomical structures and abnormalities, (2) tumor/lesion (T/L) size estimation, and (3) angle/distance (A/D) measurement. We show that current off-the-shelf VLMs perform poorly on these tasks. However, supervised and reinforcement fine-tuning (RFT) on MedVision significantly enhances performance across detection, T/L size estimation, and A/D measurement, yielding \textbf{MedVision-V0} as a strong open baseline. In the RFT stage, we design and evaluate the efficacy of process rewards, multiplicative reward composition, and multi-task RFT with curriculum learning. MedVision provides a foundation for developing VLMs with robust quantitative reasoning capabilities in medical imaging.
   \end{abstract}

   \section{Introduction}
   \label{sec:intro} Recent advances in vision-language models (VLMs)~\citep{LLaVAOneVision2024, Visual2023a, QwenVL2023, SelfSupervised2025} have made it possible to pair powerful visual encoders with flexible natural language reasoning, enabling a range of clinical applications such as image-level classification and interactive question answering. In practice, however, most medical VLM usage and evaluation have concentrated on categorical~\citep{Dataset2018a, Slake2021a, ben2019vqa} or qualitative outputs~\citep{MedICaT2020}: distinguishing “normal vs. abnormal” images, labeling a finding, or producing free-text descriptions of image content. These tasks are important, but they do not capture a critical part of routine clinical reasoning: the extraction of precise, \textit{quantitative} measurements from images that clinicians use to stage disease, plan interventions, and monitor therapy.

   Quantitative assessment is central to many medical decisions. Examples include detecting metastases and estimating tumor/lesion size for cancer staging~\citep{New2009, Proposed2024}, and measuring the Cobb angle for scoliosis diagnosis~\citep{Adolescent2015}. Despite this, the ability of modern VLMs to produce reliable, reproducible quantitative measurements (e.g., lengths, areas, volumes, angles) has received little systematic attention. Existing datasets~\citep{Dataset2018a, ben2019vqa, ben2021overview, Slake2021a, PathVQA2020, CheXpert2019a, CheXpert2024, MIMICCXR2019a, Explaining2022a, Making2022, PMCCLIP2023} emphasize classification accuracy or natural language quality rather than measurement precision. Recently, RadGPT~\citep{RadGPT2025a} presented a dataset of radiology reports with tumor location and size information, but it is limited to abdominal CT scans. Related benchmarks~\citep{GMAIMMBench2024, OmniMedVQA2024, Generalist2025, DrVDBench2025} similarly focus on qualitative tasks such as disease classification, visual question answering, and report generation. A recent work on reinforcement fine-tuning, MedVLM-R1~\citep{MedVLMR12026}, improves medical VLM accuracy on classification tasks, but leaves quantitative measurement largely unaddressed. A detailed comparison of existing datasets and benchmarks is provided in Appendix~\ref{appendix:subsec:dataset_benchmark_comparison}.

   Quantitative vision–language capability is challenging and underexplored for several reasons: measurement tasks demand precise localization across heterogeneous modalities and an understanding of geometric relationships and physical units rather than semantic categories alone; current architectures and training objectives rarely target calibrated numeric outputs; and public datasets with standardized measurement annotations remain sparse and fragmented across modalities and anatomies.

	To address these gaps, we introduce \textbf{MedVision}, a large-scale dataset and benchmark specifically created to assess and improve VLM performance on quantitative medical-image tasks. MedVision aggregates 22 public datasets spanning multiple anatomies and imaging modalities, comprising \textbf{29.0K} 3D images, \textbf{11.2M} annotated 2D slices, and \textbf{24.3M} single-instance annotations suitable for measurement tasks.\footnote{Dataset statistics for v1.0.0; the latest release is available at \url{https://medvision-vlm.github.io/explorer}.} We organize the benchmark around three representative, clinically relevant tasks: (1) detection of anatomical structures and abnormalities (localization and identification), (2) tumor/lesion (T/L) size estimation (bidirectional dimensions), and (3) angle/distance (A/D) measurement (e.g., joint angles, inter-structure distances).

   Using MedVision, we systematically evaluate open-weight VLMs in two settings: off-the-shelf inference and fine-tuning on the MedVision training splits. Off-the-shelf VLMs, even those with strong qualitative language abilities, perform poorly on quantitative tasks, failing at precise localization and producing large numeric errors, whereas supervised fine-tuning and reinforcement fine-tuning (SFT+RFT) on MedVision materially improve detection and measurement accuracy. We also analyze common failure modes and identify persistent challenges. Our \textbf{main contributions} are as follows:

   \noindent
   \textbf{\textit{(i)}} We highlight a clinically important but underappreciated gap: modern VLMs are not reliably able to produce precise quantitative measurements from medical images.
   
   \noindent
   \textbf{\textit{(ii)}} We introduce \textbf{MedVision}, a large-scale, multi-modality dataset for quantitative medical image analysis. 
   
   \noindent
   \textbf{\textit{(iii)}} We provide the first comprehensive evaluation of contemporary VLMs on detection, tumor/lesion size estimation, and angle/distance measurement. 
   
   \noindent
   \textbf{\textit{(iv)}} We train \textbf{MedVision-V0} on our dataset via supervised and reinforcement learning, which significantly outperforms all evaluated VLMs across all three quantitative tasks and generalizes to out-of-distribution imaging planes and anatomical targets.

   \noindent
   \textbf{\textit{(v)}} We design and evaluate the efficacy of process rewards, multiplicative reward composition in RFT, and multi-task RFT with curriculum learning.
   
   \noindent
   \textbf{\textit{(vi)}} We release the data, model checkpoints, and code to enable further research toward VLMs that support clinically useful quantitative reasoning.

   \section{MedVision: Dataset and Benchmark}
   \label{sec:methodology} 
\begin{figure*}[t]
      \centering
      \includegraphics[width=1\linewidth]{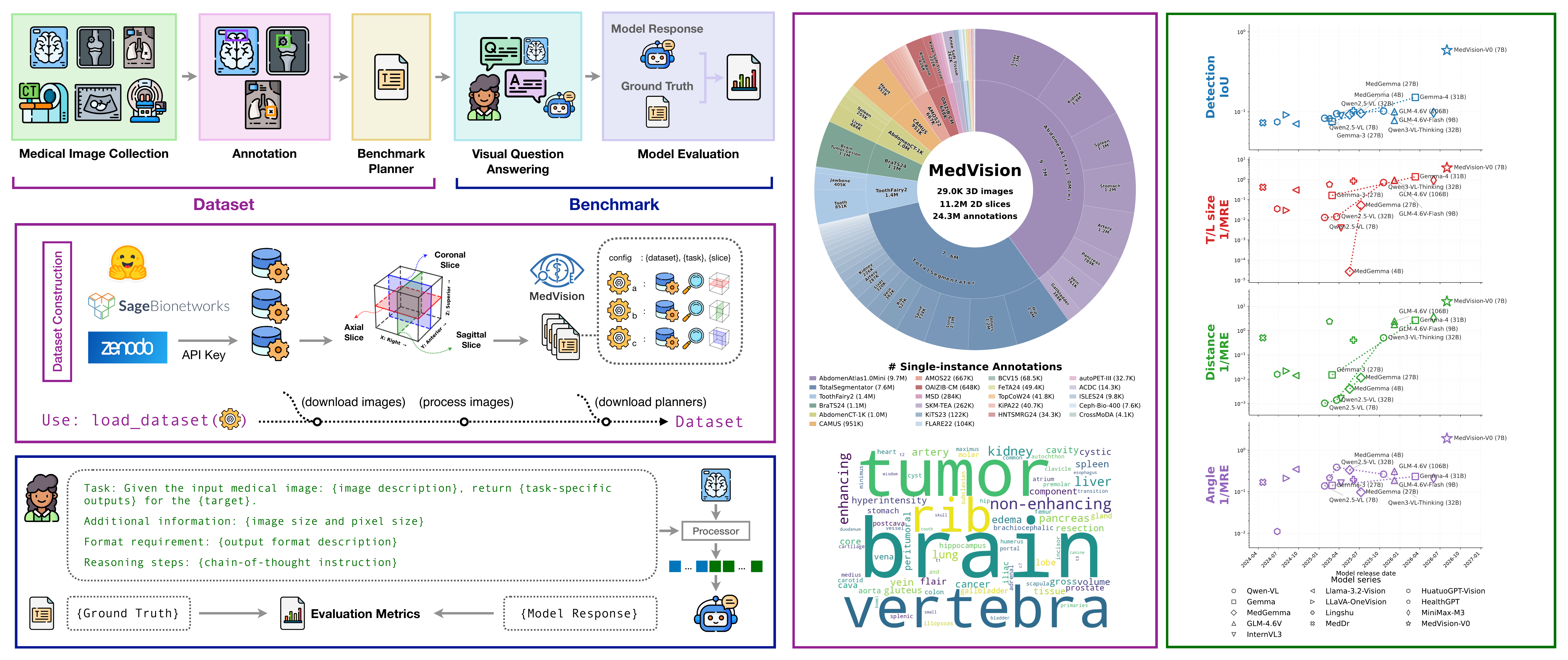}
      \caption{Overview of MedVision dataset construction and benchmark design (left), dataset summary (middle), and benchmark performance against model release date (right, details in Appendix~\ref{appendix:sec:per_model_profiles}). The MedVision dataset was constructed for out-of-the-box use: \texttt{load\_dataset} is all you need. Appendix~\ref{appendix:subsec:dataset_examples} shows examples of imaging data and quantitative annotation.}
      \label{fig:overview}
   \end{figure*}

An overview of our MedVision dataset and benchmark design is illustrated in Figure~\ref{fig:overview}. We curated a large-scale multi-anatomy and multi-modality medical image dataset with quantitative annotations (Section~\ref{subsec:dataset}). We designed a benchmark to evaluate VLMs' ability on three quantitative medical image analysis tasks (Section~\ref{subsec:benchmark_design}).

   \subsection{Dataset}
   \label{subsec:dataset} We constructed the MedVision dataset from public medical image data. Quantitative annotations are derived from the original labels. The dataset construction process is automated and reproducible, with a codebase that handles raw data downloading, image preprocessing, and annotation extraction.

   \noindent
   \textbf{Data Source \& Preprocessing.} Public medical imaging data with segmentation masks or landmark annotations were collected. We restricted the medical image modalities to those with physical spacing information in the image file header, which is essential for generating ground truth quantitative measurements. We collected 22 datasets, summarized in Table~\ref{tab:dataset}, covering a diverse range of anatomies and modalities: AbdomenAtlas~\citep{AbdomenAtlas2024a}, AbdomenCT-1K~\citep{AbdomenCT1K2022a}, ACDC~\citep{Deep2018c}, AMOS22~\citep{AMOS2022}, autoPET-III\footnote{https://autopet-iii.grand-challenge.org}, BCV15\footnote{https://doi.org/10.7303/syn3193805}, BraTS24~\citep{20242024b, Analysis2025, Brain2024b}, CAMUS~\citep{Deep2019b}, Ceph-Bio-400~\citep{Fully2016a}, CrossMoDA~\citep{CrossMoDA2023a}, FeTA24~\citep{Fetal2023}, FLARE22~\citep{Unleashing2024a}, HNTSMRG24\footnote{https://hntsmrg24.grand-challenge.org}, ISLES24\footnote{https://isles-24.grand-challenge.org}, KiPA22~\citep{Meta2021, Dense2020, Laparoscopic2011, Precise2012}, KiTS23~\citep{KiTS212023}, MSD~\citep{Medical2022h}, OAIZIB-CM~\citep{Automated2019a, CartiMorph2024, Quantifying2025}, SKM-TEA~\citep{SKMTEA2021}, ToothFairy2~\citep{Segmenting, Segmenting2025, Enhancing2024a}, TopCoW24~\citep{TopCoW2026}, and TotalSegmentator~\citep{TotalSegmentator2023}. Images are preprocessed to ensure consistent orientation and standardized format. Images are stored as 3D volumes with orientation corrected to RAS+ convention, where the first, second, and third dimensions correspond to the left-to-Right, posterior-to-Anterior, and inferior-to-Superior directions, respectively. Our data loading codebase supports flexible image slicing along any of the three anatomical planes (sagittal, coronal, and axial).

\begin{figure}[t]
      \centering
      \includegraphics[width=1\linewidth]{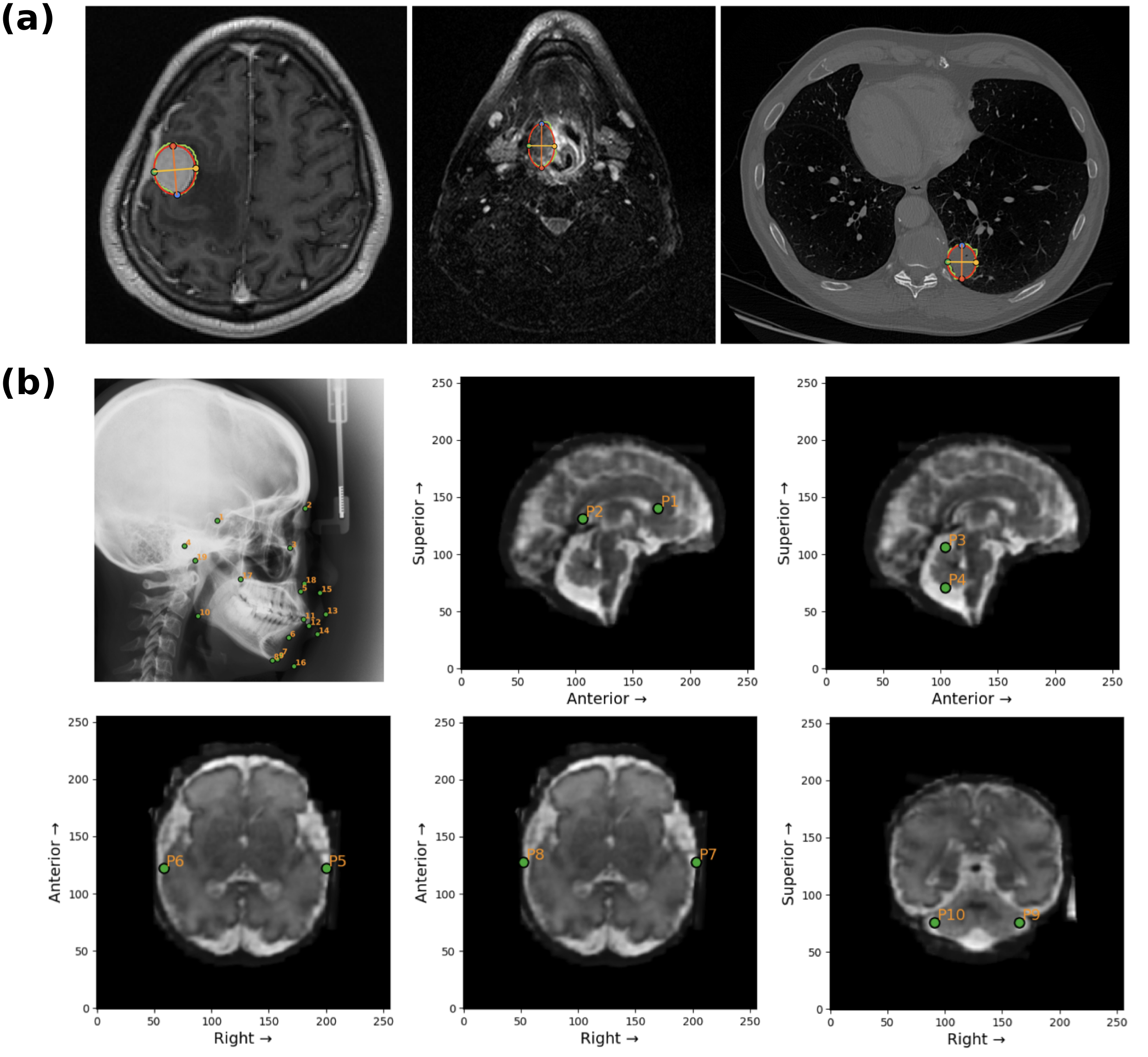}
      \caption{Quantitative annotation examples. (a) Tumor/lesion size annotation. An ellipse is fitted to the tumor/lesion mask. The physical lengths of the major and minor axes are recorded as T/L size annotation. (b) Landmarks in the Ceph-Bio-400 (top-left) and FeTA24 datasets. Ground truth angle and distance measurements are calculated from these landmarks.}
      \label{fig:tl-ad-annotation}
   \end{figure}

   \noindent
	\textbf{Quantitative Annotations.} We generated bounding box (Box), bidirectional tumor/lesion (T/L) size, and angle/distance (A/D) annotations to support the three benchmark tasks. The new annotations and metadata of images are released as the benchmark planner. \textbf{\textit{(i) Bounding Box Annotation:}} For each segmentation label in a 2D slice, a bounding box is fitted to each cluster of the binary mask. Clusters with fewer than 10 pixels in any dimension are excluded. \textbf{\textit{(ii) Tumor/Lesion Size Annotation:}} Bidirectional measurements of tumors/lesions from medical images are useful metrics and can provide quantitative information for disease assessment and monitoring. We generated T/L size annotations by calculating the lengths of the major and minor axes of the ellipse fitted to each T/L label in 2D slices (Figure~\ref{fig:tl-ad-annotation}a). Ellipse fitting is conducted in the real-world coordinate system by multiplying the image array indices with pixel sizes. \textbf{\textit{(iii) Angle/Distance Annotation:}} Angle and distance measurements are calculated from human-annotated landmarks. Figure~\ref{fig:tl-ad-annotation}b shows examples of landmarks in the two datasets. All target measurements are in \textbf{\textit{real-world units}} (e.g., mm), and models must convert image-space estimates into real-world units using the pixel/image size provided in the prompt.

   \begin{table}[t]
      \caption{Details of the MedVision dataset, with single-instance annotation counts at annotation version v1.0.0. MRI: Magnetic Resonance Imaging; CT: Computed Tomography; PET: Positron Emission Tomography; US: Ultrasound; Box: bounding box; T/L: tumor/lesion size; A/D: angle/distance. Full statistics, including multi-instance counts, in Table~\ref{appendix:tab:dataset} (Appendix~\ref{appendix:sec:dataset_stats}).}
      \label{tab:dataset}
      \centering
      \tiny
      \setlength{\tabcolsep}{4pt}
      \begin{adjustbox}
         {max width=\columnwidth}
         \begin{tabular}{lllrrr}
            \toprule \multirow{2}{*}{\textbf{Dataset}} & \multirow{2}{*}{\textbf{Anatomy}} & \multirow{2}{*}{\textbf{Modality}} & \multicolumn{3}{c}{\textbf{\# Single-instance Annotation}} \\
            \cmidrule(lr){4-6}                         &                                   &                                    & \textbf{Box} & \textbf{T/L} & \textbf{A/D} \\
            \midrule AbdomenAtlas     & abdomen       & CT      & 9,748,290 & 0      & 0 \\
            AbdomenCT-1K     & abdomen       & CT      & 1,041,588 & 0      & 0 \\
            ACDC             & heart         & MRI     & 14,271    & 0      & 0 \\
            AMOS22           & abdomen       & CT, MRI & 666,532   & 0      & 0 \\
            autoPET-III      & whole body    & CT, PET & 31,794    & 735    & 0 \\
            BCV15            & abdomen       & CT      & 68,543    & 0      & 0 \\
            BraTS24          & brain         & MRI     & 1,115,524 & 11,071 & 0 \\
            CAMUS            & heart         & US      & 951,370   & 0      & 0 \\
            Ceph-Bio-400     & head and neck & X-ray   & 0         & 0      & 7,600 \\
            CrossMoDA        & brain         & MRI     & 4,076     & 0      & 0 \\
            FeTA24           & fetal brain   & MRI     & 49,087    & 0      & 325 \\
            FLARE22          & abdomen       & CT      & 104,211   & 0      & 0 \\
            HNTSMRG24        & head and neck & MRI     & 32,029    & 1,392  & 0 \\
            ISLES24          & brain         & MRI     & 9,774     & 0      & 0 \\
            KiPA22           & kidney        & CT      & 37,647    & 3,095  & 0 \\
            KiTS23           & kidney        & CT      & 114,491   & 8,484  & 0 \\
            MSD              & multiple      & CT, MRI & 277,451   & 7,472  & 0 \\
            OAIZIB-CM        & knee          & MRI     & 648,048   & 0      & 0 \\
            SKM-TEA          & knee          & MRI     & 262,338   & 0      & 0 \\
            ToothFairy2      & tooth         & CT      & 1,413,979 & 0      & 0 \\
            TopCoW24         & brain         & CT, MRI & 41,829    & 0      & 0 \\
            TotalSegmentator & multiple      & CT, MRI & 7,603,455 & 0      & 0 \\
            \bottomrule
         \end{tabular}
      \end{adjustbox}
   \end{table}

   \subsection{Benchmark}
   \label{subsec:benchmark_design}

   \noindent
   \textbf{Model.} We evaluated a range of general and medical open-weight VLMs. General VLMs include Qwen2.5-VL~\citep{Qwen25VL2025a}, Qwen3-VL~\citep{qwen3technicalreport}, InternVL3~\citep{InternVL32025a}, Gemma3~\citep{Gemma2025}, Gemma4~\citep{Gemma2026}, GLM-4.6V~\citep{GLM45V2026}, MiniMax-M3~\citep{MiniMax2026}, Llama3.2-Vision, and LLaVA-OneVision~\citep{LLaVAOneVision2024}. Medical VLMs include Lingshu~\citep{Lingshu2025a}, MedGemma~\citep{MedGemma2026}, MedDr~\citep{Generalizable2026}, HuatuoGPT-Vision~\citep{Injecting2024a}, and HealthGPT-L14~\citep{HealthGPT2025a}. Details of these models are provided in Appendix~\ref{appendix:vlm_version}.

   \noindent
   \textbf{Prompt.} We focus on clinically relevant open-ended VQA tasks for our benchmark: (1) detection of healthy anatomical structures and abnormalities, (2) tumor/lesion size estimation, and (3) angle/distance measurement. Models are instructed to output the required estimates in a structured format. Generally, for all tasks, the input prompt includes a task description and a format requirement. Specifically, for T/L and A/D tasks, physical spacing information and chain-of-thought (CoT) instructions are added to provide necessary context and guide the model's reasoning process. \textbf{\textit{(i) Physical Spacing Information:}} A key feature of the benchmark is the careful handling of physical spacing information in the text prompt. Since each VLM adopts a different image processing strategy (see Appendix~\ref{appendix:subsec:vlm_image_processing_strategy}), we design a dedicated prompt-building pipeline to ensure the physical spacing values provided in the prompt match the actual image input, and the image size and pixel size together always reflect the physical dimensions of the target region. \textbf{\textit{(ii) CoT Instruction:}} The CoT instruction for T/L and A/D tasks provides step-by-step guidance for specific measurement tasks. In general, it instructs the model to first estimate landmark coordinates, then calculate the target distance or angle based on the predicted coordinates and the provided physical spacing information. The full CoT instructions are provided in Appendix~\ref{appendix:sec:sft_cot}.

   \noindent
   \textbf{Metric.} Model responses are parsed to extract the required numerical values. Parsing success is determined by whether the required number of numerical values can be extracted from the response: a rule-based parser reads the \verb|<answer>|\verb|</answer>| tags, and an \textbf{\textit{LLM-assisted parsing}} pipeline additionally recovers answers stated outside this format, without altering any rule-parsed value (Appendix~\ref{appendix:sec:llm_parsing}). All reported metrics are computed from the LLM-parsed responses. Success rate (SR) is calculated as the proportion of successfully parsed outputs among all samples. Other metrics are computed over successfully parsed outputs only, except the threshold metrics \verb|<metric>|$_{<k}$ and \verb|<metric>|$_{>k}$, whose denominator is the total number of samples so that they reflect both instruction following and measurement accuracy. For detection tasks, we report the region-based recall, precision, F1 score, intersection over union (IoU), IoU$_{>0.5}$, and IoU$_{0.5:0.05:0.95}$. For T/L size and A/D measurement tasks, we report the mean absolute error (MAE), mean relative error (MRE), and MRE$_{<0.1}$. Metric definitions are provided in Appendix~\ref{appendix:subsec:metric_def}.

   \noindent
   \textbf{Experimental Setting.} Details on computational resources, software environment, and code release are provided in Appendix~\ref{appendix:sec:compute_software}.

   \section{MedVision-V0 Training}
   \label{sec:fine-tuning} We established the MedVision benchmark to evaluate VLMs' capabilities on quantitative medical image analysis tasks. Our results reveal that pre-trained models struggle with medical image quantification. This performance gap motivates the need for post-training on our dataset. We aim to train a model capable of anatomical structure recognition, precise localization, tumor/lesion size estimation, and measurement via internal perception and reasoning -- an end-to-end VLM that does not rely on external tools or specialist models. To achieve this, we adopted a two-stage fine-tuning strategy, which includes supervised fine-tuning (SFT) and reinforcement fine-tuning (RFT). The SFT stage focuses on teaching the model to learn the answer formats and reasoning patterns, while the RFT stage further refines the model's outputs based on task-specific reward functions that encourage accurate intermediate estimates (e.g., landmark coordinates) and final measurements. We denote our model as \textbf{MedVision-V0}.

   \noindent
   \textbf{SFT with CoT.} At this stage, we curated a multi-task CoT dataset with 121K samples from the MedVision training set. We randomly select 110K detection, 5.5K T/L size estimation, and 5.5K A/D measurement cases from a subset of axial slices. We deliberately leave out coronal and sagittal slices to evaluate the model's generalization ability to out-of-distribution (OOD) data. A weighted random sampler is used during training to allow oversampling of minority-task cases to mitigate the data imbalance issue (Appendix~\ref{appendix:sec:temperature_sampler}). Each sample includes an image reshaped to $512\times512$ and a prompt-answer pair. The prompt follows the same format as the one used in the benchmark. The answer is structured into an internal reasoning part wrapped in \verb|<think>|\verb|</think>| tags and a final answer in \verb|<answer>|\verb|</answer>| tags. The reasoning text is generated by filling in intermediate ground truth values (e.g., landmark coordinates) into the task-specific CoT instruction template. The prompt structure and the CoT instructions and answer templates of all tasks are provided in Appendix~\ref{appendix:sec:sft_cot}.

   \noindent
   \textbf{RFT via GRPO.} For RFT, we used the GRPO algorithm~\citep{DeepSeekMath2024a} implemented in verl~\citep{HybridFlow2025}. We constructed an RFT dataset for each of the three tasks and used them sequentially (A/D $\rightarrow$ T/L $\rightarrow$ detection) for multi-step fine-tuning. An alternative \textbf{\textit{multi-task RFT}} approach is discussed in Section~\ref{subsec:sft_rft_effectiveness}. The same 121K training samples from the SFT stage are used; the image size and prompt remain the same, while the CoT answer is removed in RFT data as models learn from the reward signal. In addition to the standard format and answer rewards in GRPO, we designed \textbf{\textit{process rewards}} for T/L and A/D tasks to encourage accurate intermediate estimates. Both the process and answer rewards are calculated from $\exp(-x)$ where $x$ is the error of the model prediction. The final reward is calculated as follows: $r=r_{\text{format}}+r_{\text{process}} \cdot r_{\text{answer}}$. We ablate this \textbf{\textit{multiplicative reward composition}} against an additive alternative (Table~\ref{tab:rft_reward_ablation} in Appendix~\ref{appendix:medvision_v0_overview}). Appendix~\ref{appendix:sec:rft_grpo} details the reward computation and the training configuration.

   \begin{figure*}[t]
      \centering
      \includegraphics[width=0.9\linewidth]{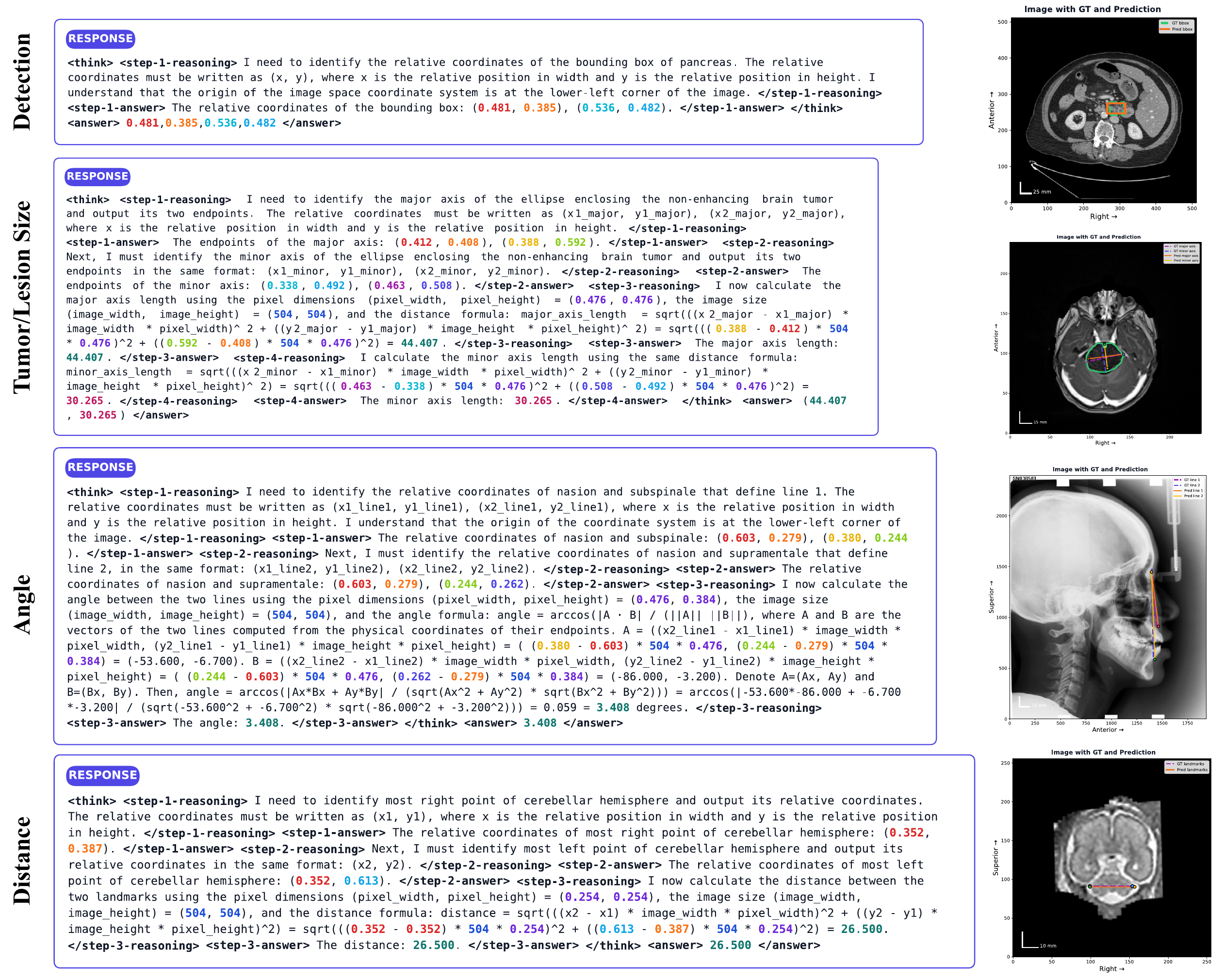}
      \caption{Model response text and the corresponding visualization of the predicted target and the ground truth, for the detection, tumor/lesion size, angle measurement, and distance measurement tasks (top to bottom).}
      \label{fig:model_reps_3tasks}
   \end{figure*}

   \section{Results \& Analysis}
   \label{sec:results} We first provide an overview of our fine-tuned model, MedVision-V0, and then quantitatively compare the performance of all evaluated models on the three benchmark task families.

   \subsection{MedVision-V0 Overview}

   \noindent
   \textbf{Qualitative Overview.} Figure~\ref{fig:model_reps_3tasks} shows example responses from MedVision-V0 on three tasks, each paired with the predicted target and the ground truth overlaid on the input image. The model is able to correctly identify the target region and produce reasonable estimates. The model's internal reasoning process, in the \verb|<think>|\verb|</think>| tags, shows that it first estimates the \textit{relative coordinates} of the target points, then computes the requested measurement from these coordinates together with the provided pixel size and image size. More examples of model responses in the detection, T/L and A/D tasks are provided in Appendix~\ref{appendix:medvision_v0_overview}.

   \noindent
   \textbf{Error Analysis.} To understand the challenge posed by MedVision, we analyze the end-to-end measurement error against the error of sub-tasks along the reasoning path: i) landmark localization, and ii) arithmetic mapping of landmark locations to final answers. We quantify the \textbf{\textit{localization error}} by the normalized L2 distance, $\|(x_{1}-x_{2},y_{1}-y_{2})\|_{2}/\sqrt{2}\in [0,1]$, where $(x_{1},y_{1}), (x_{2},y_{2})$ are relative coordinates. The model learns to use the standard formula to calculate a distance or angle. To analyze the \textbf{\textit{arithmetic error}}, we calculate the MRE between the model's prediction and the output of a Python execution of the same formula. The arithmetic error reflects the model's ability to perform the calculation, independent of the localization error. The final \textbf{\textit{measurement error}} is a coupling of the localization and arithmetic errors. Table~\ref{tab:error_analysis} shows that the localization of tumor/lesion landmarks is more challenging than that of anatomical landmarks in angle/distance tasks, since the former involves irregular and variable structures. Table~\ref{tab:error_analysis} also reveals that the arithmetic error is considerably higher in angle measurement than in distance measurement. To address this, we applied additional SFT with function calling to MedVision-V0, yielding a tool-use variant (MedVision-V0-tool). Although this eliminates the arithmetic error entirely, the final measurement error improves only marginally, indicating that \textbf{\textit{limited visual perception remains the bottleneck in quantitative medical image measurement tasks}}.

   \begin{table}[h]
      \caption{Error analysis of MedVision-V0 and a tooling variant, MedVision-V0-tool, on A/D and T/L tasks. Sample sizes are given in parentheses. normL2: normalized L2 distance; MRE: mean relative error; reported in \%.}
      \label{tab:error_analysis}
      \centering
      \tiny
      \setlength{\tabcolsep}{2pt}
      \begin{adjustbox}
         {max width=\columnwidth}
         \begin{tabular}{lrrr}
           \toprule \textbf{Model/Task} & \multicolumn{1}{c}{\textbf{Localization (normL2)}} & \multicolumn{1}{c}{\textbf{Arithmetic (MRE)}} & \multicolumn{1}{c}{\textbf{Measurement (MRE)}} \\
            \midrule \multicolumn{4}{l}{\textbf{MedVision-V0}} \\
            \quad \textbf{A/D}  & 1.6 & 12.8 & 27.7 \\
            \quad\quad Distance & 1.5 & 1.4  & 6.4  \\
            \quad\quad Angle      & 1.6 & 25.9 & 52.1 \\
            \quad \textbf{T/L}  & 6.0 & 3.4  & 26.0 \\
            \midrule \multicolumn{4}{l}{\textbf{MedVision-V0-tool}} \\
            \quad \textbf{A/D}  & 1.4 & 0.0  & 25.6 \\
            \quad\quad Distance & 1.4 & 0.0  & 6.7  \\
            \quad\quad Angle      & 1.3 & 0.0  & 47.4 \\
            \quad \textbf{T/L}  & 6.2 & 0.0  & 26.3 \\
            \bottomrule
         \end{tabular}
      \end{adjustbox}
   \end{table}

    \begin{figure*}[h]
      \centering
      \includegraphics[width=1\linewidth]{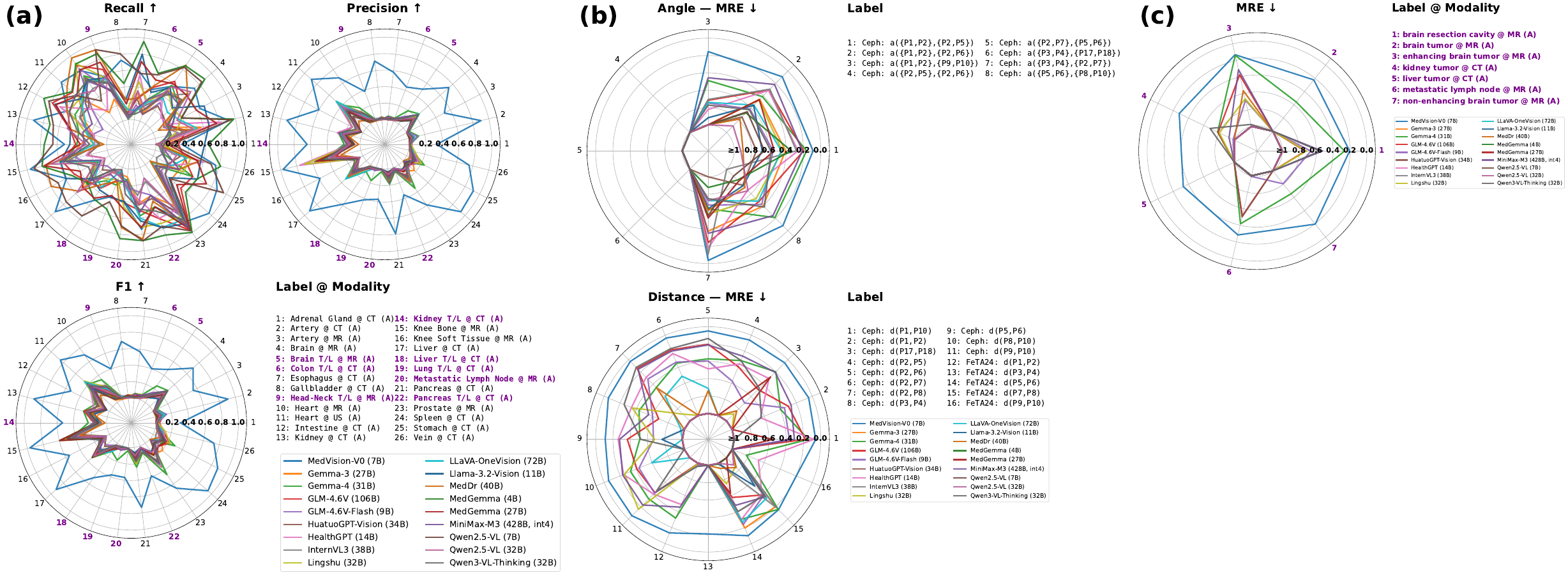}
      \caption{Performance of evaluated models on (a) detection, (b) angle/distance measurement, and (c) tumor/lesion size estimation tasks. The MRE axes are inverted so that the outermost ring corresponds to MRE $=0$ and the center to MRE $\geq 1$, i.e., \textbf{\textit{larger polygons always indicate better performance}}. In (b), MedVision-V0 attains near-zero angle MRE due to outliers; the distribution of its angle MRE is shown in Figure~\ref{fig:sft_rft_training_strategy}(b), where the median MRE for three of the angle measurements lies between 40\% and 60\%. Per-model breakdowns are provided in Figures~\ref{appendix:fig:radar_grid_part1} and~\ref{appendix:fig:radar_grid_part2} in Appendix~\ref{appendix:sec:per_model_profiles}.}
      \label{fig:radar_plots_3tasks}
   \end{figure*}

   \noindent
   \textbf{Performance.} Summary metrics for all evaluated models on all tasks are given in Section~\ref{subsec:benchmarking}. Here we provide a more detailed breakdown of MedVision-V0's performance on the detection (Table~\ref{tab:detection_per_label}) and T/L tasks (Table~\ref{tab:tl_size_medvision_labels}). The model performs better on anatomical structure detection than on tumor/lesion detection. The relative error of lesion size estimation ranges from 13\% to 33\%. Label-level detection performance and the effect of target size on detection are provided in Appendix~\ref{appendix:subsec:detection_performance}, which shows that \textbf{\textit{small-target and tumor/lesion detection are more challenging}}.

   \begin{table}[t]
      \caption{Detection performance of MedVision-V0, with the original labels categorized into 26 groups. Metrics are reported in \%. R: recall; P: precision; F1: F1 score; IoU: intersection over union; SR: success rate.}
      \label{tab:detection_per_label}
      \centering
      \tiny
      \setlength{\tabcolsep}{2pt}
      \begin{adjustbox}
         {max width=\columnwidth}
         \begin{tabular}{lrrrrrrr}
            \toprule \textbf{Label} & \multicolumn{1}{c}{\textbf{R} $\uparrow$} & \multicolumn{1}{c}{\textbf{P} $\uparrow$} & \multicolumn{1}{c}{\textbf{F1} $\uparrow$} & \multicolumn{1}{c}{\textbf{IoU} $\uparrow$} & \multicolumn{1}{c}{\textbf{SR} $\uparrow$} & \multicolumn{1}{c}{$\textbf{IoU}_{\textbf{\textgreater0.5}}$ $\uparrow$} & \multicolumn{1}{c}{$\textbf{IoU}_{\textbf{0.5:0.05:0.95}}$ $\uparrow$} \\
            \midrule \multicolumn{8}{l}{\textbf{\textit{Anatomy (18 regions, 13.4K samples)}}}    \\
            Knee Bone @ MR              & 96.3 & 97.2 & 96.7 & 94.3 & 100.0 & 99.2 & 94.2 \\
            Spleen @ CT                 & 96.1 & 93.3 & 94.3 & 90.4 & 100.0 & 98.0 & 86.8 \\
            Liver @ CT                  & 93.3 & 91.7 & 91.9 & 88.0 & 100.0 & 94.9 & 84.1 \\
            Kidney @ CT                 & 92.2 & 92.3 & 91.8 & 87.4 & 100.0 & 95.8 & 82.8 \\
            Artery @ CT                 & 91.8 & 90.3 & 90.8 & 85.3 & 100.0 & 97.3 & 78.1 \\
            Stomach @ CT                & 87.5 & 91.4 & 88.5 & 82.3 & 100.0 & 93.0 & 72.9 \\
            Heart @ US                  & 83.5 & 88.9 & 85.0 & 76.1 & 100.0 & 91.6 & 59.8 \\
            Vein @ CT                   & 75.3 & 75.4 & 73.3 & 65.6 & 100.0 & 76.2 & 51.9 \\
            Pancreas @ CT               & 74.0 & 79.3 & 73.2 & 62.6 & 100.0 & 74.5 & 41.2 \\
            Brain @ MR                  & 72.6 & 74.0 & 70.2 & 61.0 & 100.0 & 67.2 & 43.4 \\
            Heart @ MR                  & 78.7 & 68.1 & 70.2 & 60.7 & 100.0 & 67.5 & 43.0 \\
            Gallbladder @ CT            & 72.8 & 71.7 & 69.5 & 60.4 & 100.0 & 68.9 & 43.7 \\
            Prostate @ MR               & 84.4 & 59.1 & 66.3 & 55.2 & 100.0 & 62.7 & 32.3 \\
            Esophagus @ CT              & 58.1 & 58.1 & 56.9 & 45.1 & 100.0 & 44.1 & 21.6 \\
            Knee Soft Tissue @ MR       & 59.0 & 58.8 & 54.8 & 44.2 & 100.0 & 47.9 & 22.0 \\
            Artery @ MR                 & 63.9 & 51.7 & 54.8 & 44.9 & 100.0 & 51.5 & 25.1 \\
            Intestine @ CT              & 57.5 & 60.3 & 52.9 & 39.8 & 100.0 & 33.3 & 11.6 \\
            Adrenal Gland @ CT          & 31.5 & 33.2 & 29.8 & 20.2 & 100.0 & 10.2 & 2.4 \\
            \midrule \multicolumn{8}{l}{\textbf{\textit{Tumor/Lesion (8 regions, 8.5K samples)}}} \\
            Kidney Tumor/Lesion @ CT    & 51.2 & 64.4 & 53.4 & 43.9 & 100.0 & 47.7 & 25.5 \\
            Liver Tumor/Lesion @ CT     & 53.3 & 58.0 & 50.9 & 41.6 & 100.0 & 46.2 & 22.7 \\
            Brain Tumor/Lesion @ MR     & 55.4 & 51.3 & 48.6 & 39.5 & 100.0 & 42.0 & 21.5 \\
            Colon Tumor/Lesion @ CT     & 41.3 & 44.3 & 40.2 & 33.5 & 100.0 & 35.7 & 20.7 \\
            Head-Neck Tumor/Lesion @ MR & 57.6 & 38.7 & 40.1 & 33.0 & 100.0 & 34.4 & 20.4 \\
            Pancreas Tumor/Lesion @ CT  & 51.0 & 35.4 & 38.8 & 27.9 & 100.0 & 18.8 & 7.2 \\
            Lung Tumor/Lesion @ CT      & 37.8 & 42.0 & 37.5 & 30.6 & 100.0 & 33.4 & 16.9 \\
            Metastatic Lymph Node @ MR  & 41.4 & 37.0 & 36.7 & 30.4 & 100.0 & 33.5 & 17.8 \\
            \bottomrule
         \end{tabular}
      \end{adjustbox}
   \end{table}

   \begin{table}[t]
      \centering
      \caption{T/L size estimation performance of MedVision-V0 for each label. MRE, SR, and MRE$_{<0.1}$ are reported in \%, while MAE is in millimeters.}
      \label{tab:tl_size_medvision_labels}
      \tiny
      \setlength{\tabcolsep}{2pt}
      \begin{adjustbox}
         {max width=\columnwidth}
         \begin{tabular}{lrrrr}
            \toprule \textbf{Label} & \multicolumn{1}{c}{\textbf{MAE} $\downarrow$} & \multicolumn{1}{c}{\textbf{MRE} $\downarrow$} & \multicolumn{1}{c}{\textbf{SR} $\uparrow$} & \multicolumn{1}{c}{$\textbf{MRE}_{\textbf{\textless0.1}}$ $\uparrow$} \\
            \midrule enhancing brain tumor @ MR & 5.2  & 13.1 & 100.0 & 53.6 \\
            non-enhancing brain tumor @ MR      & 6.0  & 19.6 & 100.0 & 38.4 \\
            brain resection cavity @ MR         & 7.8  & 20.5 & 100.0 & 32.5 \\
            brain tumor @ MR                    & 6.2  & 22.3 & 100.0 & 38.4 \\
            kidney tumor @ CT                   & 12.5 & 27.5 & 100.0 & 18.4 \\
            metastatic lymph node @ MR          & 6.2  & 28.3 & 100.0 & 17.5 \\
            liver tumor @ CT                    & 16.4 & 33.1 & 100.0 & 19.5 \\
            \bottomrule
         \end{tabular}
      \end{adjustbox}
   \end{table}

   \subsection{Comparative Benchmarking Results}
   \label{subsec:benchmarking}

   \begin{table}[t]
      \caption{Model performance on detection tasks. Labels are categorized into anatomy and tumor/lesion groups. Metrics are reported in \%.}
      \label{tab:detection_performance}
      \centering
      \tiny
      \setlength{\tabcolsep}{2pt}
      \begin{adjustbox}
         {max width=\columnwidth}
         \begin{tabular}{lMMMMrMM}
            \toprule \textbf{Model} & \multicolumn{1}{c}{\textbf{R} $\uparrow$} & \multicolumn{1}{c}{\textbf{P} $\uparrow$} & \multicolumn{1}{c}{\textbf{F1} $\uparrow$} & \multicolumn{1}{c}{\textbf{IoU} $\uparrow$} & \multicolumn{1}{c}{\textbf{SR} $\uparrow$} & \multicolumn{1}{c}{$\textbf{IoU}_{\textbf{\textgreater0.5}}$ $\uparrow$} & \multicolumn{1}{c}{$\textbf{IoU}_{\textbf{0.5:0.05:0.95}}$ $\uparrow$} \\
            \midrule \multicolumn{8}{l}{\textbf{\textit{Anatomy (18 regions, 13.4K samples)}}}     \\
            MedVision-V0 (7B)       & \gold{81.3} & \gold{80.4} & \gold{79.1} & \gold{72.0} & 100.0 & \gold{80.1} & \gold{59.9} \\
            Gemma4 (31B)            & 34.8        & \silv{20.7} & \silv{22.6} & \silv{16.7} & 98.5  & \silv{13.5} & \silv{5.8} \\
            Lingshu (32B)           & 37.4        & \brnz{20.2} & \brnz{20.2} & \brnz{13.7} & 100.0 & 6.7 & 1.4 \\
            GLM-4.6V (106B)         & 46.5        & 16.4        & \brnz{20.2} & 13.5        & 99.9  & 6.0 & 1.8 \\
            Qwen3-VL-Thinking (32B) & 35.8        & 17.4        & 19.5        & 13.4        & 96.5  & \brnz{7.5} & \brnz{2.2} \\
            MedGemma (27B)          & 57.2        & 15.7        & 19.1        & 12.9        & 98.6  & 6.6 & 1.8 \\
            MedGemma (4B)           & \brnz{68.7} & 14.7        & 18.6        & 12.4        & 98.9  & 6.4 & 1.5 \\
            Qwen2.5-VL (32B)        & 44.8        & 14.9        & 18.4        & 12.5        & 100.0 & 6.3 & 1.8 \\
            LLaVA-OneVision (72B)   & 34.9        & 19.0        & 18.1        & 11.8        & 100.0 & 2.4 & 0.4 \\
            MiniMax-M3 (428B)       & 32.5        & 15.4        & 17.5        & 11.8        & 99.9  & 5.6 & 1.4 \\
            InternVL3 (38B)         & 31.1        & 17.0        & 17.2        & 11.5        & 100.0 & 5.3 & 1.2 \\
            Qwen2.5-VL (7B)         & \silv{69.7} & 12.2        & 16.7        & 11.3        & 99.4  & 5.6 & 1.7 \\
            HealthGPT-L14 (14B)     & 29.6        & \silv{20.7} & 16.0        & 10.2        & 98.7  & 1.8 & 0.4 \\
            GLM-4.6V-Flash (9B)     & 22.6        & 17.6        & 15.2        & 9.8         & 100.0 & 2.5 & 0.5 \\
            HuatuoGPT-Vision (34B)  & 25.6        & 17.7        & 15.1        & 9.8         & 99.5  & 2.7 & 0.6 \\
            Gemma3 (27B)            & 37.1        & 12.4        & 14.9        & 10.1        & 100.0 & 4.6 & 1.3 \\
            MedDr (40B)             & 53.6        & 11.1        & 14.6        & 9.7         & 96.8  & 4.2 & 1.1 \\
            Llama3.2-Vision (11B)   & 52.2        & 12.7        & 14.3        & 9.4         & 89.4  & 3.2 & 0.8 \\
            \midrule \multicolumn{8}{l}{\textbf{\textit{Tumor/Lesion (8 regions, 8.5K samples)}}} \\
            MedVision-V0 (7B)       & 52.4        & \gold{50.5} & \gold{46.9} & \gold{38.2} & 100.0 & \gold{40.7} & \gold{21.1} \\
            Gemma4 (31B)            & 43.6        & \silv{14.6} & \silv{18.1} & \silv{12.7} & 99.5  & \silv{8.4} & \silv{2.7} \\
            MiniMax-M3 (428B)       & 35.7        & \brnz{7.8}  & \brnz{10.2} & \brnz{6.5}  & 100.0 & \brnz{1.6} & \brnz{0.4} \\
            Lingshu (32B)           & 40.2        & 6.0         & 8.6         & 5.1         & 100.0 & 0.2 & 0.1 \\
            HealthGPT-L14 (14B)     & 29.2        & 7.0         & 8.4         & 5.1         & 97.8  & 0.5 & 0.1 \\
            LLaVA-OneVision (72B)   & 34.1        & 6.3         & 8.4         & 5.0         & 100.0 & 0.4 & 0.1 \\
            Qwen3-VL-Thinking (32B) & 33.9        & 6.6         & 8.2         & 5.2         & 98.0  & 1.3 & 0.3 \\
            InternVL3 (38B)         & 29.5        & 6.6         & 7.9         & 4.9         & 100.0 & 0.8 & 0.2 \\
            Qwen2.5-VL (32B)        & 38.5        & 5.7         & 7.7         & 4.7         & 100.0 & 0.6 & 0.2 \\
            GLM-4.6V (106B)         & 52.4        & 5.1         & 7.7         & 4.6         & 99.8  & 0.5 & 0.1 \\
            GLM-4.6V-Flash (9B)     & 26.2        & 6.1         & 7.5         & 4.7         & 100.0 & 1.0 & 0.2 \\
            MedGemma (27B)          & 54.1        & 4.6         & 7.5         & 4.3         & 97.0  & 0.1 & 0.0 \\
            MedGemma (4B)           & \gold{77.7} & 4.4         & 7.5         & 4.2         & 99.3  & 0.0 & 0.0 \\
            Qwen2.5-VL (7B)         & \silv{77.4} & 3.8         & 6.5         & 3.6         & 99.6  & 0.0 & 0.0 \\
            HuatuoGPT-Vision (34B)  & 21.6        & 5.0         & 6.4         & 3.8         & 99.0  & 0.3 & 0.1 \\
            MedDr (40B)             & \brnz{63.4} & 3.7         & 6.2         & 3.5         & 98.7  & 0.1 & 0.0 \\
            Gemma3 (27B)            & 34.3        & 4.3         & 6.1         & 3.6         & 100.0 & 0.3 & 0.0 \\
            Llama3.2-Vision (11B)   & 49.0        & 3.6         & 5.9         & 3.4         & 85.4  & 0.1 & 0.0 \\
            \bottomrule
         \end{tabular}
      \end{adjustbox}
   \end{table}

   \begin{table}[t]
      \centering
      \caption{Model performance on tumor/lesion size estimation tasks. MRE, SR, and MRE$_{<0.1}$ are reported in \%, while MAE is in millimeters.}
      \label{tab:tl_size_performance}
      \tiny
      \setlength{\tabcolsep}{2pt}
      \begin{adjustbox}
         {max width=\columnwidth}
         \begin{tabular}{lMMrM}
            \toprule \textbf{Model} & \multicolumn{1}{c}{\textbf{MAE} $\downarrow$} & \multicolumn{1}{c}{\textbf{MRE} $\downarrow$} & \multicolumn{1}{c}{\textbf{SR} $\uparrow$} & \multicolumn{1}{c}{$\textbf{MRE}_{\textbf{\textless0.1}}$ $\uparrow$} \\
            \midrule \multicolumn{5}{l}{\textbf{\textit{Tumor/Lesion (2K samples)}}} \\
            MedVision-V0 (7B)       & \gold{10.5} & \gold{26.0}  & 100.0 & \gold{23.5} \\
            Gemma4 (31B)            & \silv{21.7} & \silv{72.6}  & 98.9  & \silv{16.8} \\
            GLM-4.6V (106B)         & \brnz{31.4} & \brnz{107.3} & 92.2  & 4.1         \\
            MiniMax-M3 (428B)       & 32.7        & 108.1        & 95.8  & 5.6         \\
            GLM-4.6V-Flash (9B)     & 34.4        & 109.2        & 95.2  & 3.0         \\
            Lingshu (32B)           & 35.7        & 118.6        & 99.5  & 4.5         \\
            HealthGPT-L14 (14B)     & 51.8        & 176.0        & 100.0 & 2.2         \\
            Qwen3-VL-Thinking (32B) & 53.2        & 140.6        & 94.0  & \brnz{6.2}  \\
            MedDr (40B)             & 78.0        & 240.7        & 87.3  & 1.9         \\
            Llama3.2-Vision (11B)   & 101.2       & 329.4        & 98.3  & 0.7         \\
            Gemma3 (27B)            & 225.8       & 611.4        & 99.0  & 0.5         \\
            MedGemma (27B)          & 523.4       & 1905.0       & 57.0  & 0.5         \\
            HuatuoGPT-Vision (34B)  & 1018.6      & 2847.2       & 98.3  & 1.5         \\
            LLaVA-OneVision (72B)   & 1089.1      & 3368.0       & 100.0 & 1.2         \\
            Qwen2.5-VL (32B)        & 2054.6      & 6989.2       & 99.9  & 1.3         \\
            Qwen2.5-VL (7B)         & 2897.2      & 7680.9       & 95.4  & 0.7         \\
            InternVL3 (38B)         & 8606.8      & 25285.1      & 100.0 & 0.2         \\
            MedGemma (4B)           & 1109532.7   & 3733526.8    & 90.9  & 0.1         \\
            \bottomrule
         \end{tabular}
      \end{adjustbox}
   \end{table}

   \begin{table}[t]
      \centering
      \caption{Model performance on angle and distance measurement tasks. MRE, MRE$_{<0.1}$, and SR are reported in \%. MAEs are given in millimeters (distance) and degrees (angle).}
      \label{tab:ad_measurement_performance} \tiny
      \setlength{\tabcolsep}{2pt}
      \begin{adjustbox}
         {max width=\columnwidth}
         \begin{tabular}{lMMrM}
            \toprule \textbf{Model} & \multicolumn{1}{c}{\textbf{MAE} $\downarrow$} & \multicolumn{1}{c}{\textbf{MRE} $\downarrow$} & \multicolumn{1}{c}{\textbf{SR} $\uparrow$} & \multicolumn{1}{c}{$\textbf{MRE}_{\textbf{\textless0.1}}$ $\uparrow$} \\
            \midrule \multicolumn{5}{l}{\textbf{\textit{Distance (Ceph+FeTA, 1,100 samples)}}} \\
            MedVision-V0 (7B)       & \gold{3.6}  & \gold{6.4}   & 100.0 & \gold{81.4} \\
            MiniMax-M3 (428B)       & \silv{18.0} & \silv{30.4}  & 97.5  & \silv{22.0} \\
            GLM-4.6V (106B)         & \brnz{22.7} & 41.5         & 88.4  & 14.2        \\
            HealthGPT-L14 (14B)     & 23.1        & 42.4         & 97.6  & 18.0        \\
            Gemma4 (31B)            & 23.8        & \brnz{37.9}  & 99.8  & 10.8        \\
            GLM-4.6V-Flash (9B)     & 30.3        & 56.0         & 99.7  & 12.5        \\
            MedDr (40B)             & 106.3       & 197.6        & 91.2  & 8.5         \\
            Qwen3-VL-Thinking (32B) & 119.0       & 197.8        & 98.4  & 18.6        \\
            Lingshu (32B)           & 198.8       & 247.7        & 100.0 & \brnz{21.4} \\
            LLaVA-OneVision (72B)   & 2541.7      & 4557.2       & 100.0 & 6.1         \\
            HuatuoGPT-Vision (34B)  & 3921.6      & 6251.5       & 98.2  & 5.4         \\
            Llama3.2-Vision (11B)   & 4666.2      & 7065.8       & 98.6  & 2.1         \\
            Gemma3 (27B)            & 5054.2      & 6606.9       & 99.0  & 13.3        \\
            MedGemma (27B)          & 6239.0      & 8616.4       & 46.2  & 6.7         \\
            MedGemma (4B)           & 15222.0     & 24918.0      & 95.1  & 0.1         \\
            Qwen2.5-VL (32B)        & 56045.7     & 71734.7      & 99.9  & 6.3         \\
            InternVL3 (38B)         & 62066.9     & 62191.8      & 99.8  & 8.0         \\
            Qwen2.5-VL (7B)         & 63603.8     & 96542.0      & 98.3  & 0.5         \\
            \midrule \multicolumn{5}{l}{\textbf{\textit{Angle (Ceph, 960 samples)}}} \\
            MedVision-V0 (7B)       & \gold{4.7}  & \gold{52.1}  & 99.9  & \gold{52.0} \\
            MiniMax-M3 (428B)       & \silv{17.9} & 477.1        & 48.6  & 8.5         \\
            GLM-4.6V (106B)         & \brnz{19.7} & 322.7        & 88.8  & \silv{20.9} \\
            Gemma4 (31B)            & 23.9        & 428.0        & 88.5  & 11.7        \\
            HealthGPT-L14 (14B)     & 25.9        & 463.2        & 99.3  & 16.7        \\
            InternVL3 (38B)         & 30.3        & 616.8        & 100.0 & \brnz{20.4} \\
            Qwen3-VL-Thinking (32B) & 32.4        & 378.2        & 69.6  & 3.6         \\
            LLaVA-OneVision (72B)   & 33.6        & 473.5        & 100.0 & 2.9         \\
            Llama3.2-Vision (11B)   & 34.2        & \brnz{287.1} & 100.0 & 3.2         \\
            Lingshu (32B)           & 35.0        & 512.5        & 100.0 & 6.3         \\
            GLM-4.6V-Flash (9B)     & 35.0        & 531.7        & 99.1  & 8.1         \\
            MedGemma (4B)           & 35.8        & 298.8        & 96.7  & 6.0         \\
            Gemma3 (27B)            & 36.3        & 702.2        & 99.9  & 6.7         \\
            Qwen2.5-VL (32B)        & 41.2        & \silv{258.3} & 99.2  & 1.0         \\
            MedDr (40B)             & 45.7        & 591.5        & 94.5  & 5.3         \\
            MedGemma (27B)          & 46.4        & 1024.8       & 93.0  & 6.8         \\
            Qwen2.5-VL (7B)         & 48.0        & 724.9        & 97.6  & 2.0         \\
            HuatuoGPT-Vision (34B)  & 7070.0      & 9032.7       & 95.4  & 3.7         \\
            \bottomrule
         \end{tabular}
      \end{adjustbox}
   \end{table}

   We compare MedVision-V0 with 17 other VLMs on the three task families. The results are visualized in Figure~\ref{fig:radar_plots_3tasks} and detailed in Tables~\ref{tab:detection_performance}, \ref{tab:tl_size_performance}, and \ref{tab:ad_measurement_performance}; per-model profiles with per-sample metric distributions are given in Figures~\ref{appendix:fig:radar_grid_part1} and~\ref{appendix:fig:radar_grid_part2} in Appendix~\ref{appendix:sec:per_model_profiles}. MedVision-V0 shows superior performance across these tasks, demonstrating the effectiveness of our SFT-RFT training strategy. Among the off-the-shelf models, the strongest results come from the most recent general-purpose VLMs (Gemma4, GLM-4.6V, Qwen3-VL, and MiniMax-M3) rather than from the specialized medical VLMs (Figure~\ref{fig:overview}, right panel). Even so, every off-the-shelf model remains far behind MedVision-V0: all of them exceed $100\%$ MRE on angle measurement, and only Gemma4 falls below $100\%$ MRE on lesion size estimation. This reveals the limitation of both cutting-edge, general-purpose VLMs and specialized medical VLMs for quantitative medical image analysis, highlighting the value of our benchmark/dataset and the need for further research in this area.

   \subsection{Effectiveness of SFT-RFT Training}
   \label{subsec:sft_rft_effectiveness} We study the impact of our training pipeline by comparing our base model with SFT and SFT-RFT post-training on MedVision.

   \noindent
	   \textbf{RFT Strategy.} MedVision-V0 is trained with RFT on the three tasks sequentially (A/D $\rightarrow$ T/L $\rightarrow$ detection). To assess the effect of this schedule, we trained an ablation variant that starts from the same SFT model and undergoes a single \textbf{\textit{multi-task RFT}} stage with a curriculum learning scheme (Appendix~\ref{appendix:sec:curriculum}). The two variants achieve comparable performance (Table~\ref{tab:rft_strategy_ablation} in Appendix~\ref{appendix:medvision_v0_overview}), indicating that multi-task RFT constitutes a viable alternative to the multi-stage strategy. The two strategies nonetheless entail distinct optimization risks: sequential training is susceptible to catastrophic forgetting~\citep{Overcoming2017} of capabilities acquired in earlier stages, whereas multi-task training is subject to data imbalance and joint optimization challenges~\citep{MoDoMoDo2025a}. In our multi-task RFT, we address these challenges with \textbf{\textit{temperature-scaled task mixing}} (Appendix~\ref{appendix:sec:temperature_sampler}) and epoch-level \textbf{\textit{curriculum learning}} (Appendix~\ref{appendix:sec:curriculum}). We further ablate the reward design on the multi-task RFT model: the \textbf{\textit{multiplicative composition}} of process and answer rewards ($r_{\text{process}} \cdot r_{\text{answer}}$) matches an additive alternative on T/L size estimation and distance measurement, and performs better on angle measurement, whose reasoning trace is more sophisticated (Table~\ref{tab:rft_reward_ablation} in Appendix~\ref{appendix:medvision_v0_overview}).

	   \noindent
	   \textbf{In-distribution.} SFT dramatically improves detection precision, T/L size estimation, and A/D measurement accuracy over the base model; \textbf{\textit{the additional RFT step yields further consistent gains across all tasks}} (Table~\ref{tab:sft_rft_ablation} in Appendix~\ref{appendix:medvision_v0_overview}). SFT on MedVision also benefits other base models: Table~\ref{tab:sft_other_models} in Appendix~\ref{appendix:medvision_v0_overview} compares the base $\to$ SFT progression for Qwen2.5-VL (7B), Gemma4 (31B), and MedGemma (27B), all of which improve substantially across the three tasks.

	   \noindent
	   \textbf{Out-of-distribution Generalization.} Both SFT and SFT-RFT models generalize to unseen imaging planes (plane-OOD) and unseen anatomical targets (target-OOD), with \textbf{\textit{the SFT-RFT model (MedVision-V0) consistently outperforming the SFT-only model}} across detection and T/L size estimation (Table~\ref{tab:ood_generalization} in Appendix~\ref{appendix:medvision_v0_overview}). These settings probe generalization along multiple axes at once: even for an identical subject and target, an unseen imaging plane presents markedly different \textbf{\textit{anatomical context, target appearance, and physical extent}} (Figure~\ref{appendix:fig:planeOOD_tl} in Appendix~\ref{appendix:medvision_v0_overview}). Nonetheless, a performance gap relative to the in-distribution setting persists: anatomy detection F1 decreases from $79.1\%$ to $49.5\%$ (plane-OOD) and $36.0\%$ (target-OOD), T/L detection F1 from $46.9\%$ to $36.6\%$ and $45.5\%$, and T/L size MRE increases from $26.0\%$ to $36.4\%$ and $34.7\%$.

	   \begin{table}[tb]
	      \centering
	      \caption{Comparison of MedVision-V0 with BiomedParse on detection and T/L size estimation; BiomedParse's boxes and sizes are derived from its predicted segmentation masks. R: recall; P: precision. Metrics are reported in \%, except MAE (millimeters).}
	      \label{tab:biomedparse_comparison}
	      \tiny
	      \setlength{\tabcolsep}{2pt}
	      \begin{adjustbox}
		 {max width=\columnwidth}
		 \begin{tabular}{lrrrrrrr}
		    \toprule \textbf{Model} & \textbf{R} $\uparrow$ & \textbf{P} $\uparrow$ & \textbf{F1} $\uparrow$ & \textbf{IoU} $\uparrow$ & \textbf{SR} $\uparrow$ & $\textbf{IoU}_{\textbf{\textgreater0.5}}$ $\uparrow$ & $\textbf{IoU}_{\textbf{0.5:0.05:0.95}}$ $\uparrow$ \\
		    \midrule \multicolumn{8}{l}{\textbf{\textit{Detection: Anatomy}}}                     \\
		    MedVision-V0 (7B) & \textbf{81.3} & \textbf{80.4} & \textbf{79.1} & \textbf{72.0} & \textbf{100.0} & \textbf{80.1} & \textbf{59.9} \\
		    BiomedParse       & 53.9          & 57.4          & 53.8          & 50.2          & 64.8           & 55.3 & 44.5 \\
		    \midrule \multicolumn{8}{l}{\textbf{\textit{Detection: Tumor/Lesion}}}                \\
		    MedVision-V0 (7B) & \textbf{52.4} & \textbf{50.5} & \textbf{46.9} & \textbf{38.2} & \textbf{100.0} & \textbf{40.7} & \textbf{21.1} \\
		    BiomedParse       & 33.7          & 41.0          & 32.4          & 27.7          & 64.8           & 30.5 & 19.0 \\
		    \midrule\midrule 
		     \textbf{Model} & \textbf{MAE} $\downarrow$ & \textbf{MRE} $\downarrow$ & \textbf{SR} $\uparrow$ & $\textbf{MRE}_{\textbf{\textless0.1}}$ $\uparrow$ \\
		    \midrule \multicolumn{8}{l}{\textbf{\textit{T/L Size Estimation}}} \\
		     MedVision-V0 (7B)        & \textbf{10.5} & \textbf{26.0} & \textbf{100.0} & 23.5            \\
		    BiomedParse                       & 16.5          & 52.4          & 66.4           & \textbf{36.4}   \\
		    \bottomrule
		 \end{tabular}
	      \end{adjustbox}
	   \end{table}

	   \subsection{Comparison with a Specialist Model}
	   \label{subsec:comparison_with_specialist_model}

   We compare MedVision-V0 with BiomedParse~\citep{Foundation2025a}, a text-prompted biomedical foundation model for segmentation across diverse anatomical structures and pathologies. Unlike MedVision-V0, which directly predicts bounding box coordinates and lesion axis lengths end-to-end, BiomedParse produces pixel-level segmentation masks and relies on post-processing to derive the final detection and size estimation results. Bounding box and ellipse fitting are applied to convert BiomedParse's segmentation outputs into the target estimates. As shown in Table~\ref{tab:biomedparse_comparison}, MedVision-V0 substantially outperforms BiomedParse on both tasks. 

   We position MedVision-V0 as a generalist model for detection and measurement across anatomies and modalities from language prompts. MedVision-V0 can serve as a measurement tool within larger agentic systems, where an orchestrator (e.g., a radiology report generation model) invokes it on demand. Fine-tuning a segmentation specialist such as BiomedParse on MedVision would primarily advance pixel-level segmentation, which deviates from our goal: \textbf{\textit{assessing and improving VLMs' perception and reasoning ability for end-to-end (image-text-to-text) quantitative medical image analysis}}.

   \subsection{Clinical Decision Agreement}
   \label{subsec:cda} Absolute error metrics do not reveal whether a measurement error is large enough to alter the resulting clinical decision. We therefore complement the benchmark with a clinical decision agreement (CDA) analysis: each predicted measurement is mapped to a clinical category through a published cutoff table (\textbf{\textit{maxillary and mandibular position classification}} from cephalometric angles~\citep{Cephalometrics1953}; \textbf{\textit{kidney tumor staging}} from AJCC size cutoffs~\citep{AJCC2017}), and the predicted category is compared with the category implied by the ground truth measurement, so that any disagreement is attributable to measurement error alone. Agreement is quantified with chance-corrected, quadratic-weighted Cohen's $\kappa_w$. Off-the-shelf VLMs reach at most fair agreement~\citep{Measurement1977} on every CDA task ($\kappa_w\leq0.361$), whereas MedVision-V0 attains fair agreement on maxillary position ($\kappa_w=0.332$), moderate agreement on mandibular position ($\kappa_w=0.439$), and substantial agreement on kidney tumor staging ($\kappa_w=0.726$). The measurement gains from post-training thus translate into clinically meaningful decision agreement, a property that absolute error metrics alone do not capture. The complete protocol and per-model results are provided in Appendix~\ref{appendix:sec:cda}.

   \section{Conclusion}
   \label{sec:conclusion}

   We introduce \textbf{MedVision}, a large-scale, multi-modality, multi-anatomy medical imaging dataset with rich quantitative annotations for detection and measurement tasks. Using MedVision, we establish a comprehensive benchmark to evaluate VLMs on detection, tumor/lesion size estimation, and angle/distance measurement. Our results show that off-the-shelf VLMs perform poorly on these quantitative tasks, whereas supervised and reinforcement fine-tuning (SFT-RFT) with MedVision substantially improves accuracy. Moreover, RFT brings additional gains in both performance and generalization to out-of-distribution imaging planes and anatomical targets. Nonetheless, challenges remain, particularly in detecting and measuring small structures and highly variable tumor/lesion targets. We expect MedVision to provide a valuable foundation for advancing quantitative capabilities in medical VLMs.

   \section*{Acknowledgments}
   This work was supported by UK Research and Innovation (grant EP/S02431X/1) through the UKRI Centre for Doctoral Training in Biomedical AI at the School of Informatics, University of Edinburgh. We also acknowledge the support of the Edinburgh International Data Facility (EIDF) and the Data-Driven Innovation Programme at the University of Edinburgh.

   \section*{Limitations}
   MedVision is currently generalist in terms of being able to do a wide variety of quantitative tasks with a single VLM, without relying on external tools. However, it is specialist in that it is not simultaneously trained to do all the categorical and qualitative description tasks of mainstream or biomedically focused VLMs. Unifying MedVision-style quantitative analysis datasets and training recipes with categorical and qualitative datasets and training pipelines remains an important open challenge.

   \textbf{Ethical Considerations.} The benchmark reveals that current state-of-the-art VLMs are incapable of accurate medical image detection and measurement. The fine-tuned model MedVision-V0, although demonstrating improved performance, remains far from the accuracy and robustness required for clinical application. Such models must not be used to drive any medical diagnosis or clinical decision-making. Details on the intended use of the model and on data privacy are provided in Appendix~\ref{appendix:sec:intended_use_privacy}.

   \bibliography{MedVision}

   \appendix

   \section{Dataset and Benchmark Comparison}
   \label{appendix:subsec:dataset_benchmark_comparison} Early medical image–text dataset construction mainly targeted visual question answering (VQA), including VQA-RAD~\citep{Dataset2018a}, VQA-Med~\citep{ben2019vqa, ben2021overview}, SLAKE~\citep{Slake2021a}, and PathVQA~\citep{PathVQA2020}. These resources support structured evaluation of medical image understanding via question–answer pairs, but the evaluations remain predominantly categorical. Large-scale image-text corpora such as CheXpert~\citep{CheXpert2019a}, CheXpert Plus~\citep{CheXpert2024}, MIMIC-CXR~\citep{MIMICCXR2019a}, MIMIC-NLE~\citep{Explaining2022a}, MS-CXR~\citep{Making2022}, and PMC-OA~\citep{PMCCLIP2023} further support the development of VLMs by aligning medical images with clinical reports or natural language explanations from experts. A comparison of existing medical image–text datasets is summarized in Table~\ref{tab:medical_image_text_datasets}. Notably, these datasets lack quantitative annotations, limiting their utility for evaluating models on measurement tasks.

   Building on these resources, recent VLMs have been adapted to the medical domain, including LLaVA-Med~\citep{LLaVAmed2023}, MedDr~\citep{Generalizable2026}, HuatuoGPT-Vision~\citep{Injecting2024a}, HealthGPT-L14 (14B)~\citep{HealthGPT2025a}, MedGemma~\citep{MedGemma2026}, Lingshu~\citep{Lingshu2025a}, and MedVLM-R1~\citep{MedVLMR12026}. Several recent works have aimed to systematically evaluate medical VLMs, including GMAI-MMBench~\citep{GMAIMMBench2024}, OmniMedVQA~\citep{OmniMedVQA2024}, RadBench~\citep{Generalist2025}, and DrVD-Bench~\citep{DrVDBench2025}. These benchmarks emphasize descriptive or categorical tasks such as image understanding and report generation, with limited evaluation of quantitative measurement abilities.

   \begin{table}[h]
      \caption{Comparison of medical datasets. XR: X-ray; MR: Magnetic Resonance Imaging; CT: Computed Tomography; PET: Positron Emission Tomography; US: Ultrasound.}
      \label{tab:medical_image_text_datasets}
      \centering
      \tiny
      \setlength{\tabcolsep}{2pt}
      \begin{adjustbox}{max width=\columnwidth}
      \begin{tabular}{ccccc}
         \toprule \textbf{Dataset}                                                                      & \textbf{Size}                                              & \textbf{Anatomy} & \textbf{Modality}   & \textbf{Quantitative} \\
         \midrule \shortstack{VQA-RAD \\ \citep{Dataset2018a}}                                        & 3.5K                                                       & multiple         & XR, CT, MR          & \xmark \\
         \shortstack{VQA-Med \\ \citep{ben2019vqa, ben2021overview}}                            & 5K                                                         & multiple         & XR, CT, MR, US, PET & \xmark \\
         \shortstack{SLAKE \\ \citep{Slake2021a}} & 14K                                                        & multiple         & XR, CT, MR          & \xmark \\
         \shortstack{PathVQA \\ \citep{PathVQA2020}}                    & 33K                                                        & N/A              & pathology           & \xmark \\
         \shortstack{CheXpert \\ \citep{CheXpert2019a}}                                    & 224K                                                       & chest            & XR                  & \xmark \\
         \shortstack{CheXpert Plus \\ \citep{CheXpert2024}}          & 223K                                                       & chest            & XR                  & \xmark \\
         \shortstack{MIMIC-CXR \\ \citep{MIMICCXR2019a}}                                    & 377K                                                       & chest            & XR                  & \xmark \\
         \shortstack{MIMIC-NLE \\ \citep{Explaining2022a}}                                & 38K                                                        & chest            & XR                  & \xmark \\
         \shortstack{MS-CXR \\ \citep{Making2022}}                                     & 1.2K                                                       & chest            & XR                  & \xmark \\
         \shortstack{PMC-OA \\ \citep{PMCCLIP2023}}      & 1.6M                                                       & multiple         & diverse             & \xmark \\
         \midrule MedVision                                                                             & \shortstack{24.3M single-instance \\ 45.3M multi-instance} & multiple         & XR, CT, MR, US, PET & \cmark \\
         \bottomrule
      \end{tabular}
      \end{adjustbox}
   \end{table}

   \section{Dataset Examples}
   \label{appendix:subsec:dataset_examples} 
\begin{figure*}[t]
      \centering
      \includegraphics[width=1\linewidth]{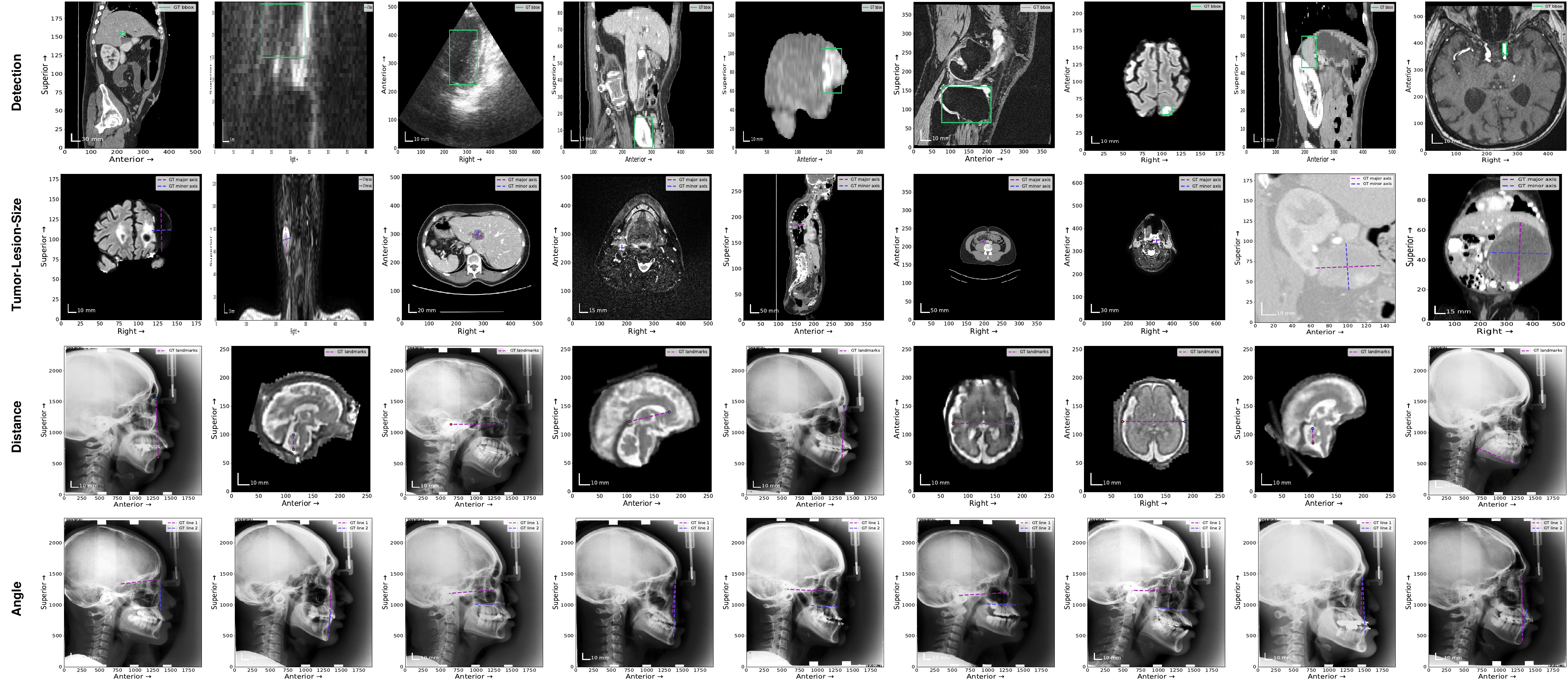}
      \caption{Preprocessed images and annotations from the MedVision dataset.}
      \label{appendix:fig:data_samples}
   \end{figure*}

Figure~\ref{appendix:fig:data_samples} shows preprocessed images of each dataset and examples of three types of quantitative annotations.

   \section{Full Dataset Statistics}
   \label{appendix:sec:dataset_stats} 
\begin{figure*}[t]
      \centering
      \includegraphics[width=1\linewidth]{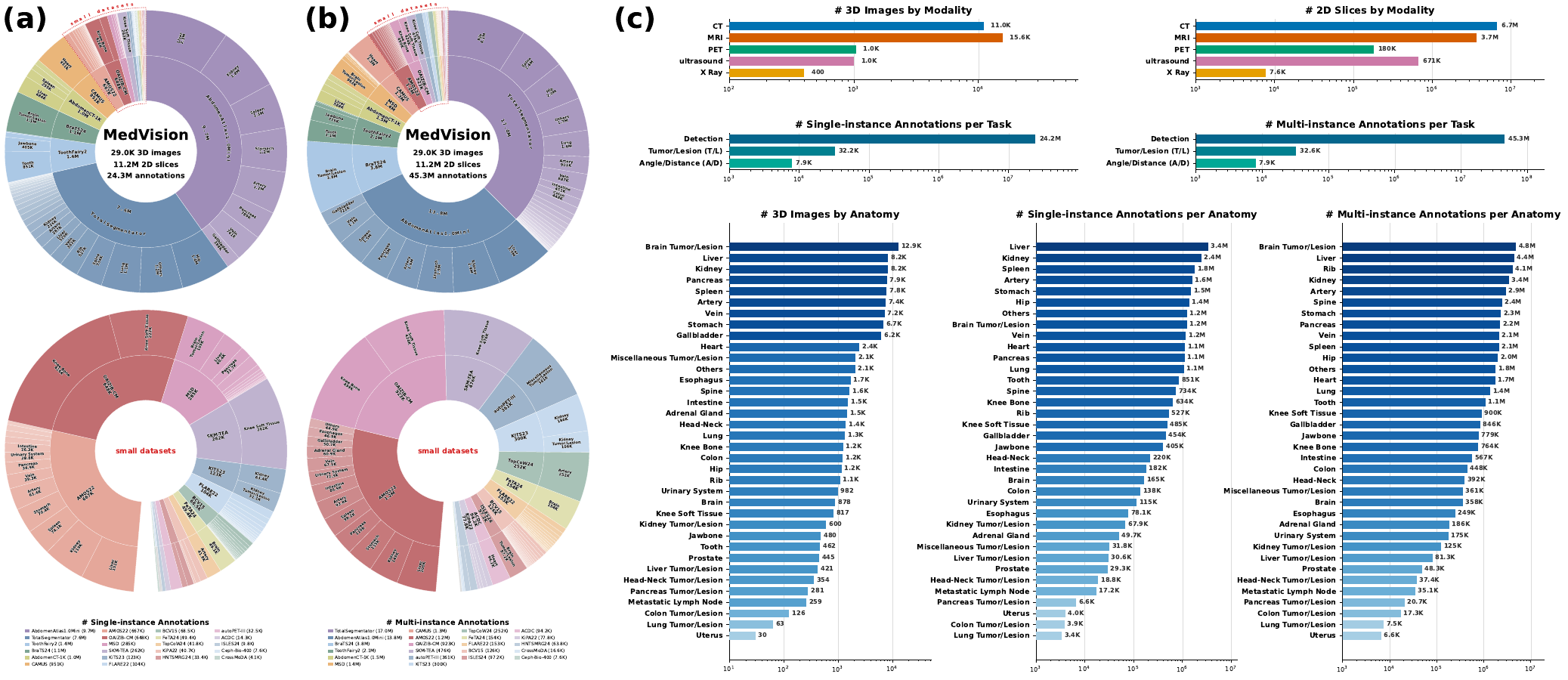}
      \caption{Composition of the MedVision dataset, computed on annotation version v1.0.0. (a) Single-instance and (b) multi-instance annotations broken down by dataset (inner ring) and anatomical subgroup (outer rings); the lower panel zooms into the smallest datasets. (c) Number of 3D images and 2D slices per modality, number of annotations per task, and number of 3D images and annotations per anatomy. The latest data release is available in the \href{https://medvision-vlm.github.io/explorer.html}{data explorer}.}
      \label{appendix:fig:dataset_stats}
   \end{figure*}

Figure~\ref{appendix:fig:dataset_stats} visualizes the composition across modalities, anatomies, and tasks, and Table~\ref{appendix:tab:dataset} provides the full statistics of the MedVision dataset v1.0.0, including the number of samples for each annotation type (bounding box, tumor/lesion size, and angle/distance). This detailed breakdown highlights the diversity and scale of the dataset, which supports comprehensive evaluation of VLMs on a wide range of quantitative medical imaging tasks.

   \begin{table*}
      [t]
      \caption{Full statistics of the MedVision dataset, computed on annotation version v1.0.0. Availability is marked next to the dataset name: $^{\diamondsuit}$ openly available; $^{\spadesuit}$ registration required; $^{\clubsuit}$ redistributed by us. Every dataset is partitioned into training and test sets at a 7:3 ratio. A benchmark sample is a (2D slice, target) pair. \textbf{Single-instance} annotations retain a target only when it appears as a single, sufficiently large instance on the slice; these are the annotations used for benchmarking and fine-tuning. \textbf{Multi-instance} annotations retain every annotated target regardless of how many instances it has. MRI: Magnetic Resonance Imaging; CT: Computed Tomography; PET: Positron Emission Tomography; US: Ultrasound; Box: bounding box; T/L: tumor/lesion size; A/D: angle/distance.}
      \label{appendix:tab:dataset}
      \centering
      \tiny
      \setlength{\tabcolsep}{12pt}
      \begin{adjustbox}{max width=\textwidth}
      \begin{tabular}{lllrrrrrrr}
         \toprule \multirow{2}{*}{\textbf{Dataset}} & \multirow{2}{*}{\textbf{Anatomy}} & \multirow{2}{*}{\textbf{Modality}} & \multirow{2}{*}{\textbf{\# 3D Image}} & \multicolumn{3}{c}{\textbf{\# Single-instance Annotation}} & \multicolumn{3}{c}{\textbf{\# Multi-instance Annotation}} \\
         \cmidrule(lr){5-7} \cmidrule(lr){8-10}     &               &         &                 & \textbf{Box}        & \textbf{T/L}    & \textbf{A/D}   & \textbf{Box}        & \textbf{T/L}    & \textbf{A/D} \\
         \cmidrule(lr){1-10} AbdomenAtlas$^{\diamondsuit}$              & abdomen       & CT      & 5,195           & 9,748,290           & 0               & 0              & 13,770,398          & 0               & 0 \\
         AbdomenCT-1K$^{\diamondsuit}$              & abdomen       & CT      & 1,000           & 1,041,588           & 0               & 0              & 1,549,325           & 0               & 0 \\
         ACDC$^{\diamondsuit\clubsuit}$             & heart         & MRI     & 300             & 14,271              & 0               & 0              & 94,160              & 0               & 0 \\
         AMOS22$^{\diamondsuit}$                    & abdomen       & CT, MRI & 360             & 666,532             & 0               & 0              & 1,215,776           & 0               & 0 \\
         autoPET-III$^{\diamondsuit\clubsuit}$      & whole body    & CT, PET & 2,076           & 31,794              & 735             & 0              & 360,638             & 749             & 0 \\
         BCV15$^{\diamondsuit\clubsuit}$            & abdomen       & CT      & 60              & 68,543              & 0               & 0              & 125,870             & 0               & 0 \\
         BraTS24$^{\diamondsuit\clubsuit}$          & brain         & MRI     & 10,632          & 1,115,524           & 11,071          & 0              & 3,767,594           & 11,093          & 0 \\
         CAMUS$^{\diamondsuit\clubsuit}$            & heart         & US      & 1,000           & 951,370             & 0               & 0              & 1,341,433           & 0               & 0 \\
         Ceph-Bio-400$^{\diamondsuit\clubsuit}$     & head and neck & X-ray   & 400             & 0                   & 0               & 7,600          & 0                   & 0               & 7,600 \\
         CrossMoDA$^{\diamondsuit\clubsuit}$        & brain         & MRI     & 105             & 4,076               & 0               & 0              & 16,623              & 0               & 0 \\
         FeTA24$^{\spadesuit}$                      & fetal brain   & MRI     & 80              & 49,087              & 0               & 325            & 153,599             & 0               & 325 \\
         FLARE22$^{\diamondsuit\clubsuit}$          & abdomen       & CT      & 50              & 104,211             & 0               & 0              & 152,954             & 0               & 0 \\
         HNTSMRG24$^{\diamondsuit}$                 & head and neck & MRI     & 300             & 32,029              & 1,392           & 0              & 62,424              & 1,416           & 0 \\
         ISLES24$^{\diamondsuit\clubsuit}$          & brain         & MRI     & 298             & 9,774               & 0               & 0              & 97,228              & 0               & 0 \\
         KiPA22$^{\diamondsuit\clubsuit}$           & kidney        & CT      & 70              & 37,647              & 3,095           & 0              & 74,690              & 3,095           & 0 \\
         KiTS23$^{\diamondsuit\clubsuit}$           & kidney        & CT      & 489             & 114,491             & 8,484           & 0              & 291,550             & 8,540           & 0 \\
         MSD$^{\diamondsuit}$                       & multiple      & CT, MRI & 3,225           & 277,451             & 7,472           & 0              & 1,438,472           & 7,674           & 0 \\
         OAIZIB-CM$^{\diamondsuit}$                 & knee          & MRI     & 507             & 648,048             & 0               & 0              & 922,989             & 0               & 0 \\
         SKM-TEA$^{\spadesuit}$                     & knee          & MRI     & 310             & 262,338             & 0               & 0              & 475,828             & 0               & 0 \\
         ToothFairy2$^{\spadesuit}$                 & tooth         & CT      & 480             & 1,413,979           & 0               & 0              & 2,131,223           & 0               & 0 \\
         TopCoW24$^{\diamondsuit\clubsuit}$         & brain         & CT, MRI & 250             & 41,829              & 0               & 0              & 251,901             & 0               & 0 \\
         TotalSegmentator$^{\diamondsuit\clubsuit}$ & multiple      & CT, MRI & 1,844           & 7,603,455           & 0               & 0              & 16,979,575          & 0               & 0 \\
         \cmidrule(lr){1-10} \textbf{Total (22)}                        &               &         & \textbf{29,031} & \textbf{24,236,327} & \textbf{32,249} & \textbf{7,925} & \textbf{45,274,250} & \textbf{32,567} & \textbf{7,925} \\
         \bottomrule
      \end{tabular}
      \end{adjustbox}
   \end{table*}

   \section{VLMs Version}
   \label{appendix:vlm_version} Table~\ref{appendix:tab:vlm_version} lists the HuggingFace model IDs of the evaluated open-weight VLMs.

   \begin{table*}
      [t]
      \caption{HuggingFace model ID of the evaluated open-weight VLMs.}
      \label{appendix:tab:vlm_version}
      \centering
      \tiny
      \begin{tabular}{ll}
         \toprule \textbf{Model} & \textbf{HuggingFace Model ID} \\
         \cmidrule(lr){1-2} MedVision-V0 (7B)       & \verb|YongchengYAO/MedVision-V0-7B| \\
         Qwen2.5-VL (7B)         & \verb|Qwen/Qwen2.5-VL-7B-Instruct| \\
         Qwen2.5-VL (32B)        & \verb|Qwen/Qwen2.5-VL-32B-Instruct| \\
         Qwen3-VL-Thinking (32B) & \verb|Qwen/Qwen3-VL-32B-Thinking| \\
         Lingshu (32B)           & \verb|lingshu-medical-mllm/Lingshu-32B| \\
         GLM-4.6V (106B)         & \verb|zai-org/GLM-4.6V| \\
         GLM-4.6V-Flash (9B)     & \verb|zai-org/GLM-4.6V-Flash| \\
         MiniMax-M3 (428B)       & \verb|cyankiwi/MiniMax-M3-AWQ-INT4| \\
         Gemma3 (27B)            & \verb|google/gemma-3-27b-it| \\
         Gemma4 (31B)            & \verb|google/gemma-4-31B-it| \\
         MedGemma (4B)           & \verb|google/medgemma-4b-it| \\
         MedGemma (27B)          & \verb|google/medgemma-27b-it| \\
         InternVL3 (38B)         & \verb|OpenGVLab/InternVL3-38B| \\
         Llama3.2-Vision (11B)   & \verb|meta-llama/Llama-3.2-11B-Vision-Instruct| \\
         LLaVA-OneVision (72B)   & \verb|llava-hf/llava-onevision-qwen2-72b-ov-hf| \\
         MedDr (40B)             & \verb|Sunanhe/MedDr_0401| \\
         HuatuoGPT-Vision (34B)  & \verb|FreedomIntelligence/HuatuoGPT-Vision-34B| \\
         HealthGPT-L14 (14B)     & \verb|lintw/HealthGPT-L14| \\
         \bottomrule
      \end{tabular}
   \end{table*}

   \section{VLM Image Processing Strategy}
   \label{appendix:subsec:vlm_image_processing_strategy} Different VLMs define their own image processing strategy in the image processor classes (or equivalent modules), where image processing pipelines including resizing, cropping, padding, and rescaling are defined. We summarized the image resizing strategies of the evaluated VLMs in Table~\ref{tab:vlm_image_resize}, upon which we designed appropriate pixel size adjustments for conveying physical spacing information to VLMs.

   \begin{table*}
      [t]
      \caption{Image resize strategies of the evaluated VLMs, i.e., how each model's own image processor transforms an input slice into the canvas its vision encoder perceives. ${\ddagger}$ aspect ratio preserved.}
      \label{tab:vlm_image_resize}
      \centering
      \tiny
      \setlength{\tabcolsep}{2pt}
      \begin{tabular}{lll}
         \toprule \textbf{Model} & \textbf{Perceived Image Size} & \textbf{Image Resize Strategy} \\
         \cmidrule(lr){1-3} MedVision-V0 (7B)       & dynamic & round each side to a multiple of 28; no pad \\
         Qwen2.5-VL (7B)         & dynamic & round each side to a multiple of 28; no pad \\
         Qwen2.5-VL (32B)        & dynamic & round each side to a multiple of 28; no pad \\
         Qwen3-VL-Thinking (32B) & dynamic & round each side to a multiple of 32; no pad \\
         Lingshu (32B)           & dynamic & round each side to a multiple of 28; no pad \\
         GLM-4.6V (106B)         & dynamic & round each side to a multiple of 28; no pad \\
         GLM-4.6V-Flash (9B)     & dynamic & round each side to a multiple of 28; no pad \\
         MiniMax-M3 (428B)       & dynamic & round each side to a multiple of 28 under a 451,584-px cap; no pad \\
         Gemma4 (31B)            & dynamic & scale to a 2,520-patch budget, then floor each side to a multiple of 48; no pad \\
         InternVL3 (38B)         & dynamic & stretch onto a dynamic grid of 448x448 tiles; no pad \\
         Llama3.2-Vision (11B)   & dynamic & scale$^{\ddagger}$ to fit within a canvas of 560x560 tiles, then pad to the tile grid \\
         LLaVA-OneVision (72B)   & dynamic & scale$^{\ddagger}$ to fit within a canvas of 384x384 tiles, then pad to the canvas \\
         Gemma3 (27B)            & 896x896 & stretch to a fixed size \\
         MedGemma (4B)           & 896x896 & stretch to a fixed size \\
         MedGemma (27B)          & 896x896 & stretch to a fixed size \\
         MedDr (40B)             & 448x448 & stretch to a fixed size (single tile, tiling off) \\
         HuatuoGPT-Vision (34B)  & 336x336 & pad to square, then CLIP-336 resize + center-crop \\
         HealthGPT-L14 (14B)     & 336x336 & pad to square, then CLIP-336 resize + center-crop \\
         \bottomrule
      \end{tabular}
   \end{table*}

   \section{LLM-Assisted Response Parsing}
   \label{appendix:sec:llm_parsing} The rule-based parser accepts only answers written inside the \verb|<answer>|\verb|</answer>| tags, so a response that states a correct measurement in any other form (e.g., \verb|\boxed{...}|, \enquote{Answer: ...}, or plain text) is scored as a parsing failure. This conflates two distinct failure modes: the inability to measure and the inability to follow the output format. Across the evaluated models, the rule-based parser rejects 7.8\% of detection, 18.3\% of T/L, and 21.0\% of A/D responses (Table~\ref{tab:llm_parsing_sr}), and most rejected responses contain a complete answer in a different wrapper. We therefore re-read every response with an LLM parser (Gemma4, 31B) whose only job is to locate the stated final answer, and compute all reported metrics from the re-parsed records.

   \noindent
   \textbf{Design.} The pipeline is built so that the LLM parser cannot corrupt a result. \textbf{\textit{(i)}} The parser sees only the response text, never the image or the ground truth, so it cannot favor or penalize any model. \textbf{\textit{(ii)}} It is not trusted to copy numbers: it must quote the sentence in which it found the answer, the quote is verified to occur in the response, and the numerical values are then re-extracted from the response text itself, so a value the model never wrote cannot be scored. \textbf{\textit{(iii)}} Recovery is strictly additive: a response parsed by the rule-based parser keeps its originally extracted values, and the LLM parser's output is used only for the responses the rule-based parser rejected. Each recovered record additionally stores the quoted sentence, so any recovered value can be audited against the original response. Algorithm~\ref{alg:llm_parsing} summarizes the parsing of one response.

\begin{algorithm*}[h]
   \caption{LLM-assisted parsing of one model response.}
   \label{alg:llm_parsing}
   \scriptsize
   \begin{algorithmic}[1]
      \Require response text $R$; required value count $k$ (detection: $k{=}4$, box coordinates; T/L: $k{=}2$, major and minor axis lengths; A/D: $k{=}1$, the measurement)
      \Statex
      \Process
      \State $V_r \gets \Call{RuleParse}{R}$ \Comment{read \texttt{<answer>} tags}
      \If{$V_r$ contains $k$ values}
         \State \Return ($V_r$, \textcolor{teal}{\textit{answer in expected format}})
      \EndIf
      \State $(\hat{V}, Q) \gets \Call{LlmLocate}{R}$ \Comment{LLM returns the stated values $\hat{V}$ and a verbatim quote $Q$ of the sentence stating them}
      \If{LLM reports no answer}
         \State \Return ($\varnothing$, \textcolor{orange}{\textit{no answer stated}})
      \EndIf
      \If{$Q$ occurs in $R$} \Comment{quote verification}
         \State $V \gets \Call{ReExtract}{R, Q, k}$ \Comment{$k$ numbers read from $R$ at $Q$, not from the LLM}
         \If{$V$ matches $\hat{V}$}
            \State \Return ($V$, \textcolor{olive}{\textit{answer in another format}})
         \EndIf
      \EndIf
      \State \Return ($\varnothing$, \textcolor{violet}{\textit{undetermined}})
      \Statex
      \Ensure the parsed values $V$, or $\varnothing$ when the response is unparsed, together with one of the four outcome labels: \textcolor{teal}{\textit{answer in expected format}}, \textcolor{olive}{\textit{answer in another format}}, \textcolor{orange}{\textit{no answer stated}}, or \textcolor{violet}{\textit{undetermined}}
   \end{algorithmic}
\end{algorithm*}

   \noindent
   \textbf{Outcomes and Effect.} Each response receives one of four outcomes: \textcolor{teal}{\textbf{\textit{answer in expected format}}} (rule-parsed), \textcolor{olive}{\textbf{\textit{answer in another format}}} (recovered and verified), \textcolor{orange}{\textbf{\textit{no answer stated}}}, or \textcolor{violet}{\textbf{\textit{undetermined}}} (the LLM parser output was unusable and the rule-based parser also failed). Only the first two outcomes count as successfully parsed. Pooled over the evaluated models, parsing success rises from $92.2\%$ to $98.6\%$ on detection, from $81.7\%$ to $94.5\%$ on T/L size estimation, and from $79.0\%$ to $93.9\%$ on A/D measurement (Table~\ref{tab:llm_parsing_sr}); the gap between the rule-parsed and LLM-parsed reports measures how much of a model's apparent failure was formatting rather than measurement.

   \begin{table}[h]
      \centering
      \caption{Success rate (SR, in \%) of the rule-based parser vs. the LLM-assisted parser (LLM), per model and task family. The last row pools all responses of each task family.}
      \label{tab:llm_parsing_sr} 
      \tiny
      \setlength{\tabcolsep}{2pt}
      \begin{adjustbox}
         {max width=\columnwidth}
         \begin{tabular}{lrrrrrr}
            \toprule  & \multicolumn{2}{c}{\textbf{Detection}} & \multicolumn{2}{c}{\textbf{T/L}} & \multicolumn{2}{c}{\textbf{A/D}} \\
            \cmidrule(lr){2-3} \cmidrule(lr){4-5} \cmidrule(lr){6-7}
            \textbf{Model} & \textbf{Rule} & \textbf{LLM} & \textbf{Rule} & \textbf{LLM} & \textbf{Rule} & \textbf{LLM} \\
            \midrule
            MedVision-V0 (7B)        & 100.0 & 100.0 & 100.0 & 100.0 & 100.0 & 100.0 \\
            Gemma3 (27B)             & 100.0 & 100.0 & 99.0 & 99.0 & 99.2 & 99.4 \\
            Gemma4 (31B)             & 98.6 & 98.9 & 98.9 & 98.9 & 94.6 & 94.6 \\
            GLM-4.6V (106B)          & 60.0 & 99.9 & 89.5 & 91.9 & 50.8 & 88.5 \\
            GLM-4.6V-Flash (9B)      & 99.9 & 100.0 & 91.2 & 94.8 & 96.6 & 99.4 \\
            HealthGPT-L14 (14B)      & 88.9 & 98.4 & 100.0 & 100.0 & 92.5 & 98.3 \\
            HuatuoGPT-Vision (34B)   & 77.7 & 99.2 & 97.8 & 98.4 & 95.1 & 96.9 \\
            InternVL3 (38B)          & 100.0 & 100.0 & 100.0 & 100.0 & 99.9 & 99.9 \\
            Lingshu (32B)            & 100.0 & 100.0 & 99.3 & 99.3 & 100.0 & 100.0 \\
            Llama3.2-Vision (11B)    & 46.5 & 87.8 & 14.5 & 97.9 & 46.3 & 99.2 \\
            LLaVA-OneVision (72B)    & 100.0 & 100.0 & 100.0 & 100.0 & 99.8 & 100.0 \\
            MedDr (40B)              & 97.2 & 97.6 & 53.7 & 87.3 & 66.2 & 92.7 \\
            MedGemma (27B)           & 96.0 & 98.0 & 53.5 & 57.5 & 44.0 & 68.0 \\
            MedGemma (4B)            & 98.4 & 98.9 & 81.1 & 90.8 & 93.2 & 95.8 \\
            MiniMax-M3 (428B)        & 99.8 & 100.0 & 85.9 & 95.9 & 54.2 & 74.7 \\
            Qwen2.5-VL (32B)         & 100.0 & 100.0 & 16.4 & 99.9 & 9.2 & 99.6 \\
            Qwen2.5-VL (7B)          & 99.4 & 99.5 & 95.9 & 95.9 & 98.0 & 98.0 \\
            Qwen3-VL-Thinking (32B)  & 96.3 & 97.2 & 93.2 & 93.5 & 83.0 & 84.9 \\
            \midrule All responses           & 92.2 & 98.6 & 81.7 & 94.5 & 79.0 & 93.9 \\
            \bottomrule
         \end{tabular}
      \end{adjustbox}
   \end{table}

   \section{Metric Definition}
   \label{appendix:subsec:metric_def}

   \noindent
   \textbf{Success Rate (SR).} For all tasks, success rate is defined as the proportion of model responses where the required numeric values can be parsed. SR indicates the model's instruction-following ability.

   \noindent
   \textbf{Detection Metrics.} For detection tasks, we calculated precision (P), recall (R), F1 score, and intersection over union (IoU) from the true positive (TP), false positive (FP), and false negative (FN) area. Given a ground truth bounding box $B_{y}$ and a predicted bounding box $B_{x}$, the TP, FP, and FN areas are defined as follows:
   \begin{align}
      \text{TP} & = \text{area}(B_{x}\cap B_{y}),  \\
      \text{FP} & = \text{area}(B_{x}- \text{TP}), \\
      \text{FN} & = \text{area}(B_{y}- \text{TP}).
   \end{align}
   Precision, recall, F1 score, and IoU are then calculated as:
   \begin{align}
      \text{Precision} & = \frac{\text{TP}}{\text{TP} + \text{FP}},                           \\
      \text{Recall}    & = \frac{\text{ TP}}{\text{TP} + \text{FN}},                          \\
      \text{F1}        & = \frac{2 \cdot \text{TP}}{2\cdot\text{TP} + \text{FP} + \text{FN}}, \\
      \text{IoU}       & = \frac{\text{TP}}{\text{TP} + \text{FP} + \text{FN}},
   \end{align}
   where the region-based F1 score is termed the Dice Similarity Coefficient (DSC) in image segmentation literature. We also reported IoU$_{>k}$, the proportion of samples with IoU greater than $k$, defined as:
   \begin{align}
      \text{IoU}_{>k} & = \frac{1}{N}\sum_{i=1}^{N}\mathbb{I}(\text{IoU}_{i}- k).
   \end{align}
   Analogous to the threshold-sweep convention in object detection benchmarks, IoU$_{0.5:0.05:0.95}$ averages IoU$_{>k}$ over 10 thresholds from 0.5 to 0.95 in steps of 0.05:
   \begin{align}
      \text{IoU}_{0.5:0.05:0.95} & = \frac{1}{10}\sum_{k \in \{0.5, 0.55, \ldots, 0.95\}}\text{IoU}_{>k}.
   \end{align}

   \noindent
   \textbf{Measurement Metrics.} For the tumor/lesion size and angle/distance measurement tasks, we calculated the mean absolute error (MAE) and mean relative error (MRE) between predictions $\mathbf{x}^{(n)}$ and ground truths $\mathbf{y}^{(n)}$ as follows:
   \begin{align}
      \text{MAE} & = \frac{1}{N}\sum_{i=1}^{N}\frac{1}{n}\sum_{j=1}^{n}|x_{i}^{j}- y_{i}^{j}|,             \\
      \text{MRE} & = \frac{1}{N}\sum_{i=1}^{N}\frac{1}{n}\sum_{j=1}^{n}(|x_{i}^{j}- y_{i}^{j}|/y_{i}^{j}),
   \end{align}
   where $N$ is the total number of samples, and $n$ is the number of values to be measured in each sample. For tumor/lesion size measurement, $n=2$ (major and minor axis lengths). For angle/distance measurement, $n=1$.

   Since the detection and measurement metrics are calculated only on the successfully parsed predictions, indicating the model prediction accuracy, we also report metrics that reflect both the instruction-following ability and prediction accuracy:
   \begin{align}
      \text{MRE}_{<k} & = \frac{1}{N}\sum_{i=1}^{N}\mathbb{I}(k - \frac{1}{n}\sum_{j=1}^{n}|x_{i}^{j}- y_{i}^{j}|/y_{i}^{j}),
   \end{align}
   where $\mathbb{I}(x)$ is the step function that returns 1 for $x\geq 0$. $\text{MRE}_{<k}$ indicates the proportion of samples with relative error less than $k$.

   \section{Computational Resources \& Software}
   \label{appendix:sec:compute_software}

   \noindent
   \textbf{Computational Resources.} Model evaluation used 4 NVIDIA H100 (80\,GB) GPUs; training of MedVision-V0 used 4 NVIDIA H200 (141\,GB) GPUs.

   \noindent
   \textbf{Software.} To support reproducibility, we publicly release all code used in this work. Dependency requirement files and Docker images are provided for every evaluated model. Reinforcement fine-tuning (RFT) was performed using verl v0.7.0.

   \section{Details of SFT with CoT}
   \label{appendix:sec:sft_cot} This appendix details the construction of the SFT-CoT training data (Section~\ref{sec:fine-tuning}): the prompt structure shared by all tasks, and the CoT instruction and answer template of each task.

   \noindent
   \textbf{Prompt Structure.} Each training sample pairs a $512\times512$ image with a prompt assembled from a fixed skeleton (Box~\ref{box:prompt_skeleton}): a task description naming the measurement target, additional information, a format requirement specifying the final answer format, and a task-specific CoT instruction listing the reasoning steps to follow. For the T/L size and A/D measurement tasks, the additional information provides the image size in pixels and the physical pixel size (pixel spacing rescaled after image resizing), which the reasoning steps use to convert relative coordinates into physical measurements; the detection prompt omits this part because its answer is expressed in relative coordinates. Placeholders are written in angle brackets.

\begin{promptbox}[label={box:prompt_skeleton}]{Prompt skeleton (all tasks)}
\begin{lstlisting}[style=promptstyle]
Task:
Given the input medical image[: <image description>], <task description>
Additional information:
The image size is <image_width> pixels (width) x <image_height> pixels (height).
The pixel size for this image is <pixel_width> <unit> (width) x <pixel_height> <unit> (height).
Format requirement:
<answer format requirement>
Reasoning steps:
<task-specific CoT instruction>
Follow the reasoning steps to get the final answer in the required format.
\end{lstlisting}
\end{promptbox}

   \noindent
   \textbf{CoT Answer Construction.} The SFT target is generated from the task's CoT answer template. Placeholders in the template are filled with ground truth values derived from the annotations: the relative coordinates of the bounding-box corners, ellipse-axis endpoints, or landmarks; the image and pixel sizes; and the intermediate and final measurement values. The resulting response walks through the declared reasoning steps, each wrapped in \verb|<step-k-reasoning>| and \verb|<step-k-answer>| tags inside the \verb|<think>| block, and ends with the final answer in \verb|<answer>| tags, so the supervised reasoning trace is consistent with the ground truth at every step. Boxes~\ref{box:cot_detection}--\ref{box:cot_angle} give the task description, format requirement, CoT instruction, and CoT answer template for the detection, T/L size estimation, distance measurement, and angle measurement tasks, respectively.

\begin{promptbox}[label={box:cot_detection}]{Detection: prompt components and CoT answer template}
{\footnotesize\textbf{Task description}}
\begin{lstlisting}[style=promptstyle]
return the coordinates of the lower-left and upper-right corner of the bounding box for the <label>.
\end{lstlisting}
{\footnotesize\textbf{Format requirement}}
\begin{lstlisting}[style=promptstyle]
The final answer must be enclosed within <answer> </answer> tags. The answer should be four decimal numbers separated by commas without any units or additional text. The first two numbers are the coordinates of the lower-left corner and the last two numbers are the coordinates of the upper-right corner of the bounding box. Use relative coordinates in the image space, where the origin is at the lower-left corner of the image. Relative coordinates should be values between 0 and 1, representing the relative positions in the image.
\end{lstlisting}
{\footnotesize\textbf{CoT instruction}}
\begin{lstlisting}[style=promptstyle]
Step 1: Identify the relative coordinates of the bounding box. The relative coordinates must be written as (x, y), where x is the relative position in width and y is the relative position in height.
Report the reasoning process and final answer within <think> </think> and <answer> </answer> tags, respectively. Inside <think> </think>, include reasoning and step results using <step-k-reasoning> </step-k-reasoning> and <step-k-answer> </step-k-answer> tags.
\end{lstlisting}
{\footnotesize\textbf{CoT answer template}}
\begin{lstlisting}[style=promptstyle]
<think>
<step-1-reasoning> I need to identify the relative coordinates of the bounding box of <label_name>. The relative coordinates must be written as (x, y), where x is the relative position in width and y is the relative position in height. I understand that the origin of the image space coordinate system is at the lower-left corner of the image. </step-1-reasoning>
<step-1-answer> The relative coordinates of the bounding box: (<coor0_w>, <coor0_h>), (<coor1_w>, <coor1_h>). </step-1-answer>
</think>
<answer> <coor0_w>,<coor0_h>,<coor1_w>,<coor1_h></answer>
\end{lstlisting}
\end{promptbox}

\begin{promptbox}[label={box:cot_tl}]{T/L size estimation: prompt components and CoT answer template}
{\footnotesize\textbf{Task description}}
\begin{lstlisting}[style=promptstyle]
estimate the major and minor axis lengths of the ellipse enclosing the <label>, in <unit>.
\end{lstlisting}
{\footnotesize\textbf{Format requirement}}
\begin{lstlisting}[style=promptstyle]
The final answer must be enclosed within <answer> </answer> tags. The answer should consist of two decimal numbers separated by a comma, without units or extra text. The first number is the major axis length, and the second is the minor axis length.
\end{lstlisting}
{\footnotesize\textbf{CoT instruction}}
\begin{lstlisting}[style=promptstyle]
Step 1: Identify the major axis (the longest diameter) of the ellipse enclosing the target region. Find its two endpoints and record their relative coordinates in the format (x, y) = (relative position in width direction, relative position in height direction). Denote the endpoints as (x1_major, y1_major) and (x2_major, y2_major).
Step 2: Identify the minor axis (the shortest diameter) of the ellipse. Find its two endpoints and record their relative coordinates in the same (x, y) format. Denote them as (x1_minor, y1_minor) and (x2_minor, y2_minor).
Step 3: Given the pixel dimensions (pixel_width, pixel_height) and image size (image_width, image_height), compute the physical length of the major axis using: major_axis_length = sqrt(((x2_major - x1_major) * image_width * pixel_width)^2 + ((y2_major - y1_major) * image_height * pixel_height)^2).
Step 4: Similarly, compute the physical length of the minor axis using: minor_axis_length = sqrt(((x2_minor - x1_minor) * image_width * pixel_width)^2 + ((y2_minor - y1_minor) * image_height * pixel_height)^2).
Report the reasoning process and final answer within <think> </think> and <answer> </answer> tags, respectively. Inside <think> </think>, include reasoning and step results using <step-k-reasoning> </step-k-reasoning> and <step-k-answer> </step-k-answer> tags.
\end{lstlisting}
{\footnotesize\textbf{CoT answer template}}
\begin{lstlisting}[style=promptstyle]
<think>
<step-1-reasoning> I need to identify the major axis of the ellipse enclosing the <label> and output its two endpoints. The relative coordinates must be written as (x1_major, y1_major), (x2_major, y2_major), where x is the relative position in width and y is the relative position in height. </step-1-reasoning>
<step-1-answer> The endpoints of the major axis: (<x1_major>, <y1_major>), (<x2_major>, <y2_major>). </step-1-answer>
<step-2-reasoning> Next, I must identify the minor axis of the ellipse enclosing the <label> and output its two endpoints in the same format: (x1_minor, y1_minor), (x2_minor, y2_minor). </step-2-reasoning>
<step-2-answer> The endpoints of the minor axis: (<x1_minor>, <y1_minor>), (<x2_minor>, <y2_minor>). </step-2-answer>
<step-3-reasoning> I now calculate the major axis length using the pixel dimensions (pixel_width, pixel_height) = (<pixel_width>, <pixel_height>), the image size (image_width, image_height) = (<image_width>, <image_height>), and the distance formula: major_axis_length = sqrt(((x2_major - x1_major) * image_width * pixel_width)^2 + ((y2_major - y1_major) * image_height * pixel_height)^2) = sqrt(((<x2_major> - <x1_major>) * <image_width> * <pixel_width>)^2 + ((<y2_major> - <y1_major>) * <image_height> * <pixel_height>)^2) = <major_axis_length>. </step-3-reasoning>
<step-3-answer> The major axis length: <major_axis_length>. </step-3-answer>
<step-4-reasoning> I calculate the minor axis length using the same distance formula: minor_axis_length = sqrt(((x2_minor - x1_minor) * image_width * pixel_width)^2 + ((y2_minor - y1_minor) * image_height * pixel_height)^2) = sqrt(((<x2_minor> - <x1_minor>) * <image_width> * <pixel_width>)^2 + ((<y2_minor> - <y1_minor>) * <image_height> * <pixel_height>)^2) = <minor_axis_length>. </step-4-reasoning>
<step-4-answer> The minor axis length: <minor_axis_length>. </step-4-answer>
</think>
<answer> (<major_axis_length>, <minor_axis_length>) </answer>
\end{lstlisting}
\end{promptbox}

\begin{promptbox}[label={box:cot_distance}]{Distance measurement: prompt components and CoT answer template}
{\footnotesize\textbf{Task description}}
\begin{lstlisting}[style=promptstyle]
estimate the distance of <measurement name> in <unit>, which is the distance between 2 landmark points: (landmark 1) <landmark 1>, (landmark 2) <landmark 2>.
\end{lstlisting}
{\footnotesize\textbf{Format requirement}}
\begin{lstlisting}[style=promptstyle]
The final answer must be enclosed within <answer> </answer> tags. The answer should be a single decimal number without units or extra text.
\end{lstlisting}
{\footnotesize\textbf{CoT instruction}}
\begin{lstlisting}[style=promptstyle]
Step 1: Identify the landmark 1 and record its relative coordinates in the format (x, y) = (relative position in width direction, relative position in height direction). Denote the coordinates as (x1, y1).
Step 2: Identify the landmark 2 and record its relative coordinates in the same (x, y) format. Denote the coordinates as (x2, y2).
Step 3: Given the pixel dimensions (pixel_width, pixel_height) and image size (image_width, image_height), compute the physical distance between the two landmarks using: distance = sqrt(((x2 - x1) * image_width * pixel_width)^2 + ((y2 - y1) * image_height * pixel_height)^2).
Report the reasoning process and final answer within <think> </think> and <answer> </answer> tags, respectively. Inside <think> </think>, include reasoning and step results using <step-k-reasoning> </step-k-reasoning> and <step-k-answer> </step-k-answer> tags.
\end{lstlisting}
{\footnotesize\textbf{CoT answer template}}
\begin{lstlisting}[style=promptstyle]
<think>
<step-1-reasoning> I need to identify <landmark 1> and output its relative coordinates. The relative coordinates must be written as (x1, y1), where x is the relative position in width and y is the relative position in height. </step-1-reasoning>
<step-1-answer> The relative coordinates of <landmark 1>: (<x1>, <y1>). </step-1-answer>
<step-2-reasoning> Next, I must identify <landmark 2> and output its relative coordinates in the same format: (x2, y2). </step-2-reasoning>
<step-2-answer> The relative coordinates of <landmark 2>: (<x2>, <y2>). </step-2-answer>
<step-3-reasoning> I now calculate the distance between the two landmarks using the pixel dimensions (pixel_width, pixel_height) = (<pixel_width>, <pixel_height>), the image size (image_width, image_height) = (<image_width>, <image_height>), and the distance formula: distance = sqrt(((x2 - x1) * image_width * pixel_width)^2 + ((y2 - y1) * image_height * pixel_height)^2) = sqrt(((<x2> - <x1>) * <image_width> * <pixel_width>)^2 + ((<y2> - <y1>) * <image_height> * <pixel_height>)^2) = <distance>. </step-3-reasoning>
<step-3-answer> The distance: <distance>. </step-3-answer>
</think>
<answer> <distance> </answer>
\end{lstlisting}
\end{promptbox}

\begin{promptbox}[label={box:cot_angle}]{Angle measurement: prompt components and CoT answer template}
{\footnotesize\textbf{Task description}}
\begin{lstlisting}[style=promptstyle]
estimate the angle of <measurement name> in <unit>, which is the angle between 2 lines: (line 1) the line connecting <landmark 1> and <landmark 2>, (line 2) the line connecting <landmark 3> and <landmark 4>.
\end{lstlisting}
{\footnotesize\textbf{Format requirement}}
\begin{lstlisting}[style=promptstyle]
The final answer must be enclosed within <answer> </answer> tags. The answer should be a single decimal number without units or extra text.
\end{lstlisting}
{\footnotesize\textbf{CoT instruction}}
\begin{lstlisting}[style=promptstyle]
Step 1: Identify line 1 and record the relative coordinates of its two endpoints in the format (x, y) = (relative position in width direction, relative position in height direction). Denote the endpoints as (x1_line1, y1_line1) and (x2_line1, y2_line1).
Step 2: Identify line 2 and record the relative coordinates of its two endpoints in the same (x, y) format. Denote them as (x1_line2, y1_line2) and (x2_line2, y2_line2).
Step 3: Given the pixel dimensions (pixel_width, pixel_height) and image size (image_width, image_height), compute the angle between the two lines using the formula: angle = arccos(|A · B| / (||A|| ||B||)), where A and B are the vectors of the two lines computed from the physical coordinates of their endpoints. A = ((x2_line1 - x1_line1) * image_width * pixel_width, (y2_line1 - y1_line1) * image_height * pixel_height) and B = ((x2_line2 - x1_line2) * image_width * pixel_width, (y2_line2 - y1_line2) * image_height * pixel_height). Denote A=(Ax, Ay) and B=(Bx, By). Then, angle = arccos(|Ax*Bx + Ay*By| / (sqrt(Ax^2 + Ay^2) * sqrt(Bx^2 + By^2))).
Report the reasoning process and final answer within <think> </think> and <answer> </answer> tags, respectively. Inside <think> </think>, include reasoning and step results using <step-k-reasoning> </step-k-reasoning> and <step-k-answer> </step-k-answer> tags.
\end{lstlisting}
{\footnotesize\textbf{CoT answer template}}
\begin{lstlisting}[style=promptstyle]
<think>
<step-1-reasoning> I need to identify the relative coordinates of <landmark 1> and <landmark 2> that define line 1. The relative coordinates must be written as (x1_line1, y1_line1), (x2_line1, y2_line1), where x is the relative position in width and y is the relative position in height. </step-1-reasoning>
<step-1-answer> The relative coordinates of <landmark 1> and <landmark 2>: (<x1_line1>, <y1_line1>), (<x2_line1>, <y2_line1>). </step-1-answer>
<step-2-reasoning> Next, I must identify the relative coordinates of <landmark 3> and <landmark 4> that define line 2, in the same format: (x1_line2, y1_line2), (x2_line2, y2_line2). </step-2-reasoning>
<step-2-answer> The relative coordinates of <landmark 3> and <landmark 4>: (<x1_line2>, <y1_line2>), (<x2_line2>, <y2_line2>). </step-2-answer>
<step-3-reasoning> I now calculate the angle between the two lines using the pixel dimensions (pixel_width, pixel_height) = (<pixel_width>, <pixel_height>), the image size (image_width, image_height) = (<image_width>, <image_height>), and the angle formula: angle = arccos(|A · B| / (||A|| ||B||)), where A and B are the vectors of the two lines computed from the physical coordinates of their endpoints. A = ((x2_line1 - x1_line1) * image_width * pixel_width, (y2_line1 - y1_line1) * image_height * pixel_height) = ( (<x2_line1> - <x1_line1>) * <image_width> * <pixel_width>, (<y2_line1> - <y1_line1>) * <image_height> * <pixel_height>) = (<Ax>, <Ay>). B = ((x2_line2 - x1_line2) * image_width * pixel_width, (y2_line2 - y1_line2) * image_height * pixel_height) = ( (<x2_line2> - <x1_line2>) * <image_width> * <pixel_width>, (<y2_line2> - <y1_line2>) * <image_height> * <pixel_height>) = (<Bx>, <By>). Denote A=(Ax, Ay) and B=(Bx, By). Then, angle = arccos(|Ax*Bx + Ay*By| / (sqrt(Ax^2 + Ay^2) * sqrt(Bx^2 + By^2))) = arccos(|<Ax>*<Bx> + <Ay>*<By>| / (sqrt(<Ax>^2 + <Ay>^2) * sqrt(<Bx>^2 + <By>^2))) = <angle> = <angle_degree> degrees. </step-3-reasoning>
<step-3-answer> The angle: <angle_degree>. </step-3-answer>
</think>
<answer> <angle_degree> </answer>
\end{lstlisting}
\end{promptbox}

   \section{Details of RFT via GRPO}
   \label{appendix:sec:rft_grpo} This appendix details the reward used in the RFT stage (Section~\ref{sec:fine-tuning}) and the GRPO training configuration. Algorithm~\ref{alg:rft_reward} summarizes the reward computation for one rollout.

   \noindent
   \textbf{Reward Components.} Each rollout is scored by up to three components, each in $[0, 1]$. The two accuracy-based components share one error-to-reward mapping $\rho(e)=\exp(-e)$: an exact prediction earns a reward of 1, the reward decays smoothly as the error grows, and an unparseable value earns 0. Because predicted coordinates are relative positions in $[0, 1]$ and measurement errors are relative errors, all errors are dimensionless and of comparable scale across tasks. \textbf{\textit{(i)}} The \textbf{\textit{format reward}} encourages responses that follow the required output structure, so that the reasoning steps and the final answer remain machine-parseable. For the T/L and A/D tasks, it combines a soft reasoning-structure score with a binary check of the final answer format, weighted 0.8 and 0.2, respectively; the reasoning-structure score grants partial credit for the presence of the \verb|<think>| block, the per-step reasoning and answer tags (i.e., \verb|<step-k-reasoning>| and \verb|<step-k-answer>|), well-formed step values, and the correct step order. For detection, only the binary answer-format check is used. \textbf{\textit{(ii)}} The \textbf{\textit{process reward}} supervises the reasoning chain itself, granting credit for accurate intermediate estimates rather than only for the final answer. Each CoT step declares one intermediate result: localization steps predict relative coordinates (e.g., landmarks or axis endpoints), and measurement steps predict an intermediate measurement value. The value in each \verb|<step-k-answer>| block is parsed and compared with its ground truth: localization steps are scored by the point-displacement error, and measurement steps by the relative error. Concretely, the localization error of a point is its Euclidean displacement from the ground truth point in relative coordinates, divided by $\sqrt{2}$, the diagonal of the unit square, so that it lies in $[0, 1]$; a step declaring multiple points is scored by its worst-localized point. Each step error is mapped through $\rho$, and the process reward is the mean of the per-step rewards (4 steps for T/L, 3 for A/D). \textbf{\textit{(iii)}} The \textbf{\textit{answer reward}} scores the final answer: the mean relative error of the parsed values for T/L and A/D, and the overlap error $e=(1-\text{CIoU})/2$ for detection. CIoU~\citep{DistanceIoU2020} augments IoU with a normalized center-distance penalty and an aspect-ratio consistency penalty:
   \begin{align}
      \text{CIoU} &= \text{IoU} - \frac{d^2}{c^2} - \alpha v, \\
      v &= \frac{4}{\pi^2}\left(\arctan\frac{w}{h} - \arctan\frac{\hat{w}}{\hat{h}}\right)^{2},
   \end{align}
   where $d$ is the distance between the centers of the predicted and ground truth boxes, $c$ is the diagonal length of the smallest box enclosing both, $(w, h)$ and $(\hat{w}, \hat{h})$ are the ground truth and predicted box dimensions, and $\alpha = v / \big((1-\text{IoU}) + v\big)$ down-weights the aspect-ratio penalty when the overlap is poor. Since $\text{CIoU} \in [-1, 1]$, the overlap error lies in $[0, 1]$, and unlike plain IoU, which is uniformly zero for any non-overlapping prediction, it provides a graded learning signal everywhere.

   \noindent
   \textbf{Reward Combination.} For the T/L and A/D tasks, the final reward is $r=r_{\text{format}}+r_{\text{process}} \cdot r_{\text{answer}}$. The multiplicative form makes the final-answer credit conditional on a verifiable chain of intermediate estimates: a rollout cannot collect the full answer reward from a final number that is inconsistent with its own reasoning trace, and accurate intermediate localizations pay off only through an accurate final measurement. The detection answer is a single localization step, so no process reward is defined and the reward reduces to $r=r_{\text{format}}+r_{\text{answer}}$.

\begin{algorithm*}[h]
   \caption{Reward computation for one RFT rollout.}
   \label{alg:rft_reward}
   \scriptsize
   \begin{algorithmic}[1]
      \Require rollout text $R$; task $t$; ground truth answer $\mathbf{y}$; ground truth step values $\mathbf{s}_1,\ldots,\mathbf{s}_n$ (T/L: $n{=}4$; A/D: $n{=}3$; detection: $n{=}0$); error-to-reward map $\rho(e)=\exp(-e)$
      \Statex
      \Process
      \State $r_{\text{format}} \gets \Call{FormatScore}{R}$ \Comment{tag-structure adherence, in $[0,1]$}
      \For{$k = 1,\ldots,n$} \Comment{process reward from CoT steps}
         \State $\hat{\mathbf{s}}_k \gets \Call{ParseStep}{R, k}$ \Comment{value in \texttt{<step-k-answer>}}
         \If{$\hat{\mathbf{s}}_k = \varnothing$}
            \State $r_k \gets 0$
         \Else
            \State $e_k \gets \Call{StepError}{\hat{\mathbf{s}}_k, \mathbf{s}_k}$ \Comment{localization step: worst-case point displacement; measurement step: relative error}
            \State $r_k \gets \rho(e_k)$
         \EndIf
      \EndFor
      \State $r_{\text{process}} \gets \frac{1}{n}\sum_{k=1}^{n} r_k$
      \State $\hat{\mathbf{y}} \gets \Call{ParseAnswer}{R}$ \Comment{values in \texttt{<answer>}}
      \If{$\hat{\mathbf{y}} = \varnothing$}
         \State $r_{\text{answer}} \gets 0$
      \Else
         \If{$t$ is detection}
            \State $e_a \gets (1-\text{CIoU}(\hat{\mathbf{y}}, \mathbf{y}))/2$ \Comment{overlap error}
         \Else
            \State $e_a \gets \frac{1}{d}\sum_{j=1}^{d} |\hat{y}_j - y_j| / y_j$ \Comment{mean relative error; $d$: number of answer values}
         \EndIf
         \State $r_{\text{answer}} \gets \rho(e_a)$
      \EndIf
      \If{$t$ is detection}
         \State $r \gets r_{\text{format}} + r_{\text{answer}}$ \Comment{no intermediate steps}
      \Else
         \State $r \gets r_{\text{format}} + r_{\text{process}} \cdot r_{\text{answer}}$
      \EndIf
      \Statex
      \Ensure scalar reward $r$
   \end{algorithmic}
\end{algorithm*}

   \noindent
   \textbf{Training Configuration.} RFT uses the GRPO implementation in verl~\citep{HybridFlow2025} with full-parameter fine-tuning, and the three sequential stages (A/D $\rightarrow$ T/L $\rightarrow$ detection) share one configuration, each stage initialized from the checkpoint of the previous one: 8 rollouts per prompt (the GRPO group size), a training batch of 256 prompts per step, with policy updates performed on mini-batches of 128 prompts, a constant learning rate of $3\times10^{-6}$, KL regularization toward the reference policy in the loss with coefficient 0.01 (low-variance estimator), and maximum prompt and response lengths of 4,096 tokens. Rollouts are generated with vLLM on 4 H200 GPUs (Appendix~\ref{appendix:sec:compute_software}). The multi-task RFT ablation (Table~\ref{tab:rft_strategy_ablation}) keeps this configuration in a single stage over the task mixture, with two additions: temperature-scaled task mixing (Appendix~\ref{appendix:sec:temperature_sampler}) that flattens the task ratio toward balance (approximately 42\%, 29\%, and 29\% of draws per epoch for detection, T/L, and A/D, respectively), and an epoch-level curriculum (Appendix~\ref{appendix:sec:curriculum}) that progressively filters out solved samples.

   \section{Temperature-Scaled Task Mixing}
   \label{appendix:sec:temperature_sampler} The multi-task training mixture is highly imbalanced: 110K detection samples against 5.5K T/L and 5.5K A/D samples (Section~\ref{sec:fine-tuning}). Sampling proportionally to these counts would devote about 91\% of all draws to detection, whereas uniform task probabilities would ignore the size of the detection set entirely. The SFT stage and the multi-task RFT ablation therefore rebalance the mixture with temperature-scaled task mixing, which interpolates between these two extremes.

   \noindent
   \textbf{Mechanism.} Samples are grouped into three tasks (angle and distance measurements are merged into A/D). For a task $t$ with $N_t$ samples, the task-level sampling probability is
   \begin{align}
      P(t) = \frac{N_t^{1/T}}{\sum_{u} N_u^{1/T}},
   \end{align}
   where $T > 0$ is the mixing temperature: $T{=}1$ recovers the natural task proportions, and increasing $T$ flattens the probabilities toward uniform. Each sample of task $t$ receives the weight $P(t)/N_t$, which is uniform within a task and reproduces the task-level probabilities exactly under weighted random sampling. One epoch draws as many samples as the training set contains, \textit{with replacement}, so a minority-task sample can be drawn more than once per epoch while a majority-task sample may be skipped; the sampler is seeded for reproducibility.

   \noindent
   \textbf{Instantiations in SFT and RFT.} The SFT stage uses $T{=}5$, giving task probabilities of approximately 47.7\% for detection and 26.2\% each for T/L and A/D (natural proportions: 90.9\%, 4.5\%, and 4.5\%). The multi-task RFT ablation uses $T{=}8$, giving approximately 42.1\% for detection and 29.0\% each for T/L and A/D. When composed with the epoch-level curriculum (Appendix~\ref{appendix:sec:curriculum}), the task counts and sample weights are recomputed over the active set ($S_t$ in Algorithm~\ref{alg:curriculum}) at every epoch, and the sampler falls back to unweighted sampling if fewer than two distinct tasks remain active.

   \section{Epoch-Level Curriculum Learning for Multi-Task RFT}
   \label{appendix:sec:curriculum} In GRPO, the advantage is normalized within the group of rollouts of one prompt. Once the policy reliably solves a prompt, all rollouts in its group receive nearly identical rewards, the within-group variance collapses, and the advantage approaches zero: the sample contributes almost no gradient while still consuming a full group of rollouts each time it is drawn. The multi-task RFT run therefore adopts an epoch-level curriculum that progressively removes solved samples from training, concentrating rollout compute on the samples the policy still fails. Unlike the classic easy-to-hard curriculum~\citep{Curriculum2009}, the schedule emphasizes hard examples, in line with dynamic-sampling practice in reinforcement learning of large language models~\citep{DAPO2025}. Algorithm~\ref{alg:curriculum} summarizes the procedure.

   \noindent
   \textbf{Sample Pools and Evidence.} For each task, samples are partitioned into two pools: a \textbf{\textit{hard pool}} $H_t$ of samples under active training and an \textbf{\textit{easy pool}} $E_t$ of samples classified as solved (notation as in Algorithm~\ref{alg:curriculum}). The first epoch trains on the full dataset. Within each epoch, the mean reward and the mean final-answer error of every drawn sample are computed over all of its rollouts. Averaging over rollouts estimates the policy's expected performance on the sample: a sample counts as solved only when the policy solves it reliably, not when a single rollout succeeds by chance. At the end of the epoch, these epoch-level means update a persistent per-sample record comprising exponential moving averages (EMA, weight $\alpha{=}0.4$) of the reward and of the answer error ($\bar{r}_i$ and $\bar{e}_i$), together with the numbers of consecutive observed epochs in which the sample's error remained below, or above, the task-specific error threshold $g_t$ ($c_i^{+}$ and $c_i^{-}$): the former is required for promotion to the easy pool, and the latter for demotion back to the hard pool. The threshold $g_t$ matches the benchmark's own success criteria: MRE $<0.1$ for the T/L and A/D tasks, and overlap error $(1-\text{CIoU})/2<0.25$ for detection (implying IoU $>0.5$). This record makes pool assignment robust to the stochasticity of a single epoch's rollouts: although one qualifying epoch is formally sufficient for promotion in our configuration ($\nu^{+}{=}1$), an unfavorable epoch lowers the EMA, and several favorable epochs are then required before the promotion criterion is met.

   \noindent
   \textbf{Pool Updates.} At the end of each epoch, the following updates are applied to each task. \textbf{\textit{(i) Promotion.}} The hard-pool samples drawn during the epoch are ranked by EMA reward $\bar{r}_i$, and at most the top 20\% ($f{=}0.2$) of the task's current training set is eligible for promotion; of these, only samples whose EMA answer error $\bar{e}_i$ falls below the threshold $g_t$ and whose passing count $c_i^{+}$ reaches the required number of consecutive epochs $\nu^{+}$ are moved to the easy pool. \textbf{\textit{(ii) Retention mix-in.}} To mitigate catastrophic forgetting of solved samples, the \textit{most recently promoted easy samples} are mixed back into training (the mix-in set $M_t$); their share $s_t$ increases with the task's solved fraction $p_t$, up to a maximum composition of 30\% easy ($m{=}0.3$) and 70\% hard samples once half of the task is solved ($p^{*}{=}0.5$). The promotion fraction $f$ and the maximum mix-in share $m$, like the other constants in this section, are tunable hyperparameters; we report the values used in our experiments. \textbf{\textit{(iii) Auditing and demotion.}} A rotating audit subset $A_t$ (5\% of the hard pool per epoch, $a{=}0.05$) re-evaluates the \textit{least recently audited easy samples}; a mixed-in or audited sample (in $M_t \cup A_t$) whose EMA error $\bar{e}_i$ exceeds $\lambda{=}1.5$ times the task's error threshold $g_t$ and whose failing count $c_i^{-}$ reaches the required number of consecutive epochs $\nu^{-}$ is returned to the hard pool, which corrects erroneous promotions and detected forgetting. The margin $\lambda$ introduces hysteresis, preventing borderline samples from oscillating between the two pools. Mix-in and audit are complementary but distinct: the mix-in keeps the \textit{most recently promoted samples} under continued training to prevent forgetting, whereas the audit rotates through the \textit{least recently audited samples} so that every easy sample is eventually re-evaluated. Neither changes pool membership by itself: mixed-in and audited samples remain in the easy pool $E_t$ while temporarily rejoining the active set $S_t$, and since they are the only easy samples with fresh rollouts, they are the only demotion candidates, demoted only when the above criterion is met. \textbf{\textit{(iv) Per-task minimum.}} At least 10\% ($\phi{=}0.1$) of each task's original samples remain in training, so that a nearly solved task is never excluded entirely: while the task remains solved, its rollout groups yield near-zero advantages and thus contribute little gradient, and if performance on the task degrades, the demotion rule returns its samples to training.

   \noindent
   \textbf{Composition with Task Balancing.} The temperature-scaled task-mixing distribution (Appendix~\ref{appendix:sec:temperature_sampler}) is recomputed over the active set $S_t$ at every epoch, so task-level rebalancing adapts as the large detection pool shrinks relative to the T/L and A/D pools. Epochs shorten as the pools shrink.

\begin{algorithm*}[h]
   \caption{Epoch-level curriculum for multi-task RFT.}
   \label{alg:curriculum}
   \scriptsize
   \begin{algorithmic}[1]
      \Require hard pool $H_t$ (all $N_t$ samples of task $t$); easy pool $E_t \gets \varnothing$; mix-in and audit subsets $M_t, A_t \gets \varnothing$; per-sample EMAs $\bar{r}_i, \bar{e}_i$ and counters $c_i^{+}, c_i^{-} \gets 0$; error threshold $g_t$ (T/L, A/D: MRE $0.1$; detection: overlap error $0.25$); promotion fraction $f{=}0.2$; maximum mix-in share $m{=}0.3$, reached at solved fraction $p^{*}{=}0.5$; hysteresis margin $\lambda{=}1.5$; audit fraction $a{=}0.05$; per-task minimum fraction $\phi{=}0.1$; EMA weight $\alpha{=}0.4$; required consecutive passing (failing) epochs $\nu^{+}{=}1$ ($\nu^{-}{=}1$)
      \Statex
      \Process
      \State $S_t \gets H_t$ for every task $t$
      \For{epoch $k = 1, 2, \ldots$}
         \State run GRPO on $\bigcup_t S_t$; compute the mean reward $r_i^{(k)}$ and mean answer error $e_i^{(k)}$ of every drawn sample $i$ over its rollouts, where $D_t^{(k)}$ denotes the samples of task $t$ drawn in epoch $k$
         \For{each task $t$}
            \For{$i \in D_t^{(k)}$}
               \State $\bar{r}_i \gets \alpha\, r_i^{(k)} + (1{-}\alpha)\,\bar{r}_i$; \quad $\bar{e}_i \gets \alpha\, e_i^{(k)} + (1{-}\alpha)\,\bar{e}_i$ \Comment{update the reward and error EMAs}
               \State $c_i^{+} \gets \mathbb{I}[e_i^{(k)} < g_t] \cdot (c_i^{+}{+}1)$; \quad $c_i^{-} \gets \mathbb{I}[e_i^{(k)} \geq g_t] \cdot (c_i^{-}{+}1)$ \Comment{count consecutive passing/failing epochs}
            \EndFor
            \State $T_t \gets$ the top $f \cdot |S_t|$ samples of $D_t^{(k)} \cap H_t$ ranked by EMA reward $\bar{r}_i$ \Comment{promotion candidates: at most a fraction $f$ of the active set}
            \State $P_t \gets \{\, i \in T_t : \bar{e}_i < g_t,\ c_i^{+} \geq \nu^{+} \,\}$ \Comment{promote: error below threshold for $\nu^{+}$ consecutive epochs}
            \State $Q_t \gets \{\, i \in M_t \cup A_t : \bar{e}_i \geq \lambda\, g_t,\ c_i^{-} \geq \nu^{-} \,\}$ \Comment{demote: error regressed beyond the hysteresis margin}
            \State $E_t \gets (E_t \cup P_t) \setminus Q_t$; \quad $H_t \gets (H_t \setminus P_t) \cup Q_t$ \Comment{apply promotions and demotions to the pools}
            \State $p_t \gets |E_t| / N_t$; \quad $s_t \gets m \cdot \min(1,\, p_t/p^{*})$ \Comment{solved fraction; mix-in share}
            \State $M_t \gets$ the $\mathrm{round}\big(s_t\,|H_t| / (1{-}s_t)\big)$ most recently promoted samples of $E_t$ \Comment{mix solved samples back in against forgetting}
            \State $A_t \gets$ the $\mathrm{round}(a\,|H_t|)$ least recently audited samples of $E_t$ \Comment{rotating audit of the easy pool}
            \State $S_t \gets H_t \cup M_t \cup A_t$, extended with the most recently promoted samples of $E_t$ until $|S_t| \geq \mathrm{round}(\phi\, N_t)$ \Comment{enforce the per-task minimum active size}
         \EndFor
         \State recompute the temperature-scaled task-mixing distribution over $\bigcup_t S_t$
      \EndFor
      \Statex
      \Ensure per-epoch active sets $S_t$ that concentrate rollout compute on unsolved samples while retaining solved samples for re-evaluation
   \end{algorithmic}
\end{algorithm*}

   \section{MedVision-V0 Overview}
   \label{appendix:medvision_v0_overview}

\begin{figure*}[h]
      \begin{center}
         \includegraphics[width=1\linewidth]{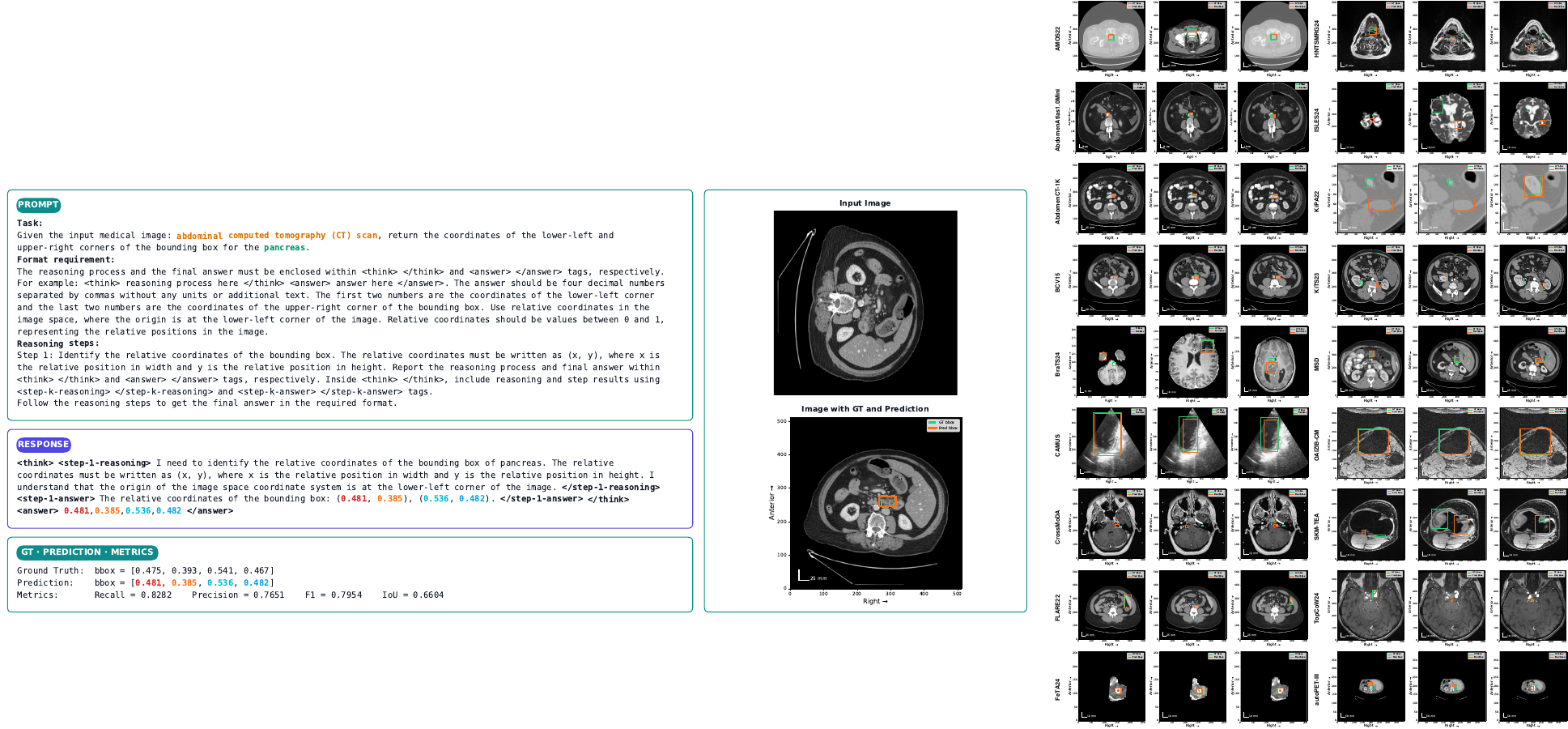}
      \end{center}
      \caption{Model responses in the detection task. Left: prompt, model response, and the ground truth, prediction and metrics of one example. Middle: the input image, and the image with the ground truth and predicted bounding boxes. Right: random samples of model predictions, grouped by dataset.}
      \label{appendix:fig:examples_detection}
   \end{figure*}

\begin{figure*}[h]
      \begin{center}
         \includegraphics[width=1\linewidth]{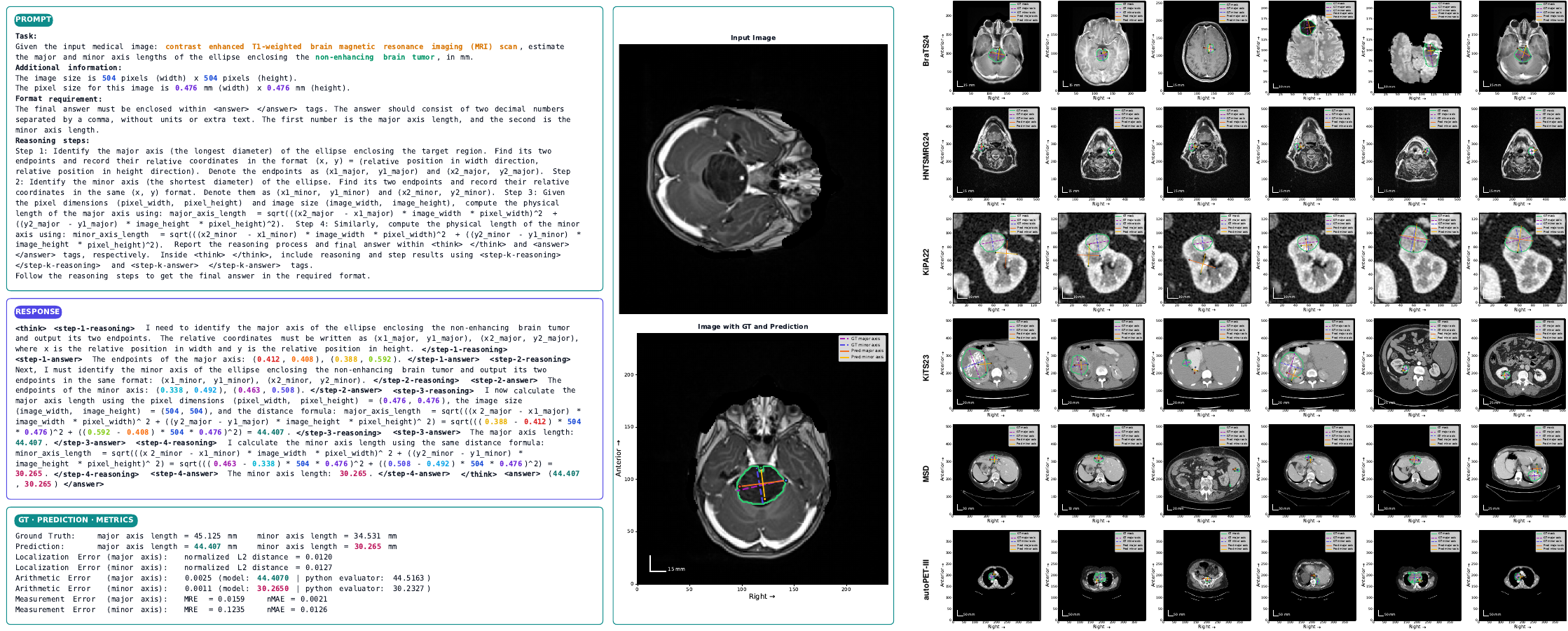}
      \end{center}
      \caption{Model responses in the T/L size task. Left: prompt, model response, and the ground truth, prediction and metrics of one example. Middle: the input image, and the image with the ground truth and predicted major and minor axes. Right: random samples of model predictions, grouped by dataset.}
      \label{appendix:fig:examples_tl}
   \end{figure*}

\begin{figure*}[h]
      \begin{center}
         \includegraphics[width=1\linewidth]{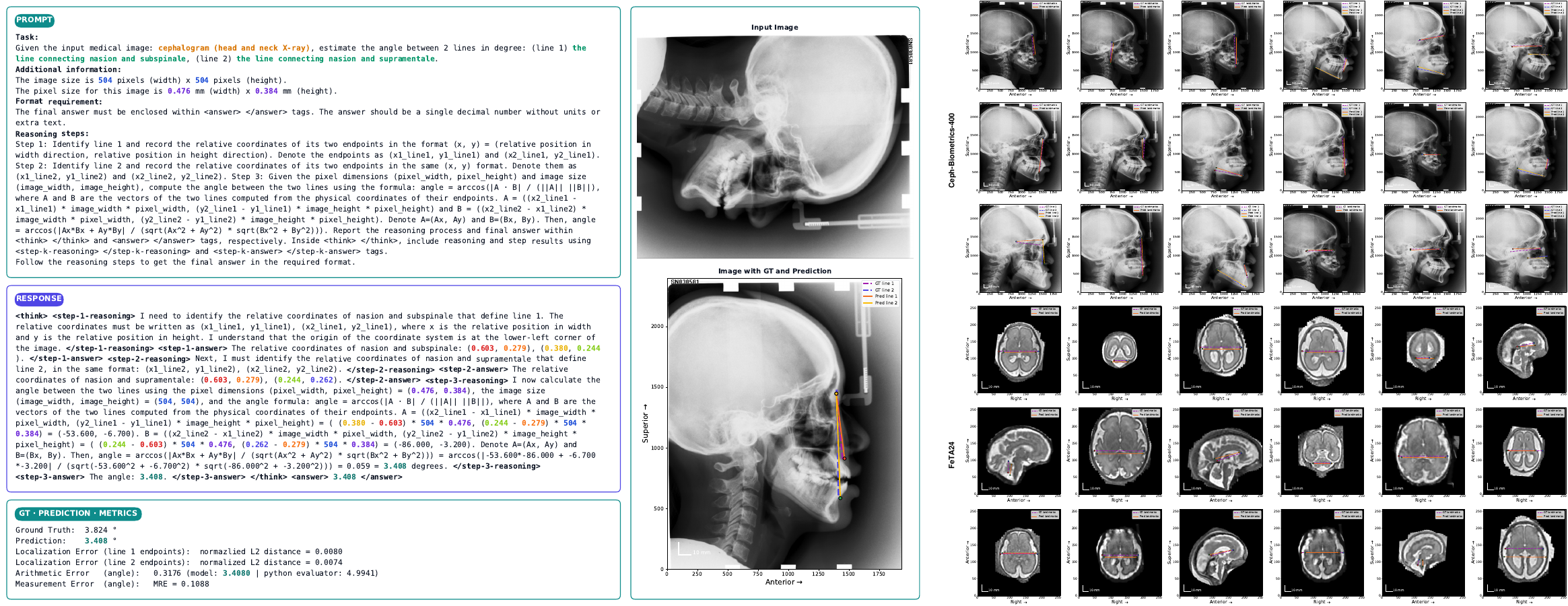}
      \end{center}
      \caption{Model responses in the A/D task. Left: prompt, model response, and the ground truth, prediction and metrics of one example. Middle: the input image, and the image with the ground truth and predicted landmarks and measurement. Right: random samples of model predictions, grouped by dataset.}
      \label{appendix:fig:examples_ad}
   \end{figure*}

   \noindent
   \textbf{Qualitative Results.} Figures~\ref{appendix:fig:examples_detection},~\ref{appendix:fig:examples_tl}, and~\ref{appendix:fig:examples_ad} show MedVision-V0 responses for the detection, T/L size and A/D tasks, respectively. In each figure, the left column gives the full prompt, the model response, and the ground truth, prediction and metrics for one example; the middle column shows the input image and the same image with the ground truth and prediction overlaid; and the right panel shows random samples of model predictions, grouped by dataset.

\begin{figure*}[h]
      \centering
      \includegraphics[width=1\linewidth]{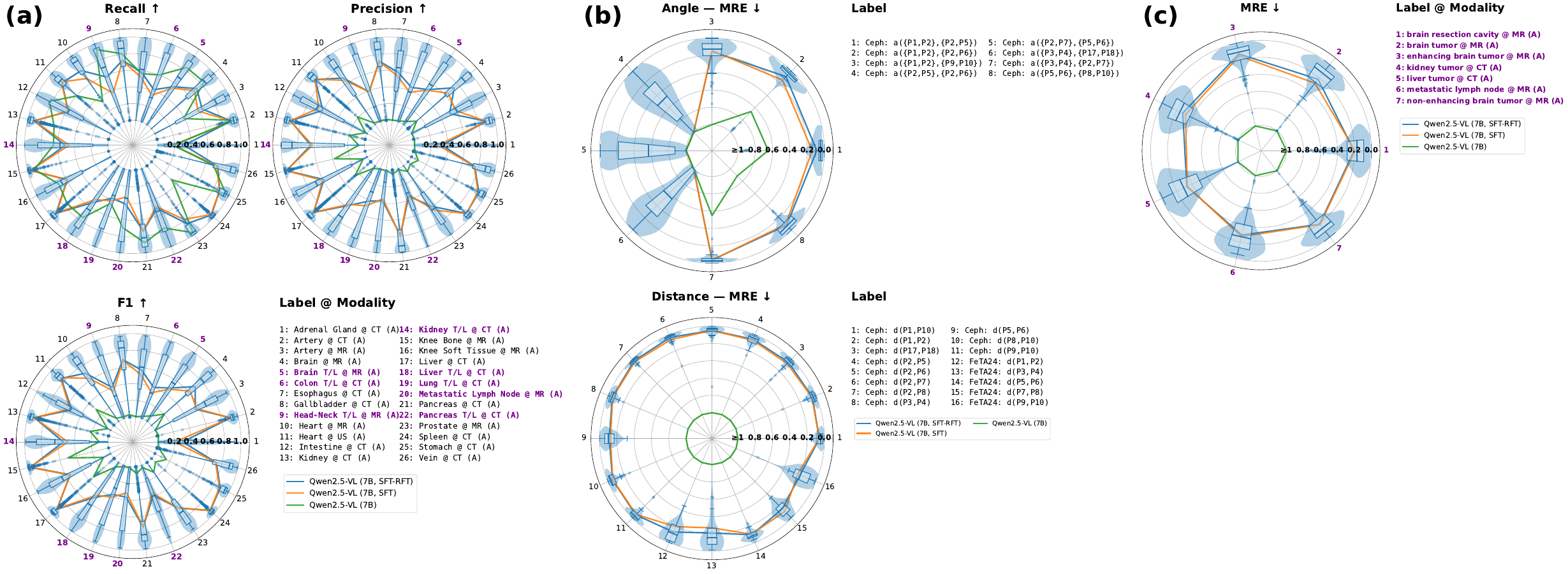}
      \caption{Effectiveness of the SFT-RFT training strategy on the three benchmark tasks. Each panel shows the average performance metric of each label group for the base (Qwen2.5-VL-7B), SFT, and SFT-RFT (MedVision-V0) models; violin plots show the distribution of metric values within each group for the SFT-RFT model.}
      \label{fig:sft_rft_training_strategy}
   \end{figure*}

   \noindent
   \textbf{Quantitative Results.} Figure~\ref{fig:sft_rft_training_strategy} summarizes the performance gains from each training stage (base $\to$ SFT $\to$ SFT-RFT): the average performance metric in each label group is presented for the base, SFT, and SFT-RFT models, and the distribution of metric values in each group for the SFT-RFT model is visualized as a violin plot. Table~\ref{tab:sft_rft_ablation} reports the full quantitative ablation across all three tasks.

   \noindent
   \textbf{Base Model Ablation.} Table~\ref{tab:sft_other_models} compares the base $\to$ SFT progression across three base models (Qwen2.5-VL, Gemma4, and MedGemma), all of which improve substantially across the three tasks.

   \noindent
   \textbf{Multi-Task RFT.} The RFT stage of MedVision-V0 trains on the three tasks sequentially (A/D $\rightarrow$ T/L $\rightarrow$ detection; Section~\ref{sec:fine-tuning}), and this task order is a design choice. To examine its impact, we train an ablation variant that starts from the same SFT model and replaces the three sequential stages with a single RFT stage on a mixture of all task data, using a task-balancing sampler and an epoch-level curriculum (Appendix~\ref{appendix:sec:curriculum}). Table~\ref{tab:rft_strategy_ablation} compares the two variants on the detection, T/L size estimation, and A/D measurement benchmarks. The multi-task variant is nearly identical to the sequential model on detection (F1 $78.2\%$ vs. $79.1\%$ on anatomy; $47.4\%$ vs. $46.9\%$ on T/L), T/L size estimation (MRE $26.7\%$ vs. $26.0\%$), and distance measurement (MRE $6.6\%$ vs. $6.4\%$), and performs better on angle measurement (MAE $3.7$ vs. $4.7$ degrees; MRE $41.9\%$ vs. $52.1\%$). As A/D is optimized in the first stage of the sequential schedule, the sequential model's weaker angle measurement may be attributable to catastrophic forgetting during the subsequent T/L and detection stages. These results indicate that multi-task RFT is a viable alternative to the multi-stage strategy.

   \noindent
   \textbf{RFT Reward Design.} For the T/L and A/D tasks, the final reward combines the process and answer rewards multiplicatively, $r=r_{\text{format}}+r_{\text{process}} \cdot r_{\text{answer}}$, making the final-answer credit conditional on accurate intermediate estimates (Appendix~\ref{appendix:sec:rft_grpo}). To assess this design choice, we train an ablation variant of the multi-task RFT model that instead combines the two rewards additively ($r=r_{\text{format}}+r_{\text{process}}+r_{\text{answer}}$), with all other training settings unchanged. As the detection task defines no process reward, the comparison concerns the T/L size estimation and A/D measurement benchmarks (Table~\ref{tab:rft_reward_ablation}). The two designs perform comparably on T/L size estimation (MRE $26.7\%$ vs. $26.9\%$) and distance measurement (MRE $6.6\%$ vs. $6.7\%$), while the multiplicative design performs better on angle measurement (MRE $41.9\%$ vs. $48.1\%$). These results support the multiplicative composition adopted in MedVision-V0.

\begin{figure*}[h]
      \centering
      \includegraphics[width=1\linewidth]{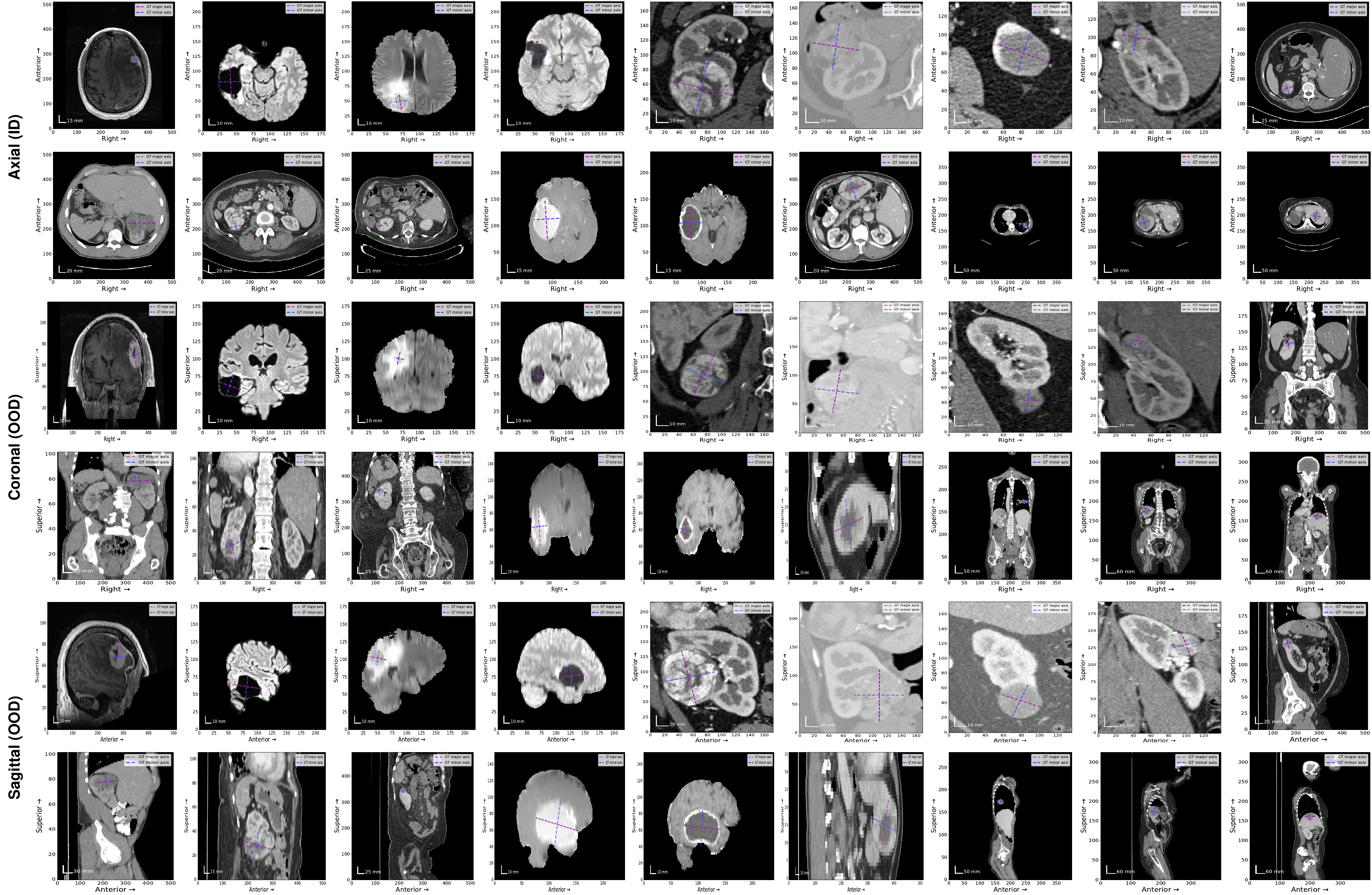}
      \caption{Examples of the Plane-OOD setting in the T/L size estimation task. Panels are grouped by imaging plane: axial (in-distribution; the only plane used during training) and coronal and sagittal (OOD; held out from training). Panels at the same grid position within each plane group show the same subject and target, with the ground truth major and minor axes overlaid and an L-shaped bar indicating the physical scale. Although the subject and target are identical, the OOD planes present markedly different \textbf{\textit{anatomical context, target appearance, and physical extent}}, which the model must generalize across.}
      \label{appendix:fig:planeOOD_tl}
   \end{figure*}

\begin{figure*}[h]
      \centering
      \includegraphics[width=1\linewidth]{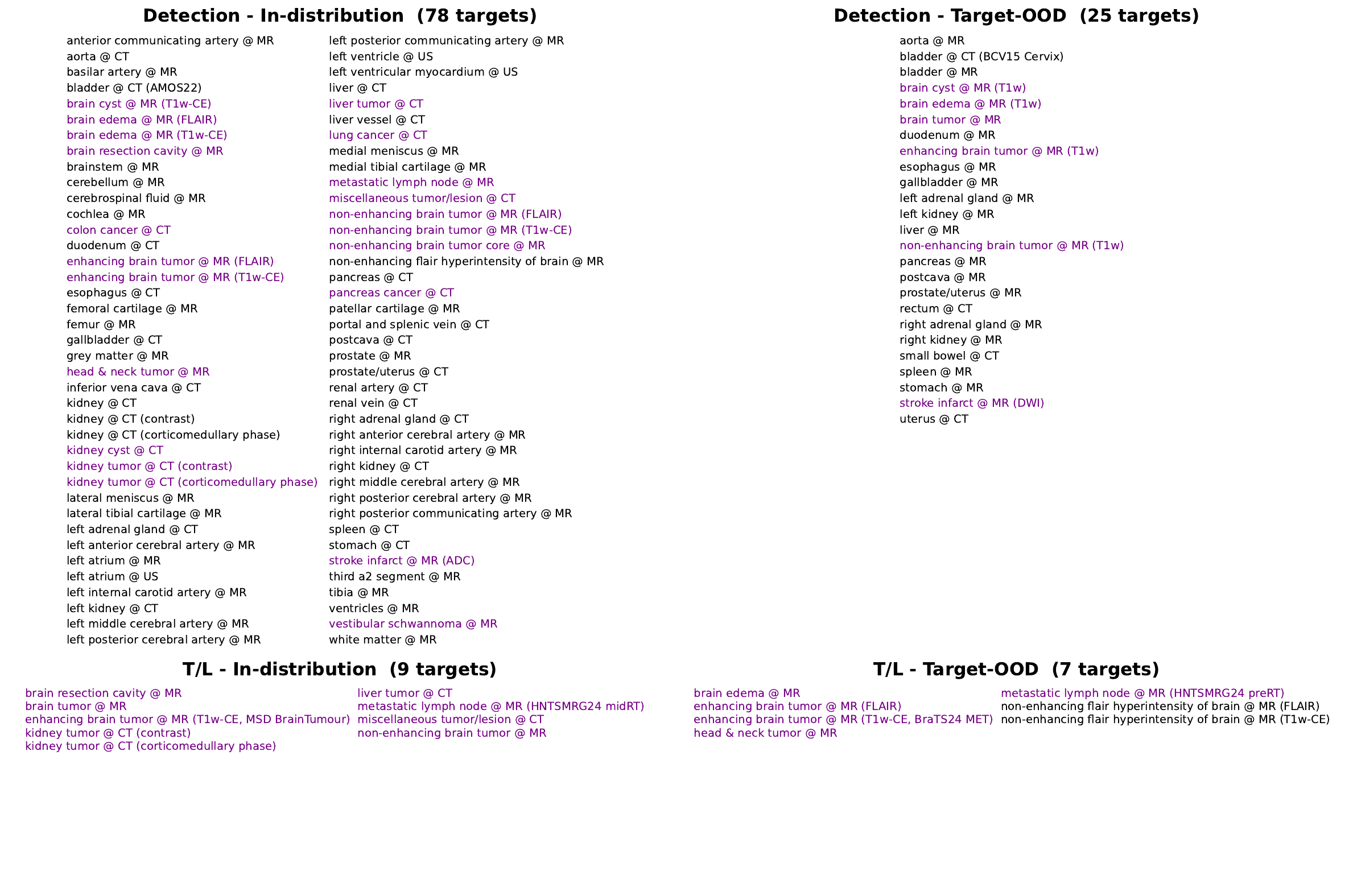}
      \caption{In-distribution and Target-OOD targets of the detection (top) and T/L size estimation (bottom) tasks. Each entry is one benchmark target, written as label @ modality, with the source dataset or MR sequence given in parentheses where needed to distinguish targets. The Target-OOD targets are entirely absent from the training label set and constitute the Target-OOD test sets in Table~\ref{tab:ood_generalization}.}
      \label{appendix:fig:targetOOD_labels}
   \end{figure*}

   \noindent
   \textbf{OOD Ablation.} Table~\ref{tab:ood_generalization} presents out-of-distribution generalization results. We evaluate two OOD settings: \textbf{\textit{(i) Plane-OOD}}, where test samples involve imaging planes not seen during training for a given anatomical target (e.g., a structure trained on axial CT slices evaluated on coronal or sagittal views); and \textbf{\textit{(ii) Target-OOD}}, where test samples involve anatomical targets (structures or tumors) entirely absent from the training label set; Figure~\ref{appendix:fig:targetOOD_labels} lists the in-distribution and Target-OOD targets of the detection and T/L size estimation tasks. Figure~\ref{appendix:fig:planeOOD_tl} illustrates the Plane-OOD setting in the T/L size estimation task: each grid position shows the same subject and target sliced along the axial (in-distribution) plane and the coronal and sagittal (OOD) planes, revealing the markedly different \textbf{\textit{anatomical context, target appearance, and physical extent}} that the model must generalize across. Across both OOD settings, SFT-RFT consistently outperforms the SFT model, indicating that reinforcement learning fosters stronger generalization beyond memorized training distributions. This observation is consistent with the finding that RL generalizes better than SFT~\citep{SFT2025a}. MedVision-V0 nonetheless trails its in-distribution performance: anatomy detection F1 drops from $79.1\%$ (Table~\ref{tab:detection_performance}) to $49.5\%$ (plane-OOD) and $36.0\%$ (target-OOD), T/L detection F1 from $46.9\%$ to $36.6\%$ and $45.5\%$, and T/L size MRE rises from $26.0\%$ (Table~\ref{tab:tl_size_performance}) to $36.4\%$ and $34.7\%$.

   \begin{table*}[h]
      \centering
      \caption{Training progression of MedVision-V0: base (Qwen2.5-VL-7B) $\to$ SFT $\to$ SFT-RFT (MedVision-V0) across all three tasks. Parentheses give the change from the preceding stage (SFT vs.\ base; SFT-RFT vs.\ SFT; \textcolor{green!55!black}{green}: improvement; \textcolor{red!75!black}{red}: decline; \textcolor{gray}{gray}: no change).}
      \label{tab:sft_rft_ablation} 
      \tiny
      \setlength{\tabcolsep}{2pt}
      \begin{adjustbox}
         {max width=\textwidth}
         \begin{tabular}{lrrrrrrr}
            \toprule 
             \textbf{Model} & \textbf{R} $\uparrow$ & \textbf{P} $\uparrow$ & \textbf{F1} $\uparrow$ & \textbf{IoU} $\uparrow$ & \textbf{SR} $\uparrow$ & $\textbf{IoU}_{\textbf{\textgreater0.5}}$ $\uparrow$ & $\textbf{IoU}_{\textbf{0.5:0.05:0.95}}$ $\uparrow$ \\
            \midrule \multicolumn{8}{l}{\textbf{\textit{Detection: Anatomy (18 regions, 13.4K samples)}}} \\
            base             & 69.7 & 12.2 & 16.7 & 11.3 & 99.4 & 5.6 & 1.7 \\
            \quad SFT        & 79.8\impv{+10.1} & 80.5\impv{+68.3} & 78.3\impv{+61.6} & 71.2\impv{+59.9} & 100.0\impv{+0.6} & 79.3\impv{+73.7} & 59.0\impv{+57.3} \\
            \quad\quad SFT-RFT & 81.3\impv{+1.5} & 80.4\decl{-0.1} & 79.1\impv{+0.8} & 72.0\impv{+0.8} & 100.0\nochg{0.0} & 80.1\impv{+0.8} & 59.9\impv{+0.9} \\
            \midrule \multicolumn{8}{l}{\textbf{\textit{Detection: Tumor/Lesion (8 regions, 8.5K samples)}}} \\
            base             & 77.4 & 3.8 & 6.5 & 3.6 & 99.6 & 0.0 & 0.0 \\
            \quad SFT        & 47.9\decl{-29.5} & 49.3\impv{+45.5} & 44.1\impv{+37.6} & 35.4\impv{+31.8} & 100.0\impv{+0.4} & 37.0\impv{+37.0} & 17.9\impv{+17.9} \\
            \quad\quad SFT-RFT & 52.4\impv{+4.5} & 50.5\impv{+1.2} & 46.9\impv{+2.8} & 38.2\impv{+2.8} & 100.0\nochg{0.0} & 40.7\impv{+3.7} & 21.1\impv{+3.2} \\
            \midrule \midrule  \textbf{Model} & \textbf{MAE} $\downarrow$ & \textbf{MRE} $\downarrow$ & \textbf{SR} $\uparrow$ & $\textbf{MRE}_{\textbf{\textless0.1}}$ $\uparrow$ \\
            \midrule \multicolumn{8}{l}{\textbf{\textit{Tumor/Lesion Size (2K samples)}}} \\
            base             & 2897.2 & 7680.9 & 95.4 & 0.7 \\
            \quad SFT        & 11.3\impv{-2885.9} & 28.6\impv{-7652.3} & 100.0\impv{+4.6} & 20.6\impv{+19.9} \\
            \quad\quad SFT-RFT & 10.5\impv{-0.8} & 26.0\impv{-2.6} & 100.0\nochg{0.0} & 23.5\impv{+2.9} \\
            \midrule \midrule \textbf{Model} & \textbf{MAE} $\downarrow$ & \textbf{MRE} $\downarrow$ & \textbf{SR} $\uparrow$ & $\textbf{MRE}_{\textbf{\textless0.1}}$ $\uparrow$ \\
            \midrule \multicolumn{8}{l}{\textbf{\textit{Distance (Ceph+FeTA, 1,100 samples)}}} \\
            base             & 63603.8 & 96542.0 & 98.3 & 0.5 \\
            \quad SFT        & 4.2\impv{-63599.6} & 7.5\impv{-96534.5} & 100.0\impv{+1.7} & 77.7\impv{+77.2} \\
            \quad\quad SFT-RFT & 3.6\impv{-0.6} & 6.4\impv{-1.1} & 100.0\nochg{0.0} & 81.4\impv{+3.7} \\
            \midrule \multicolumn{8}{l}{\textbf{\textit{Angle (Ceph, 960 samples)}}} \\
            base             & 48.0 & 724.9 & 97.6 & 2.0 \\
            \quad SFT        & 5.8\impv{-42.2} & 53.3\impv{-671.6} & 100.0\impv{+2.4} & 48.8\impv{+46.8} \\
            \quad\quad SFT-RFT & 4.7\impv{-1.1} & 52.1\impv{-1.2} & 99.9\decl{-0.1} & 52.0\impv{+3.2} \\
            \bottomrule
         \end{tabular}
      \end{adjustbox}
   \end{table*}

   \begin{table*}[h]
      \centering
      \caption{Training progression base $\to$ SFT for three base models: Qwen2.5-VL (7B), Gemma4 (31B), and MedGemma (27B), across all three tasks. For SFT rows, parentheses give the change from the corresponding base model (\textcolor{green!55!black}{green}: improvement; \textcolor{red!75!black}{red}: decline).}
      \label{tab:sft_other_models} 
      \tiny
      \setlength{\tabcolsep}{2pt}
      \begin{adjustbox}
         {max width=\textwidth}
         \begin{tabular}{lrrrrrrr}
            \toprule  \textbf{Model} & \textbf{R} $\uparrow$ & \textbf{P} $\uparrow$ & \textbf{F1} $\uparrow$ & \textbf{IoU} $\uparrow$ & \textbf{SR} $\uparrow$ & $\textbf{IoU}_{\textbf{\textgreater0.5}}$ $\uparrow$ & $\textbf{IoU}_{\textbf{0.5:0.05:0.95}}$ $\uparrow$ \\
            \midrule \multicolumn{8}{l}{\textbf{\textit{Detection: Anatomy (18 regions, 13.4K samples)}}} \\
            Qwen2.5-VL       & 69.7 & 12.2 & 16.7 & 11.3 & 99.4 & 5.6 & 1.7 \\
            \quad + SFT      & 79.8\impv{+10.1} & 80.5\impv{+68.3} & 78.3\impv{+61.6} & 71.2\impv{+59.9} & 100.0\impv{+0.6} & 79.3\impv{+73.7} & 59.0\impv{+57.3} \\
            Gemma4           & 34.8 & 20.7 & 22.6 & 16.7 & 98.5 & 13.5 & 5.8 \\
            \quad + SFT      & 72.1\impv{+37.3} & 72.2\impv{+51.5} & 70.1\impv{+47.5} & 61.8\impv{+45.1} & 100.0\impv{+1.5} & 68.4\impv{+54.9} & 46.6\impv{+40.8} \\
            MedGemma         & 57.2 & 15.7 & 19.1 & 12.9 & 98.6 & 6.6 & 1.8 \\
            \quad + SFT      & 48.6\decl{-8.6} & 49.5\impv{+33.8} & 44.7\impv{+25.6} & 33.8\impv{+20.9} & 100.0\impv{+1.4} & 29.8\impv{+23.2} & 11.8\impv{+10.0} \\
            \midrule \multicolumn{8}{l}{\textbf{\textit{Detection: Tumor/Lesion (8 regions, 8.5K samples)}}} \\
            Qwen2.5-VL       & 77.4 & 3.8 & 6.5 & 3.6 & 99.6 & 0.0 & 0.0 \\
            \quad + SFT      & 47.9\decl{-29.5} & 49.3\impv{+45.5} & 44.1\impv{+37.6} & 35.4\impv{+31.8} & 100.0\impv{+0.4} & 37.0\impv{+37.0} & 17.9\impv{+17.9} \\
            Gemma4           & 43.6 & 14.6 & 18.1 & 12.7 & 99.5 & 8.4 & 2.7 \\
            \quad + SFT      & 52.8\impv{+9.2} & 54.1\impv{+39.5} & 48.8\impv{+30.7} & 39.9\impv{+27.2} & 100.0\impv{+0.5} & 44.4\impv{+36.0} & 21.8\impv{+19.1} \\
            MedGemma         & 54.1 & 4.6 & 7.5 & 4.3 & 97.0 & 0.1 & 0.0 \\
            \quad + SFT      & 12.4\decl{-41.7} & 15.4\impv{+10.8} & 10.6\impv{+3.1} & 7.0\impv{+2.7} & 100.0\impv{+3.0} & 2.5\impv{+2.4} & 0.7\impv{+0.7} \\
            \midrule \midrule \textbf{Model} & \textbf{MAE} $\downarrow$ & \textbf{MRE} $\downarrow$ & \textbf{SR} $\uparrow$ & $\textbf{MRE}_{\textbf{\textless0.1}}$ $\uparrow$ \\
            \midrule \multicolumn{8}{l}{\textbf{\textit{Tumor/Lesion Size (2K samples)}}} \\
            Qwen2.5-VL       & 2897.2 & 7680.9 & 95.4 & 0.7 \\
            \quad + SFT      & 11.3\impv{-2885.9} & 28.6\impv{-7652.3} & 100.0\impv{+4.6} & 20.6\impv{+19.9} \\
		    Gemma4           & 21.7 & 72.6 & 98.9 & 16.8 \\
		    \quad + SFT      & 10.5\impv{-11.2} & 28.1\impv{-44.5} & 100.0\impv{+1.1} & 25.0\impv{+8.2} \\
		    MedGemma         & 523.4 & 1905.0 & 57.0 & 0.5 \\
		    \quad + SFT      & 17.0\impv{-506.4} & 42.1\impv{-1862.9} & 100.0\impv{+43.0} & 10.3\impv{+9.8} \\
		    \midrule \midrule  \textbf{Model} & \textbf{MAE} $\downarrow$ & \textbf{MRE} $\downarrow$ & \textbf{SR} $\uparrow$ & $\textbf{MRE}_{\textbf{\textless0.1}}$ $\uparrow$ \\
		    \midrule \multicolumn{8}{l}{\textbf{\textit{Distance (Ceph+FeTA, 1,100 samples)}}} \\
		    Qwen2.5-VL       & 63603.8 & 96542.0 & 98.3 & 0.5 \\
		    \quad + SFT      & 4.2\impv{-63599.6} & 7.5\impv{-96534.5} & 100.0\impv{+1.7} & 77.7\impv{+77.2} \\
		    Gemma4           & 23.8 & 37.9 & 99.8 & 10.8 \\
		    \quad + SFT      & 6.2\impv{-17.6} & 10.1\impv{-27.8} & 100.0\impv{+0.2} & 60.2\impv{+49.4} \\
		    MedGemma         & 6239.0 & 8616.4 & 46.2 & 6.7 \\
		    \quad + SFT      & 5.2\impv{-6233.8} & 9.1\impv{-8607.3} & 100.0\impv{+53.8} & 68.6\impv{+61.9} \\
		    \midrule \multicolumn{8}{l}{\textbf{\textit{Angle (Ceph, 960 samples)}}} \\
		    Qwen2.5-VL       & 48.0 & 724.9 & 97.6 & 2.0 \\
		    \quad + SFT      & 5.8\impv{-42.2} & 53.3\impv{-671.6} & 100.0\impv{+2.4} & 48.8\impv{+46.8} \\
		    Gemma4           & 23.9 & 428.0 & 88.5 & 11.7 \\
		    \quad + SFT      & 4.3\impv{-19.6} & 57.3\impv{-370.7} & 100.0\impv{+11.5} & 49.8\impv{+38.1} \\
		    MedGemma         & 46.4 & 1024.8 & 93.0 & 6.8 \\
		    \quad + SFT      & 5.2\impv{-41.2} & 63.4\impv{-961.4} & 100.0\impv{+7.0} & 44.4\impv{+37.6} \\
		    \bottomrule
		 \end{tabular}
	      \end{adjustbox}
	   \end{table*}

	   \begin{table*}[h]
	      \centering
	      \caption{RFT data-mixture ablation: sequential three-stage RFT (A/D $\rightarrow$ T/L $\rightarrow$ detection; MedVision-V0) vs. single-stage RFT on a mixture of all task data, starting from the same SFT model. Detection metrics, MRE, SR, and MRE$_{<0.1}$ are reported in \%. MAEs are given in millimeters (T/L size, distance) and degrees (angle).}
	      \label{tab:rft_strategy_ablation} 
	      \tiny
	      \setlength{\tabcolsep}{2pt}
	      \begin{adjustbox}
		 {max width=\textwidth}
		 \begin{tabular}{lrrrrrrr}
		    \toprule  \textbf{Model} & \textbf{R} $\uparrow$ & \textbf{P} $\uparrow$ & \textbf{F1} $\uparrow$ & \textbf{IoU} $\uparrow$ & \textbf{SR} $\uparrow$ & $\textbf{IoU}_{\textbf{\textgreater0.5}}$ $\uparrow$ & $\textbf{IoU}_{\textbf{0.5:0.05:0.95}}$ $\uparrow$ \\
		    \midrule \multicolumn{8}{l}{\textbf{\textit{Detection: Anatomy (18 regions, 13.4K samples)}}} \\
		    Sequential RFT   & 81.3 & 80.4 & 79.1 & 72.0 & 100.0 & 80.1 & 59.9 \\
		    Multi-task RFT   & 80.4 & 79.3 & 78.2 & 71.2 & 100.0 & 79.6 & 59.4 \\
		    \midrule \multicolumn{8}{l}{\textbf{\textit{Detection: Tumor/Lesion (8 regions, 8.5K samples)}}} \\
		    Sequential RFT   & 52.4 & 50.5 & 46.9 & 38.2 & 100.0 & 40.7 & 21.1 \\
		    Multi-task RFT   & 50.2 & 52.9 & 47.4 & 38.5 & 100.0 & 40.3 & 20.5 \\
		    \midrule \midrule  \textbf{Model} & \textbf{MAE} $\downarrow$ & \textbf{MRE} $\downarrow$ & \textbf{SR} $\uparrow$ & $\textbf{MRE}_{\textbf{\textless0.1}}$ $\uparrow$ \\
		    \midrule \multicolumn{8}{l}{\textbf{\textit{Tumor/Lesion Size (2K samples)}}} \\
		    Sequential RFT   & 10.5 & 26.0 & 100.0 & 23.5 \\
		    Multi-task RFT   & 10.5 & 26.7 & 100.0 & 23.4 \\
		    \midrule \multicolumn{8}{l}{\textbf{\textit{Distance (Ceph+FeTA, 1,100 samples)}}} \\
		    Sequential RFT   & 3.6 & 6.4 & 100.0 & 81.4 \\
		    Multi-task RFT   & 3.6 & 6.6 & 100.0 & 82.0 \\
		    \midrule \multicolumn{8}{l}{\textbf{\textit{Angle (Ceph, 960 samples)}}} \\
		    Sequential RFT   & 4.7 & 52.1 & 99.9 & 52.0 \\
		    Multi-task RFT   & 3.7 & 41.9 & 100.0 & 52.4 \\
		    \bottomrule
		 \end{tabular}
	      \end{adjustbox}
	   \end{table*}

	   \begin{table*}[h]
	      \centering
	      \caption{RFT reward-design ablation: multiplicative composition of the process and answer rewards ($r=r_{\text{format}}+r_{\text{process}} \cdot r_{\text{answer}}$; default) vs. additive composition ($r=r_{\text{format}}+r_{\text{process}}+r_{\text{answer}}$), both trained as a single-stage multi-task RFT from the same SFT model. The detection task defines no process reward and is therefore not affected by the composition. MRE, SR, and MRE$_{<0.1}$ are reported in \%. MAEs are given in millimeters (T/L size, distance) and degrees (angle).}
	      \label{tab:rft_reward_ablation}
	      \tiny
	      \setlength{\tabcolsep}{2pt}
	      \begin{adjustbox}
		 {max width=\textwidth}
		 \begin{tabular}{lrrrr}
		    \toprule  \textbf{Reward} & \textbf{MAE} $\downarrow$ & \textbf{MRE} $\downarrow$ & \textbf{SR} $\uparrow$ & $\textbf{MRE}_{\textbf{\textless0.1}}$ $\uparrow$ \\
		    \midrule \multicolumn{5}{l}{\textbf{\textit{Tumor/Lesion Size (2K samples)}}} \\
		    $r_{\text{process}} \cdot r_{\text{answer}}$ (default)   & 10.5 & 26.7 & 100.0 & 23.4 \\
		    $r_{\text{process}} + r_{\text{answer}}$   & 10.1 & 26.9 & 100.0 & 25.2 \\
		    \midrule \multicolumn{5}{l}{\textbf{\textit{Distance (Ceph+FeTA, 1,100 samples)}}} \\
		    $r_{\text{process}} \cdot r_{\text{answer}}$ (default)   & 3.6 & 6.6 & 100.0 & 82.0 \\
		    $r_{\text{process}} + r_{\text{answer}}$   & 3.6 & 6.7 & 100.0 & 80.9 \\
		    \midrule \multicolumn{5}{l}{\textbf{\textit{Angle (Ceph, 960 samples)}}} \\
		    $r_{\text{process}} \cdot r_{\text{answer}}$ (default)   & 3.7 & 41.9 & 100.0 & 52.4 \\
		    $r_{\text{process}} + r_{\text{answer}}$   & 3.8 & 48.1 & 100.0 & 50.3 \\
		    \bottomrule
		 \end{tabular}
	      \end{adjustbox}
	   \end{table*}

	   \begin{table*}[h]
	      \centering
	      \caption{OOD generalization of the base model (Qwen2.5-VL-7B), SFT, and SFT-RFT (MedVision-V0) on detection and T/L size estimation tasks. Parentheses give the change from the preceding stage (SFT vs.\ base; SFT-RFT vs.\ SFT; \textcolor{green!55!black}{green}: improvement; \textcolor{red!75!black}{red}: decline; \textcolor{gray}{gray}: no change).}
	      \label{tab:ood_generalization} 
	      \tiny
	      \setlength{\tabcolsep}{2pt}
	      \begin{adjustbox}
		 {max width=\textwidth}
		 \begin{tabular}{lrrrrrrr}
		    \toprule 
		    \textbf{Model} & \textbf{R} $\uparrow$ & \textbf{P} $\uparrow$ & \textbf{F1} $\uparrow$ & \textbf{IoU} $\uparrow$ & \textbf{SR} $\uparrow$ & $\textbf{IoU}_{\textbf{\textgreater0.5}}$ $\uparrow$ & $\textbf{IoU}_{\textbf{0.5:0.05:0.95}}$ $\uparrow$ \\
		    \midrule \multicolumn{8}{l}{\textbf{\textit{Plane-OOD Detection: Anatomy (25 regions, 25.6K samples)}}} \\
		    base             & 32.6 & 9.1 & 11.6 & 7.5 & 72.9 & 2.7 & 0.9 \\
		    \quad SFT        & 48.5\impv{+15.9} & 55.7\impv{+46.6} & 49.0\impv{+37.4} & 43.7\impv{+36.2} & 100.0\impv{+27.1} & 47.4\impv{+44.7} & 34.6\impv{+33.7} \\
		    \quad\quad SFT-RFT & 49.3\impv{+0.8} & 54.7\decl{-1.0} & 49.5\impv{+0.5} & 44.3\impv{+0.6} & 100.0\nochg{0.0} & 47.5\impv{+0.1} & 35.6\impv{+1.0} \\
		    \midrule \multicolumn{8}{l}{\textbf{\textit{Plane-OOD Detection: Tumor/Lesion (16 regions, 15.2K samples)}}} \\
		    base             & 43.8 & 5.2 & 8.3 & 4.9 & 73.2 & 0.5 & 0.1 \\
		    \quad SFT        & 30.1\decl{-13.7} & 43.8\impv{+38.6} & 30.7\impv{+22.4} & 23.9\impv{+19.0} & 100.0\impv{+26.8} & 22.9\impv{+22.4} & 10.7\impv{+10.6} \\
		    \quad\quad SFT-RFT & 35.7\impv{+5.6} & 48.8\impv{+5.0} & 36.6\impv{+5.9} & 28.9\impv{+5.0} & 100.0\nochg{0.0} & 28.1\impv{+5.2} & 13.8\impv{+3.1} \\
		    \midrule \midrule 
		    \textbf{Model} & \textbf{R} $\uparrow$ & \textbf{P} $\uparrow$ & \textbf{F1} $\uparrow$ & \textbf{IoU} $\uparrow$ & \textbf{SR} $\uparrow$ & $\textbf{IoU}_{\textbf{\textgreater0.5}}$ $\uparrow$ & $\textbf{IoU}_{\textbf{0.5:0.05:0.95}}$ $\uparrow$ \\
		    \midrule \multicolumn{8}{l}{\textbf{\textit{Target-OOD Detection: Anatomy (27 regions, 5.6K samples)}}} \\
		    base             & 21.5 & 5.6 & 7.3 & 4.7 & 62.3 & 1.2 & 0.2 \\
		    \quad SFT        & 31.3\impv{+9.8} & 44.7\impv{+39.1} & 32.9\impv{+25.6} & 28.5\impv{+23.8} & 98.6\impv{+36.3} & 29.5\impv{+28.3} & 21.2\impv{+21.0} \\
		    \quad\quad SFT-RFT & 34.8\impv{+3.5} & 47.0\impv{+2.3} & 36.0\impv{+3.1} & 31.2\impv{+2.7} & 98.8\impv{+0.2} & 31.7\impv{+2.2} & 23.3\impv{+2.1} \\
		    \midrule \multicolumn{8}{l}{\textbf{\textit{Target-OOD Detection: Tumor/Lesion (3 regions, 7.3K samples)}}} \\
		    base             & 34.1 & 2.6 & 4.4 & 2.5 & 44.7 & 0.1 & 0.0 \\
		    \quad SFT        & 40.6\impv{+6.5} & 50.0\impv{+47.4} & 40.2\impv{+35.8} & 32.7\impv{+30.2} & 99.8\impv{+55.1} & 35.4\impv{+35.3} & 17.9\impv{+17.9} \\
		    \quad\quad SFT-RFT & 47.4\impv{+6.8} & 51.5\impv{+1.5} & 45.5\impv{+5.3} & 38.0\impv{+5.3} & 99.9\impv{+0.1} & 41.7\impv{+6.3} & 23.5\impv{+5.6} \\
		    \midrule \midrule 
		    \textbf{Model} & \textbf{MAE} $\downarrow$ & \textbf{MRE} $\downarrow$ & \textbf{SR} $\uparrow$ & $\textbf{MRE}_{\textbf{\textless0.1}}$ $\uparrow$ \\
		    \midrule
		    \multicolumn{8}{l}{\textbf{\textit{Plane-OOD Tumor/Lesion Size (16 labels, 4.5K samples)}}} \\
		    base             & 16240.6 & 44054.2 & 96.9 & 0.4 \\
		    \quad SFT        & 20.3\impv{-16220.3} & 42.1\impv{-44012.1} & 100.0\impv{+3.1} & 6.8\impv{+6.4} \\
		    \quad\quad SFT-RFT & 18.2\impv{-2.1} & 36.4\impv{-5.7} & 100.0\nochg{0.0} & 8.1\impv{+1.3} \\
		    \midrule \midrule \textbf{Model} & \textbf{MAE} $\downarrow$ & \textbf{MRE} $\downarrow$ & \textbf{SR} $\uparrow$ & $\textbf{MRE}_{\textbf{\textless0.1}}$ $\uparrow$ \\
		    \midrule
		    \multicolumn{8}{l}{\textbf{\textit{Target-OOD Tumor/Lesion Size (5 labels, 764 samples)}}} \\
		    base             & 25989.2 & 132518.7 & 99.2 & 0.1 \\
		    \quad SFT        & 10.9\impv{-25978.3} & 41.6\impv{-132477.1} & 100.0\impv{+0.8} & 14.1\impv{+14.0} \\
		    \quad\quad SFT-RFT & 9.5\impv{-1.4} & 34.7\impv{-6.9} & 100.0\nochg{0.0} & 17.7\impv{+3.6} \\
		    \bottomrule
		 \end{tabular}
	      \end{adjustbox}
	   \end{table*}

	   \section{Per-Model Performance Profiles}
	   \label{appendix:sec:per_model_profiles}
\begin{figure*}[h]
      \centering
      \includegraphics[width=1\linewidth]{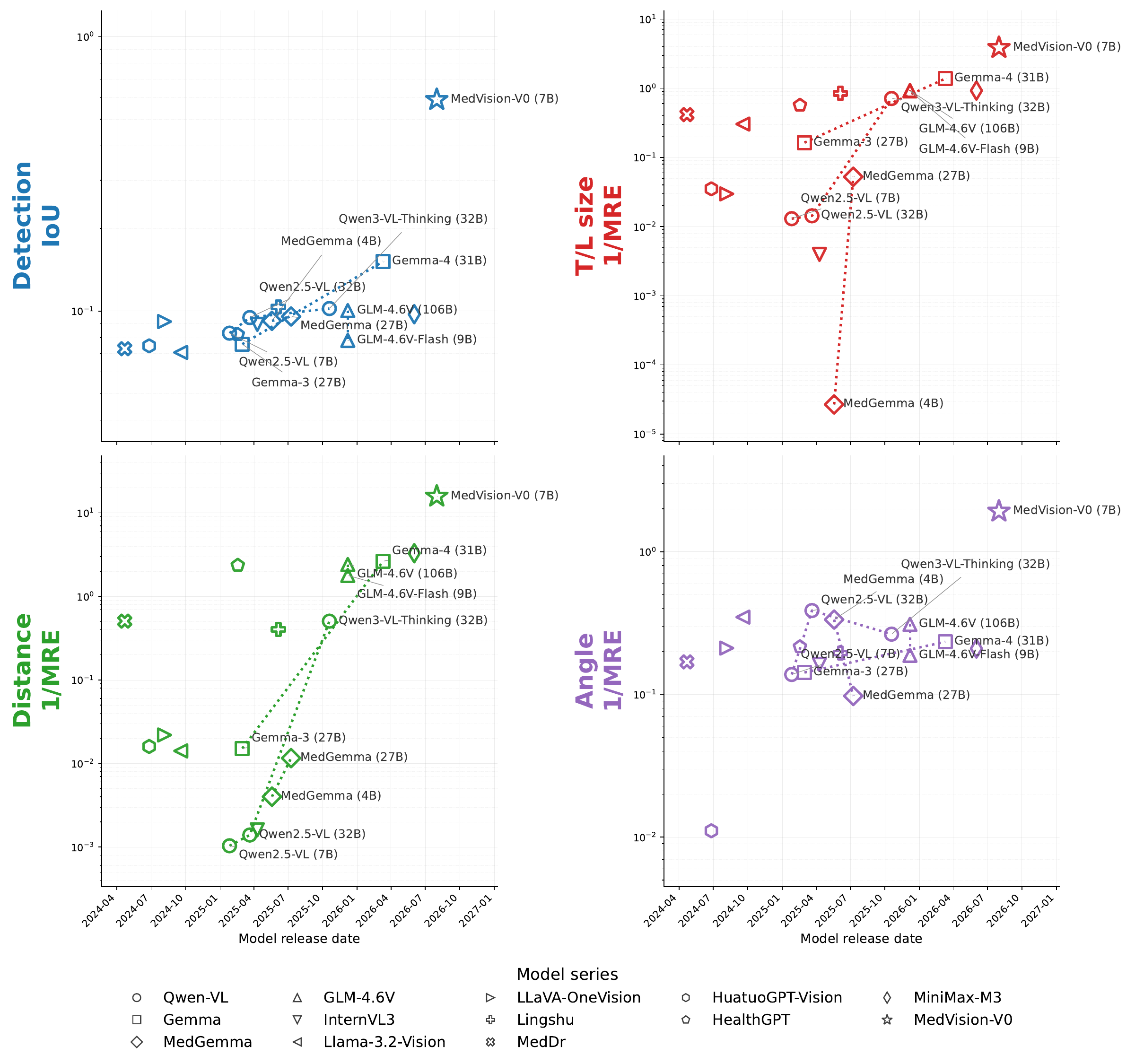}
      \caption{Overview of model performance against model release date. From left to right and top to bottom: detection IoU, and the inverse mean relative error (1/MRE) of T/L size estimation, distance measurement, and angle measurement, on a logarithmic scale; \textbf{\textit{higher is better in all panels}}. Markers denote model series, dotted lines connect successive releases within a series, and the star marks MedVision-V0.}
      \label{appendix:fig:leaderboard_timeline}
   \end{figure*}

\begin{figure*}[p]
      \centering
      \begin{adjustbox}
         {max width=\textwidth, max height=0.88\textheight}
         \includegraphics{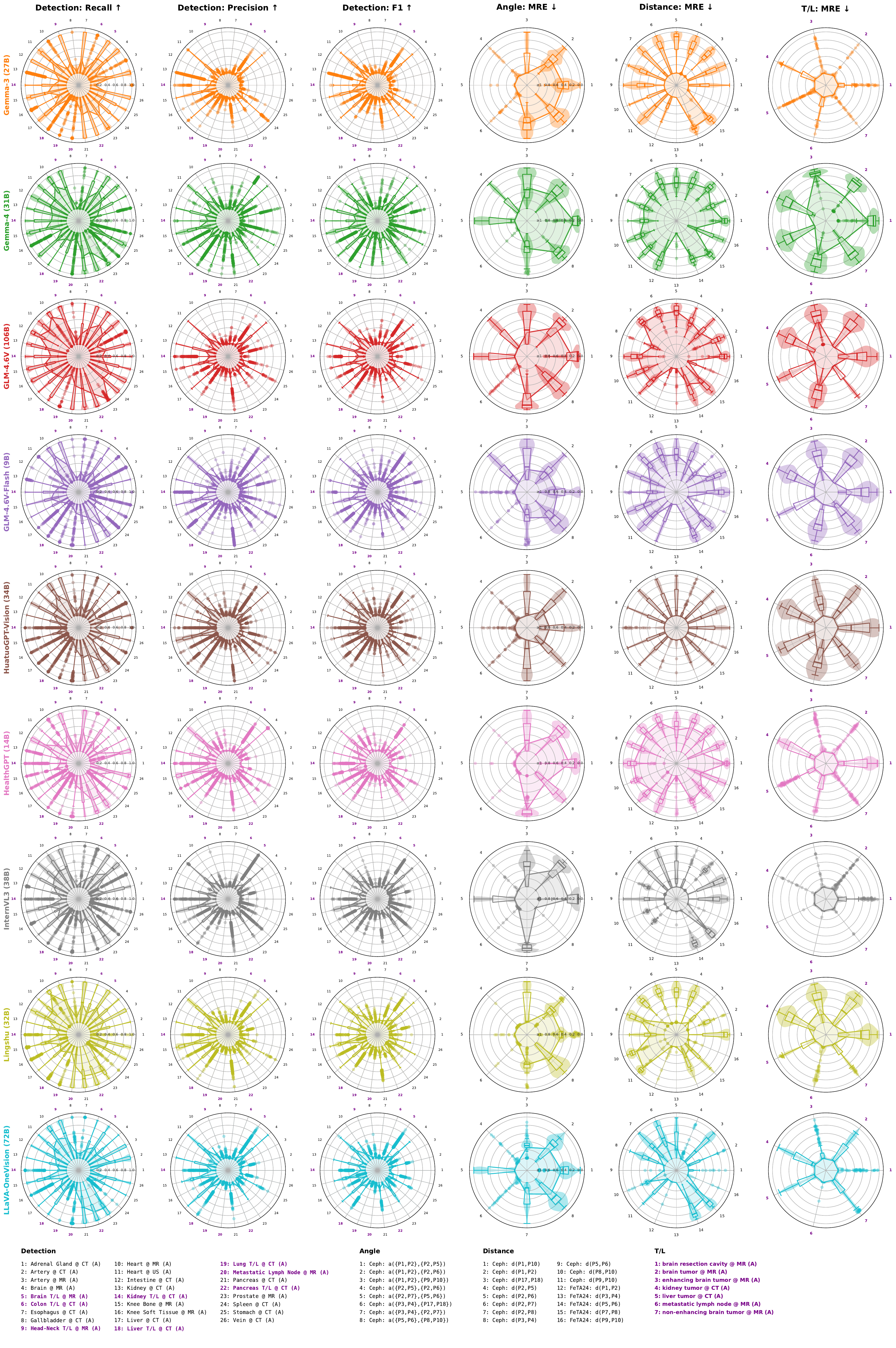}
      \end{adjustbox}
	   \caption{Per-model performance profiles of the evaluated VLMs (part 1 of 2; the first nine models). Each row shows one model's results as six radar charts: recall, precision, and F1 in the detection task; MRE in the angle and distance measurement tasks; and MRE in the T/L size task. The MRE axes are inverted so that the outermost ring corresponds to MRE $=0$ and the center to MRE $\geq 1$, i.e., \textbf{\textit{larger polygons always indicate better performance}}. Violin and box plots on each spoke show the model's per-sample metric distribution, with dots marking outliers beyond the $1.5\times$IQR whiskers. MedVision-V0 is shown in Figure~\ref{fig:sft_rft_training_strategy}.}
      \label{appendix:fig:radar_grid_part1}
   \end{figure*}

\begin{figure*}[p]
      \centering
      \begin{adjustbox}
         {max width=\textwidth, max height=0.88\textheight}
         \includegraphics{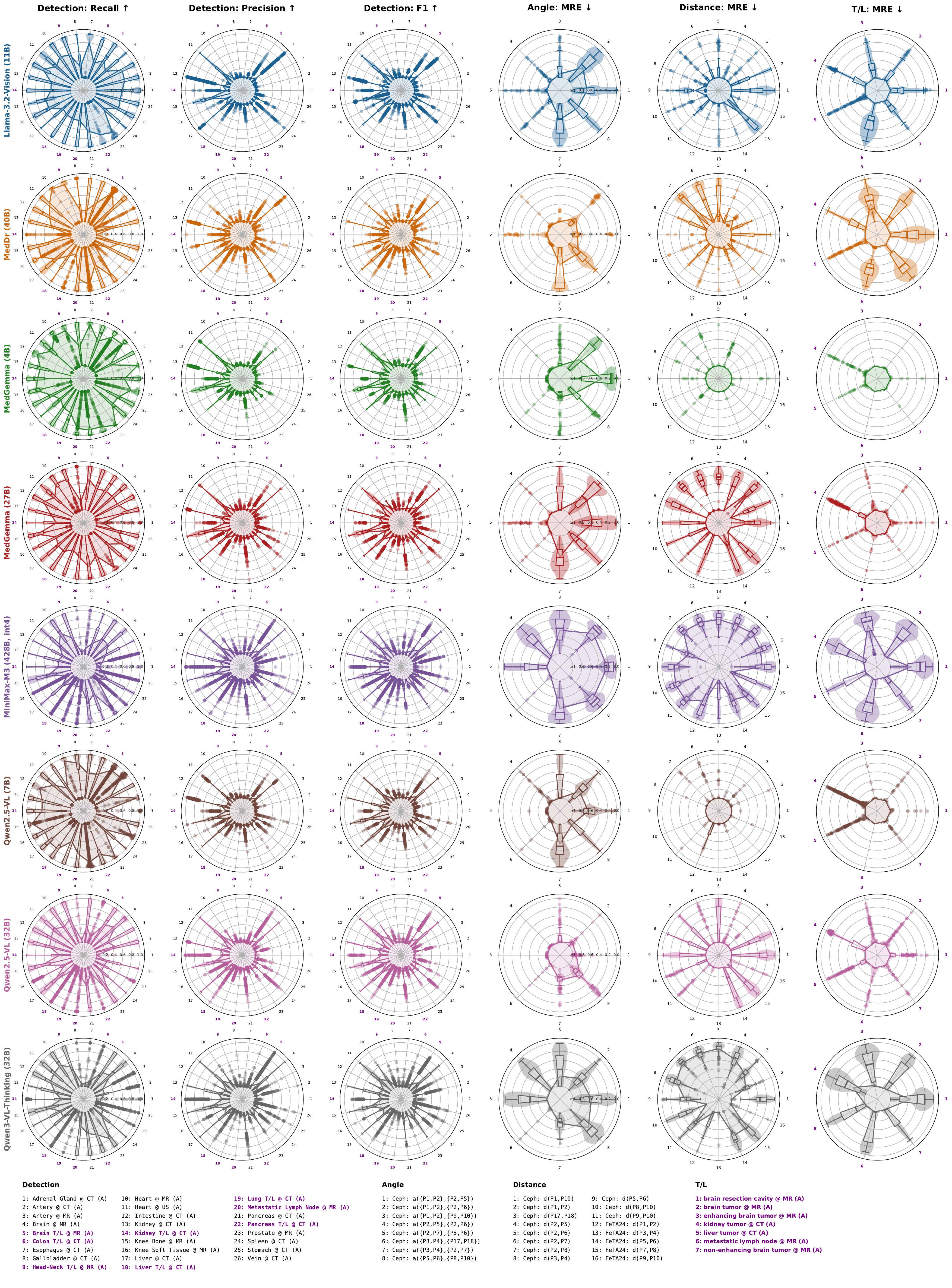}
      \end{adjustbox}
      \caption{Per-model performance profiles of the evaluated VLMs (part 2 of 2; the remaining eight models).}
      \label{appendix:fig:radar_grid_part2}
   \end{figure*}

	   Figure~\ref{appendix:fig:leaderboard_timeline} provides a general overview of model performance against model release date on the three benchmark task families. Figures~\ref{appendix:fig:radar_grid_part1} and~\ref{appendix:fig:radar_grid_part2} present per-model performance profiles for the 17 evaluated VLMs: each row shows one model's results as six radar charts (detection recall, precision, and F1; angle and distance MRE; T/L size MRE), with the model's per-sample metric distributions overlaid as violin and box plots on each spoke.

	   \section{Details of Detection Performance}
	   \label{appendix:subsec:detection_performance}
\begin{figure*}[h]
	      \begin{center}
		 \includegraphics[width=1\linewidth]{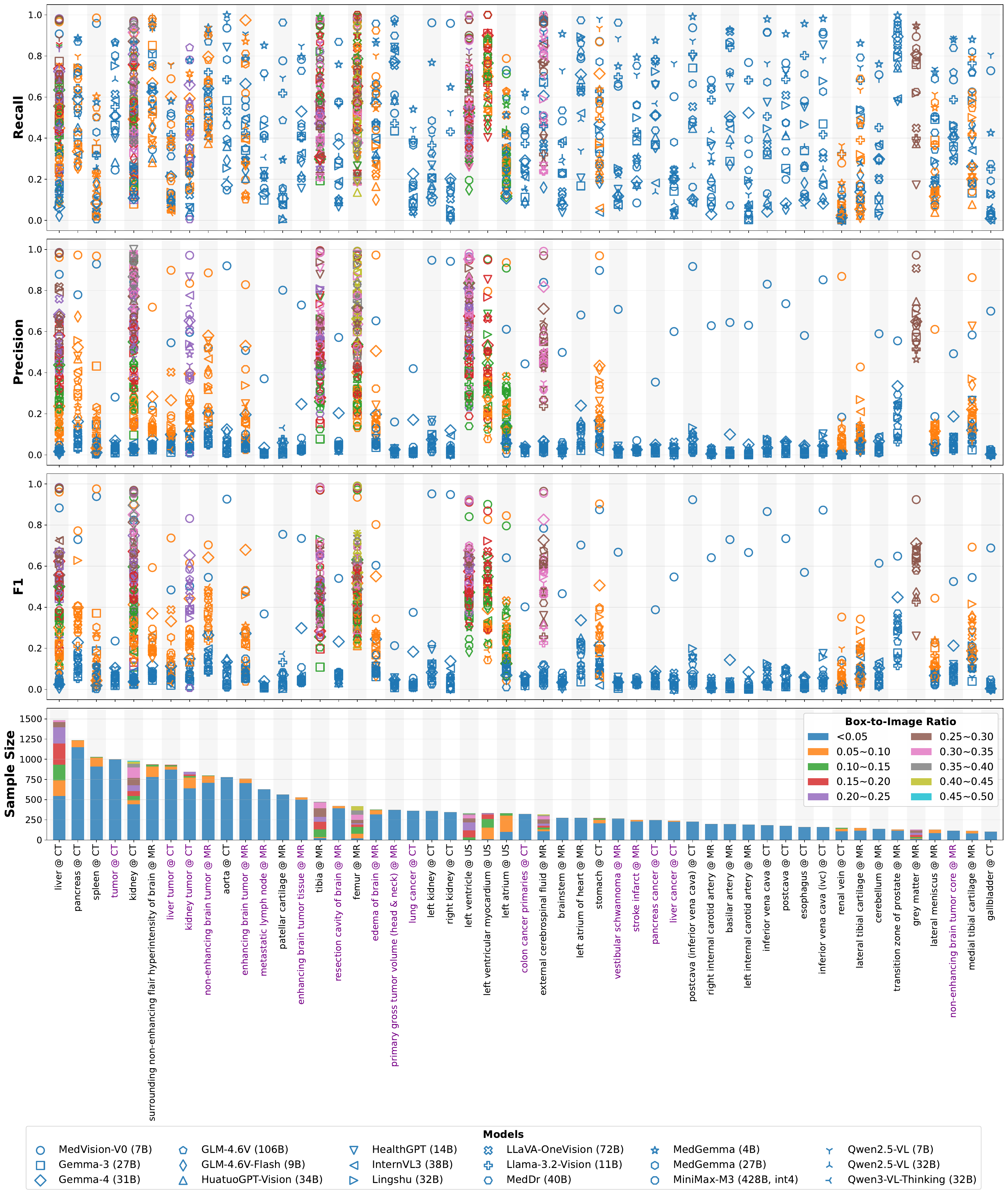}
	      \end{center}
	      \caption{Label-level detection performance of the evaluated VLMs at each box-image-ratio group, where box-image-ratio is the area ratio of the bounding box to the image. Top: recall, precision, and F1 score for each label and box-image-ratio group. Bottom: the composition of box-image-ratio groups for each label.}
	      \label{appendix:fig:detection_label_level}
	   \end{figure*}

	   \noindent
	   \textbf{Label-Level Detection Performance.} Figure~\ref{appendix:fig:detection_label_level} shows the label-level detection performance of VLMs at each box-image-ratio group. Box-image-ratio is defined as the area ratio of the bounding box to the image. Labels are grouped into different box-image-ratio ranges. We visualized the composition of box-image-ratio groups for each label at the bottom of Figure~\ref{appendix:fig:detection_label_level}. For each box-image-ratio group and label combination, we visualized recall, precision, and F1 score. Figure~\ref{appendix:fig:detection_label_level} is a detailed map of VLM performance at various target sizes and labels.

\begin{figure*}[h]
	      \begin{center}
		 \includegraphics[width=1\linewidth]{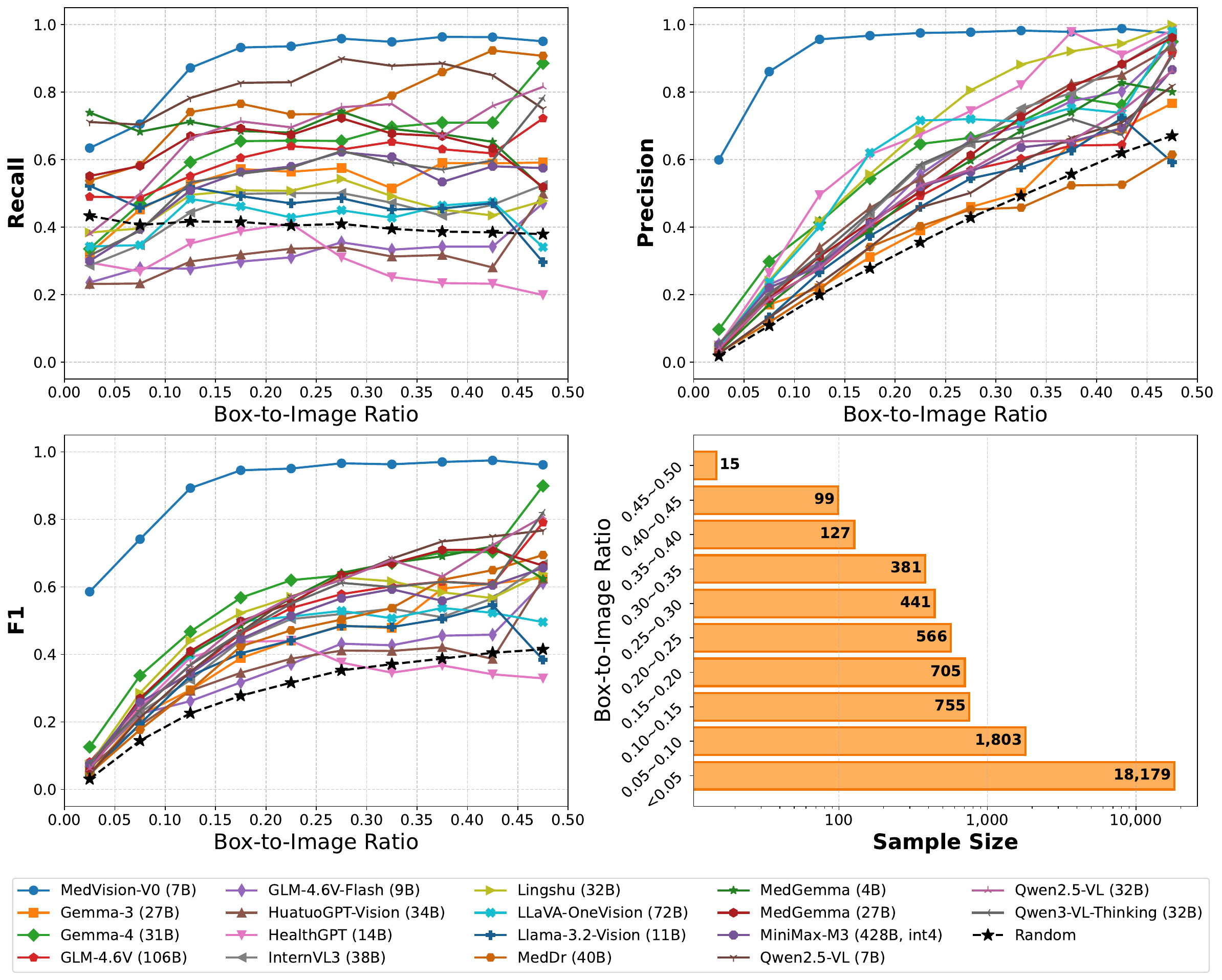}
	      \end{center}
	      \caption{Effect of target size, quantified by box-image-ratio (the area ratio of the bounding box to the image), on detection performance. All detection targets are grouped by box-image-ratio, and each model's overall detection performance is computed within each group. \textbf{\textit{Random}} denotes a random-guessing baseline in which each sample's metrics are averaged over 100 uniformly sampled boxes.}
	      \label{appendix:fig:effect_of_target_size_detection_performance}
	   \end{figure*}

	   \noindent
	   \textbf{Effect of Target Size.} Figure~\ref{appendix:fig:detection_label_level} shows a trend of increasing detection performance with larger target sizes (box-image-ratio). To further illustrate this effect, we grouped all detection targets into different box-image-ratio groups, and calculated the overall detection performance for each group. As shown in Figure~\ref{appendix:fig:effect_of_target_size_detection_performance}, precision and recall are positively correlated with relative target size (box-image-ratio). As a reference, the figure includes a random-guessing baseline (\textbf{\textit{Random}}): for each test sample, we simulate 100 random boxes whose width and height are drawn uniformly at random (from a 3-pixel minimum up to the full image extent) and whose position is drawn uniformly among all placements fully inside the image; per-sample metrics are averaged over the 100 simulations and then aggregated per box-image-ratio group in the same way as for the evaluated models. The baseline's rising precision and F1 reflect that larger targets are easier to overlap by chance. MedVision-V0 consistently outperforms all off-the-shelf models across all target sizes.

	   \section{Clinical Decision Agreement Analysis}
	   \label{appendix:sec:cda} This appendix details the clinical decision agreement (CDA) analysis summarized in Section~\ref{subsec:cda}. CDA re-scores the parsed benchmark outputs of all evaluated models: each predicted measurement is converted into a clinical category through a published cutoff table, and the predicted category is compared with the category derived from the ground truth measurement. Because both sides pass through the same cutoff table, any disagreement is attributable to measurement error alone.

	   \noindent
	   \textbf{CDA Tasks.} We evaluate three CDA tasks. \textbf{\textit{(i--ii) Maxillary and mandibular position:}} the anteroposterior position of the maxilla and the mandible of the Ceph-Bio-400 samples, revealed by the SNA (sella--nasion--A point) and SNB (sella--nasion--B point) angles, respectively. Each is a three-class task: an angle within the Steiner norm band ($82\pm2$ degrees for SNA, $80\pm2$ degrees for SNB)~\citep{Cephalometrics1953} indicates a normal position, an angle below the band a retrusive position, and an angle above the band a protrusive position. \textbf{\textit{(iii) AJCC kidney tumor staging:}} the kidney tumor category from tumor size, a four-class task (T staging: T1a $\leq40$ mm, T1b $>$40--70 mm, T2a $>$70--100 mm, T2b $>100$ mm)~\citep{AJCC2017}, applied to the per-slice T/L size predictions on the KiPA22 and KiTS23 datasets. Tumor size measured on cross-sectional imaging is the primary criterion for the clinical T staging of organ-confined kidney tumors~\citep{AJCC2017, Update2021}, so these cutoffs constitute clinically established decision boundaries for size measurement. The benchmark measures the tumor's major axis on each annotated 2D slice, a standard clinical practice consistent with the RECIST guideline (v1.1) for tumor response assessment~\citep{New2009}. The kidney tumor staging task therefore compares each slice-level prediction with the category of the ground truth measurement, evaluating whether the model's measurement error is large enough to change the staging category.

	   \noindent
	   \textbf{Metrics.} Accuracy (Acc) is the fraction of parsed predictions whose category matches the reference, and success rate (SR) is the fraction of predictions from which a measurement could be parsed. Chance-corrected agreement is measured with Cohen's $\kappa$ and quadratic-weighted $\kappa_w$. Cohen's $\kappa$ measures exact-category agreement beyond chance:
	   \begin{align}
	      \kappa = \frac{p_o - p_e}{1 - p_e},
	   \end{align}
	   where $p_o$ is the observed fraction of samples whose predicted and reference categories match, and $p_e$ is the matching fraction expected by chance from the marginal distributions. All three CDA tasks have ordered categories, and $\kappa_w$, which gives partial credit to near-miss categories, is the primary statistic:
	   \begin{align}
	      \kappa_w = 1 - \frac{\sum_{i=1}^{k}\sum_{j=1}^{k} w_{ij} O_{ij}}{\sum_{i=1}^{k}\sum_{j=1}^{k} w_{ij} E_{ij}}, \quad w_{ij} = \frac{(i-j)^{2}}{(k-1)^{2}},
	   \end{align}
	   where $k$ is the number of categories, $O_{ij}$ is the observed fraction of samples with predicted category $i$ and reference category $j$, $E_{ij}$ is the fraction expected by chance from the marginal distributions, and $w_{ij}$ is the quadratic disagreement weight; the unweighted $\kappa$ corresponds to the special case with $w_{ij}=0$ for $i=j$ and $w_{ij}=1$ otherwise. We interpret values with the standard bands (0.21--0.40 fair, 0.41--0.60 moderate, 0.61--0.80 substantial)~\citep{Measurement1977}. Uncertainty is estimated with a percentile bootstrap over whole imaging volumes (4,000 resamples), from which we also derive a one-sided p-value for agreement above chance. Because $\kappa$ depends on how close the cohort's values lie to the cutoffs, it is comparable across models within a CDA task but not across tasks.

	   \noindent
	   \textbf{Results.} Table~\ref{tab:cda_agreement} reports the decision agreement of all 18 models on the three CDA tasks. On the cephalometric tasks, no off-the-shelf model except LLaVA-OneVision on SNB ($\kappa_w=0.243$) achieves agreement significantly above chance. On the kidney tumor staging task, off-the-shelf models reach at most fair ordinal agreement ($\kappa_w\leq0.361$), and the weakest collapse toward a single category: InternVL3 assigns $99.1\%$ of its parsed predictions to T2b and MedGemma (4B) assigns $81.4\%$, although only $6.8\%$ of the ground truth measurements fall in that category. This collapse reflects a failure of physical scale: these models emit tumor sizes far beyond the plausible anatomical range regardless of the input image (e.g., InternVL3's median predicted major axis is $5{,}590$ mm against a ground truth median of $38$ mm), so nearly every prediction exceeds the highest staging cutoff of $100$ mm and folds into the top category. MedVision-V0 is the only model whose agreement is significantly above chance on all three CDA tasks, attaining fair agreement on SNA ($\kappa_w=0.332$), moderate agreement on SNB ($\kappa_w=0.439$), and substantial agreement on kidney tumor staging ($\kappa_w=0.726$) at full parsing coverage. Although $\kappa_w$ values are not strictly comparable across CDA tasks, the lower agreement on the two angle-derived tasks than on the size-derived staging task mirrors the benchmark results: angle measurement remains harder for MedVision-V0 than length measurements such as landmark distance and tumor size (Tables~\ref{tab:ad_measurement_performance} and \ref{tab:tl_size_performance}).

	   \begin{table*}[h]
	      \centering
	      \caption{Clinical decision agreement of all evaluated models: the category of the predicted measurement vs. the category of the ground truth measurement, both derived through the same published cutoff table. Acc: category agreement on parsed samples; $\kappa$: Cohen's kappa; $\kappa_w$: quadratic-weighted kappa; CI and p: 95\% percentile bootstrap confidence interval and one-sided p-value for agreement above chance, computed for $\kappa_w$ by resampling whole imaging volumes (4,000 resamples); SR: success rate, the fraction of predictions from which a measurement could be parsed, in \%. $\kappa$ values are not comparable across CDA tasks.}
	      \label{tab:cda_agreement} 
	      \tiny
	      \setlength{\tabcolsep}{2pt}
	      \begin{adjustbox}
		 {max width=\columnwidth}
		 \begin{tabular}{lrrrrrr}
		    \toprule  \textbf{Model} & \textbf{Acc} $\uparrow$ & $\kappa$ $\uparrow$ & $\kappa_w$ $\uparrow$ & \textbf{95\% CI} & \textbf{p} & \textbf{SR} $\uparrow$ \\
            \midrule \multicolumn{7}{l}{\textbf{\textit{SNA maxillary position (Ceph-Bio-400; 120 samples)}}} \\
            MedVision-V0 (7B)       & 0.458 & 0.183 & 0.332 & [0.163, 0.484] & \textless 0.001 & 100.0 \\
            InternVL3 (38B)         & 0.358 & 0.078 & 0.119 & [-0.048, 0.283] & 0.080 & 100.0 \\
            Gemma4 (31B)            & 0.393 & 0.066 & -0.033 & [-0.294, 0.223] & 0.609 & 46.7 \\
            Lingshu (32B)           & 0.325 & 0.052 & 0.090 & [-0.029, 0.206] & 0.071 & 100.0 \\
            Qwen2.5-VL (7B)         & 0.303 & 0.022 & 0.037 & [-0.039, 0.115] & 0.176 & 99.2 \\
            GLM-4.6V-Flash (9B)     & 0.322 & 0.021 & 0.004 & [-0.073, 0.090] & 0.457 & 98.3 \\
            Llama3.2-Vision (11B)   & 0.300 & 0.017 & -0.002 & [-0.031, 0.029] & 0.556 & 100.0 \\
            LLaVA-OneVision (72B)   & 0.300 & 0.000 & 0.001 & [-0.169, 0.174] & 0.499 & 100.0 \\
            Qwen2.5-VL (32B)        & 0.283 & 0.000 & 0.000 & [0.000, 0.000] & 1.000 & 100.0 \\
            Gemma3 (27B)            & 0.283 & -0.006 & 0.014 & [-0.093, 0.116] & 0.404 & 100.0 \\
            HuatuoGPT-Vision (34B)  & 0.293 & -0.014 & -0.044 & [-0.190, 0.112] & 0.712 & 96.7 \\
            MedDr (40B)             & 0.278 & -0.014 & -0.025 & [-0.123, 0.070] & 0.709 & 90.0 \\
            Qwen3-VL-Thinking (32B) & 0.295 & -0.018 & 0.001 & [-0.056, 0.070] & 0.539 & 79.2 \\
            MedGemma (4B)           & 0.291 & -0.020 & -0.023 & [-0.194, 0.148] & 0.606 & 97.5 \\
            MiniMax-M3 (428B)       & 0.250 & -0.066 & -0.027 & [-0.269, 0.190] & 0.590 & 30.0 \\
            HealthGPT-L14 (14B)     & 0.250 & -0.082 & -0.132 & [-0.289, 0.027] & 0.951 & 100.0 \\
            MedGemma (27B)          & 0.233 & -0.082 & -0.153 & [-0.332, 0.030] & 0.948 & 75.0 \\
            GLM-4.6V (106B)         & 0.272 & -0.121 & -0.018 & [-0.164, 0.132] & 0.595 & 85.8 \\
            \midrule \multicolumn{7}{l}{\textbf{\textit{SNB mandibular position (Ceph-Bio-400; 120 samples)}}} \\
            MedVision-V0 (7B)       & 0.483 & 0.227 & 0.439 & [0.292, 0.575] & \textless 0.001 & 100.0 \\
            LLaVA-OneVision (72B)   & 0.425 & 0.153 & 0.243 & [0.097, 0.392] & \textless 0.001 & 100.0 \\
            HealthGPT-L14 (14B)     & 0.375 & 0.055 & 0.090 & [-0.088, 0.258] & 0.155 & 100.0 \\
            MedGemma (27B)          & 0.367 & 0.055 & 0.122 & [-0.066, 0.305] & 0.100 & 90.8 \\
            MedDr (40B)             & 0.345 & 0.050 & 0.080 & [-0.057, 0.211] & 0.110 & 96.7 \\
            MiniMax-M3 (428B)       & 0.341 & 0.029 & 0.050 & [-0.209, 0.315] & 0.357 & 36.7 \\
            GLM-4.6V-Flash (9B)     & 0.308 & 0.020 & 0.012 & [-0.045, 0.067] & 0.331 & 100.0 \\
            Llama3.2-Vision (11B)   & 0.308 & 0.018 & 0.019 & [-0.024, 0.079] & 0.256 & 100.0 \\
            Lingshu (32B)           & 0.300 & -0.001 & -0.002 & [-0.103, 0.102] & 0.526 & 100.0 \\
            Qwen2.5-VL (7B)         & 0.294 & -0.007 & -0.011 & [-0.067, 0.041] & 0.667 & 99.2 \\
            Gemma3 (27B)            & 0.283 & -0.021 & -0.032 & [-0.133, 0.060] & 0.744 & 100.0 \\
            Qwen2.5-VL (32B)        & 0.277 & -0.028 & -0.034 & [-0.107, 0.026] & 0.852 & 99.2 \\
            GLM-4.6V (106B)         & 0.281 & -0.034 & -0.107 & [-0.242, 0.020] & 0.948 & 95.0 \\
            Qwen3-VL-Thinking (32B) & 0.302 & -0.035 & -0.043 & [-0.129, 0.025] & 0.882 & 71.7 \\
            Gemma4 (31B)            & 0.273 & -0.037 & -0.017 & [-0.159, 0.119] & 0.583 & 91.7 \\
            InternVL3 (38B)         & 0.283 & -0.049 & -0.077 & [-0.231, 0.074] & 0.842 & 100.0 \\
            HuatuoGPT-Vision (34B)  & 0.263 & -0.082 & -0.120 & [-0.277, 0.038] & 0.927 & 95.0 \\
            MedGemma (4B)           & 0.250 & -0.092 & -0.154 & [-0.283, -0.026] & 0.993 & 96.7 \\
            \midrule \multicolumn{7}{l}{\textbf{\textit{AJCC kidney tumor staging (KiPA22 + KiTS23; 1,025 samples)}}} \\
            MedVision-V0 (7B)       & 0.657 & 0.379 & 0.726 & [0.428, 0.858] & \textless 0.001 & 100.0 \\
            Gemma4 (31B)            & 0.402 & 0.203 & 0.361 & [0.166, 0.522] & \textless 0.001 & 97.9 \\
            Qwen3-VL-Thinking (32B) & 0.405 & 0.159 & 0.342 & [0.106, 0.509] & \textless 0.001 & 96.7 \\
            GLM-4.6V-Flash (9B)     & 0.356 & 0.153 & 0.238 & [0.063, 0.421] & \textless 0.001 & 95.9 \\
            Lingshu (32B)           & 0.364 & 0.140 & 0.266 & [0.079, 0.439] & \textless 0.001 & 99.9 \\
            GLM-4.6V (106B)         & 0.352 & 0.127 & 0.235 & [0.050, 0.408] & 0.002 & 93.2 \\
            HuatuoGPT-Vision (34B)  & 0.459 & 0.120 & 0.185 & [0.051, 0.277] & \textless 0.001 & 98.0 \\
            MiniMax-M3 (428B)       & 0.323 & 0.113 & 0.266 & [0.123, 0.387] & \textless 0.001 & 95.0 \\
            HealthGPT-L14 (14B)     & 0.267 & 0.101 & 0.182 & [0.052, 0.313] & 0.001 & 100.0 \\
            MedGemma (27B)          & 0.180 & 0.051 & 0.086 & [0.017, 0.170] & 0.004 & 51.0 \\
            Gemma3 (27B)            & 0.198 & 0.050 & 0.088 & [0.020, 0.173] & 0.003 & 98.6 \\
            MedDr (40B)             & 0.326 & 0.026 & 0.095 & [-0.031, 0.213] & 0.072 & 83.6 \\
            LLaVA-OneVision (72B)   & 0.220 & 0.025 & 0.032 & [-0.038, 0.100] & 0.173 & 100.0 \\
            MedGemma (4B)           & 0.149 & 0.019 & 0.039 & [0.009, 0.075] & 0.005 & 90.4 \\
            Qwen2.5-VL (32B)        & 0.142 & 0.014 & 0.060 & [0.005, 0.128] & 0.015 & 99.9 \\
            InternVL3 (38B)         & 0.073 & 0.004 & -0.000 & [-0.002, 0.002] & 0.550 & 100.0 \\
            Llama3.2-Vision (11B)   & 0.286 & 0.002 & 0.046 & [-0.049, 0.147] & 0.207 & 99.4 \\
            Qwen2.5-VL (7B)         & 0.092 & -0.059 & -0.051 & [-0.215, 0.049] & 0.658 & 92.3 \\
            \bottomrule
         \end{tabular}
      \end{adjustbox}
   \end{table*}

   \section{Intended Use, License \& Data Privacy}
   \label{appendix:sec:intended_use_privacy}

   \noindent
   \textbf{Intended Use.} The MedVision dataset, benchmark, and the fine-tuned model MedVision-V0 are released for research purposes only. They are intended to support the study of quantitative reasoning in vision-language models, reproducible benchmarking, and the development of stronger models. None of these artifacts is a medical device, none has been validated in a clinical setting, and none may be used for diagnosis, treatment planning, patient management, or any other clinical decision-making. As reported in Section~\ref{sec:results}, even the best model evaluated here retains measurement errors above clinical tolerances. Any use beyond research would require independent clinical validation and the appropriate regulatory approval.

   \noindent
   \textbf{License.} MedVision is released under the Creative Commons Attribution 4.0 International (CC-BY 4.0) license. Users are permitted to utilize, adapt, and build upon this dataset for both academic and commercial purposes, provided that appropriate credit is given. Users are reminded that MedVision acts as a meta-dataset built upon various publicly available source datasets. While the annotations provided by MedVision are covered by the CC-BY 4.0 license, any downstream application must continue to comply with the specific usage terms and licensing requirements stipulated by the curators of the original raw imaging data. It is the responsibility of the user to ensure that their application of this data aligns with the license agreements of all constituent source datasets.

   \noindent
   \textbf{Data Privacy.} All source imaging datasets incorporated in MedVision were publicly released in anonymized form by their respective curators. The annotations contributed by MedVision (bounding boxes, tumor/lesion size measurements, and angle/distance measurements) are purely geometric descriptors of image content and contain no information that could identify individual subjects.
\end{document}